%% file: main.tex
\documentclass[letterpaper, 10 pt, conference]{ieeeconf}
\IEEEoverridecommandlockouts
\usepackage{cite, balance}
\usepackage{amsmath,amssymb,amsfonts}

\usepackage{amsthm}
\usepackage{algorithmic}
\usepackage{mathtools}
\usepackage[font=small,skip=0pt]{caption}

\usepackage{babel}
\usepackage{xurl}

\theoremstyle{definition} 
\newtheoremstyle{italdefinition} 
{} 
{} 
{} 
{} 
{\itshape} 
{:} 
{3pt} 
{} 
\theoremstyle{italdefinition}
\newtheorem{definition}{Definition}

\usepackage{caption}
\usepackage{ifthen}
\newboolean{includeFigures}
\setboolean{includeFigures}{true}  
\newboolean{includeAppendix}
\setboolean{includeAppendix}{false}
\usepackage{subcaption}

\makeatletter
\renewcommand\paragraph{\@startsection{paragraph}{4}{\z@}%
                                    {0.3ex \@plus1ex \@minus.2ex}%
                                    {-1em}%
                                    {\normalfont\normalsize\bfseries}}
\makeatother

\usepackage[utf8]{inputenc}
\usepackage{pgfplots}
\DeclareUnicodeCharacter{2212}{−}
\usepgfplotslibrary{groupplots,dateplot}
\usetikzlibrary{patterns,shapes.arrows}
\pgfplotsset{compat=newest}
\pgfplotsset{every tick label/.append style={font=\tiny}}
\usetikzlibrary{shapes.geometric}
\usetikzlibrary{calc}

\usepackage[linesnumbered,ruled,vlined]{algorithm2e}
\usepackage{amssymb}

\usepackage{titlesec}
\titlespacing*{\section}{0pt}{4pt}{2pt}

\usepackage{hyperref}
\hypersetup{
    colorlinks=true,
    linkcolor=blue,
    filecolor=magenta,      
    urlcolor=blue,
    pdftitle={Overleaf Example},
    pdfpagemode=FullScreen,
    }
\usepackage{graphicx}
\usepackage{textcomp}
\usepackage{xcolor}
\def\BibTeX{{\rm B\kern-.05em{\sc i\kern-.025em b}\kern-.08em
    T\kern-.1667em\lower.7ex\hbox{E}\kern-.125emX}}

\title{\LARGE \bf
Quasi-static Soft Fixture Analysis of Rigid and Deformable Objects 
}

\author{Yifei Dong$^{*}$ and Florian T. Pokorny$^{*}$
\thanks{* The authors are with the division of Robotics, Perception and Learning, KTH Royal %
Institute of Technology, 100 44 Stockholm, Sweden, {\tt\small \{yifeid, fpokorny\}@kth.se}. Funded %
by the European Commission under the Horizon Europe Framework Programme project SoftEnable, grant %
number 101070600, \url{http://softenable.eu/}}
}

\begin{document}

\maketitle
\thispagestyle{empty}
\pagestyle{empty}

\begin{abstract}
We present a sampling-based approach to reasoning about the caging-based manipulation of rigid and a simplified class of deformable 3D objects subject to energy constraints. Towards this end, we propose the notion of soft fixtures extending earlier work on energy-bounded caging to include a broader set of energy function constraints and settings, such as gravitational and elastic potential energy of 3D deformable objects. Previous methods focused on establishing provably correct algorithms to compute lower bounds or analytically exact estimates of escape energy for a very restricted class of known objects with low-dimensional C-spaces, such as planar polygons. We instead propose a practical sampling-based approach that is applicable in higher-dimensional C-spaces but only produces a sequence of upper-bound estimates that, however, appear to converge rapidly to actual escape energy. We present 8 simulation experiments demonstrating the applicability of our approach to various complex quasi-static manipulation scenarios. Quantitative results indicate the effectiveness of our approach in providing upper-bound estimates for escape energy in quasi-static manipulation scenarios. Two real-world experiments also show that the computed normalized escape energy estimates appear to correlate strongly with the probability of escape of an object under randomized pose perturbation\footnote{For additional
material, visit \url{https://sites.google.com/view/softfixture/}.}.
\end{abstract}

\section{Introduction}
Fixturing, which is the task of restraining the poses of a rigid object under noise and external perturbation, forms a key part of robotic manipulation and serves as an essential step towards achieving robust and dexterous manipulation~\cite{bicchi2000robotic}. Classical grasping approaches, such as form and force closure \cite{nguyen1988constructing}, utilize point contacts to attain complete control over the pose of a grasped object, but are restricted to rigid objects and are susceptible to grasp instability issues arising from perceptual noise and external disturbances. Furthermore, these approaches only capture a small subset of strategies humans employ to restrain the poses of an object - for example, when we utilize gravity in combination with support surfaces from tools and our hands to interact with soft, deformable or fragile objects such as clothes or food items. In this work, we take the first steps towards an extension of fixturing to ``soft fixtures'', where ``soft'' indicates both that we consider not just ``hard'' collision constraints but also ``soft constraints'' formulated in terms of maximal thresholds of a suitable energy function and that the framework is applicable to soft objects. Our approach generalizes the idea of energy-bounded cages of rigid bodies - where a rigid object is constrained to a bounded path component by collisions and a simple potential energy function. Unlike past works \cite{mahler2016energy, mahler2018synthesis} on energy-bounded cages, which employed volumetric configuration space (C-space) representations and were limited to the gravitational potential energy of simple 2D rigid objects, we show that a sampling-based approach to the study of soft fixtures can be applied with more general energy functions and even challenging settings involving both deformable and rigid bodies. The proposed ``soft fixture''-based approach can, in our view, be seen as a starting point for an additional avenue towards robotic manipulation that, like classical form and force closure, is fully physics-based and explainable, but opens the possibility to describe and analyze novel interaction scenarios involving both rigid and deformable objects.

The primary contributions of this work\footnote{This work is an extension of our workshop contribution\cite{dongsoft}.} are:
(1) An extension of caging-based manipulation to soft fixtures capable of capturing a novel class of rigid and deformable object manipulation scenarios involving potential energy constraints; 
(2) The introduction of sampling-based algorithms for the estimation of upper bounds of escape energy of soft fixtures that appear to converge rapidly; 
(3) A first analysis of escape energy in practical quasi-static manipulation scenarios involving diverse rigid and deformable objects, supported by physical experiments that assess the correlation between normalized escape energy and grasp stability.

\ifthenelse{\boolean{includeFigures}}{
\begin{figure*}[!t]
    \includegraphics[width=\linewidth]{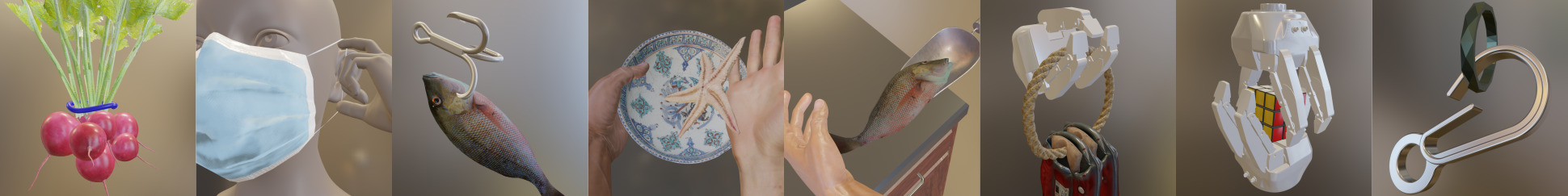}
    \vspace{-5pt}
    \caption{Soft fixture scenarios: (1) radishes held together with a rubber band, (2) putting on a mask, (3) hooking a frozen fish with a fishing hook, (4) catching a starfish with a bowl, (5) scooping of a fish with a shovel, (6) grasping a chain of transfusion bags, (7) manipulating a Rubik cube with two grippers, (8) snap lock mechanisms. The examples include textures and meshes by \cite{mesh1,mesh2,mesh3,mesh4,mesh5,mesh6,mesh7,mesh8,mesh9,mesh10,mesh11,mesh12,mesh13,mesh14,mesh15,mesh16,mesh17, robotiq}.}
    \vspace{-2pt}
    \label{fig-scenarios}
\end{figure*}
}{}

\section{Related Work}
\paragraph*{Caging Grasps}
The original notion of caging, introduced by Kuperberg \cite{kuperberg1990problems}, addresses the problem of preventing a polygon from escaping arbitrarily far away using a set of fixed point obstacles, ensuring that its path-component in the collision-free C-space remains bounded. Building on this idea, Rimon et al.\cite{rimon1996caging, rimon1999caging} extended caging to robotic grasping and developed a theory for caging-based grasps for planar objects. In \cite{allen2015two, allen2014two}, methods for finding two-finger cage formations of planar polygons and 3D polyhedra by means of contact-space search were determined. Additionally, Rodriguez et al. \cite{rodriguez2012caging} demonstrated that caging grasps can be regarded as a viable means of achieving immobilizing grasps via finger closure. Subsequent works, such as those by \cite{pokorny2013grasping, varava2016caging}, have applied caging grasps to certain classes of 3D rigid and deformable objects, analyzing topological and geometric features. However, these methods are limited to objects with specific features, such as holes or double forks. Makapunyo et al. \cite{makapunyo2012measurement} pioneered the concept of partial caging, characterizing it as a finger arrangement allowing limited escape movements. Satoshi et al. \cite{makita2015evaluation} further explored partial caging concerning circular objects within a planar hand. Varava et al. \cite{welle2021partial} defined a clearance-based quality measure and employed deep neural networks to generate a 2D partial caging dataset. Gravity caging is proposed in \cite{jiang2012learning} to derive placing areas where objects could be partially caged and held under gravitational force. Mahler et al. \cite{mahler2016energy, mahler2018synthesis} introduced energy-bounded caging of 2D rigid objects, extending the caging paradigm with gravitational potential energy constraints through cell-based decomposition of C-space and the application of persistent homology. Advancing along this trajectory, we show that it is possible to incorporate generalized potential energies and complex 3D rigid and deformable objects with 6-18 dimensional C-spaces using the sampling-based approach we propose here.


\paragraph*{Sampling-based Planning for Manipulation}
Sampling-based motion planning methods encompass methods like PRMs\cite{kavraki1996probabilistic} and RRTs \cite{lavalle2001randomized}, alongside asymptotically optimal approaches such as RRT* and PRM* \cite{karaman2011sampling}. Notably, Batch Informed Trees (BIT*) \cite{gammell2015batch} strikes a balance between graph-search and sampling-based techniques, prioritizing search by potential solution quality akin to A*, while maintaining asymptotic optimality like RRT* \cite{gammell2020batch}. Building upon BIT*, our method optimizes a customized cost function that accelerates finding a solution minimizing the maximal value along a feasible path. A similar problem was defined in \cite{solovey2017efficient} and solved with a bottleneck tree. Manipulation problems are often approached using sampling-based motion planners\cite{gammell2021asymptotically}. For instance, Schmitt et al. \cite{schmitt2017optimal} address manipulation tasks as a multi-modal planning problem in high-dimensional C-space, utilizing a PRM*-based approach for offline manipulation planning. Shome et al. \cite{shome2017improving} enhance computational scalability for dual-arm systems without explicitly constructing a roadmap in the composite space. CBiRRT \cite{berenson2009manipulation, simeon2004manipulation} explores constraint manifolds in manipulation and automates motion primitive generation by exploring lower-dimensional C-space manifolds \cite{cheng2021contact}.

\section{Preliminaries and Definitions}
In the following, we introduce a notion of potential energy for rigid and deformable objects that we shall study in this work. We then introduce the problem of estimating the minimum energy required for a bounded 3D object $\mathcal{O}\subset\mathbb{R}^3$ to escape from its initial configuration subject to fixed environmental constraints, including support surfaces of finite size, such as a table-top or static robot grippers and in-hand tools. 

\paragraph*{Articulated Objects with Potential Energy}
Three types of objects are considered in our problem setting: (1) 3D rigid objects; (2) 3D deformable objects like fish that yield when subjected to external forces and absorb or release energy through a spring-like mechanism; (3) linear deformable objects, such as ropes and elastic bands. With the exception of elastic bands, we model all types as links or point masses interconnected by compliant revolute joints. We denote by $\mathcal{X}\subset \mathrm{SE}(3)\times \mathbb{R}^{n_j}$ the C-space of an 3D object with $n_l$ links connected by $n_j$ revolute joints, each angle constrained to $[a_i,b_i] \subset [-\pi,\pi]$ $(1 \le i \le n_j)$, with a special rigid object case where $n_j=0$. An element $\boldsymbol{x}\in\mathcal{X}$, described as $\boldsymbol{x}=(\boldsymbol{r}, \boldsymbol{q}, \boldsymbol{\alpha})$, encompasses the position $\boldsymbol{r}$ and orientation (a unit quaternion) $\boldsymbol{q}$ of the center of mass of the base link of the object and a vector of revolute joint angles $\boldsymbol{\alpha}$. The collision space $\mathcal{X}_{\text{obs}}$ indicates a set of configurations for which $\mathcal{O}$ penetrates at least one of the rigid obstacles in the workspace.
The free C-space is given by $\mathcal{X}_{\text{free}} = cl(\mathcal{X} \setminus \mathcal{X}_{\text{obs}})$, where $cl(\cdot)$ denotes the closure of a set. We assume that torsion springs are affixed at each of the revolute joints, resulting in non-zero elastic potential energy for joint configurations other than those with $\boldsymbol{\alpha}=\boldsymbol{0}$. We consider the following energy function $E:\mathcal{X}\to \mathbb{R}\cup\{\infty\}$, combining gravitational and elastic potential energy \cite{young2020university}:
\begin{equation} \label{eq1}
E(\boldsymbol{x}) = 
\sum_{i=1}^{n_l} m_i g z_i +\sum_{j=1}^{n_j} \frac{1}{2} k_j {\alpha}_{j}^2, \quad \boldsymbol{x}\in \mathcal{X}_{\text{free}}.
\end{equation}
Here, $g$ denotes the gravitational acceleration constant, and $z_i$ represents the height of the center of mass of the $i$'th link in the world coordinates (gravity in $-z$ direction). ${\alpha}_{j}$, $k_j$ denote the angle and the stiffness coefficient of the $j$'th joint ($k_j=0$ for non-elastic ropes), respectively. Configurations in collision space $\mathcal{X}_{\text{obs}}$ have infinite energy.

For a closed loop linear elastic deformable object, such as a rubber band, we select $n_b$ equally spaced key points along it and assume these points are loosely interconnected by non-compressible tension springs. Its C-space is approximated by $\mathcal{X}\subset \mathbb{R}^{3n_b}$ with an element $\boldsymbol{x}=(\boldsymbol{r}_1, \boldsymbol{r}_2, ..., \boldsymbol{r}_{n_b})$. 
The energy function is given by
\begin{equation} \label{eq2}
E_b(\boldsymbol{x}) = 
\sum_{i=1}^{n_b} \frac{1}{2} k_i (\max\{\|{\boldsymbol{r}_i}-{\boldsymbol{r}_{i-1}}\|-L, 0\})^2,
\end{equation}
where $\boldsymbol{x} \in \mathcal{X}_{\text{free}}$, $L$ is the rest length of the springs,
${\boldsymbol{r}_0}$ and ${\boldsymbol{r}_{n_b}}$ indicate identical key points along the enclosed loop.
For this scenario, we do not model gravitational effects and mass is not explicitly modeled in the
energy function.
\ifthenelse{\boolean{includeFigures}}{
\begin{figure}
    \includegraphics[width=\linewidth]{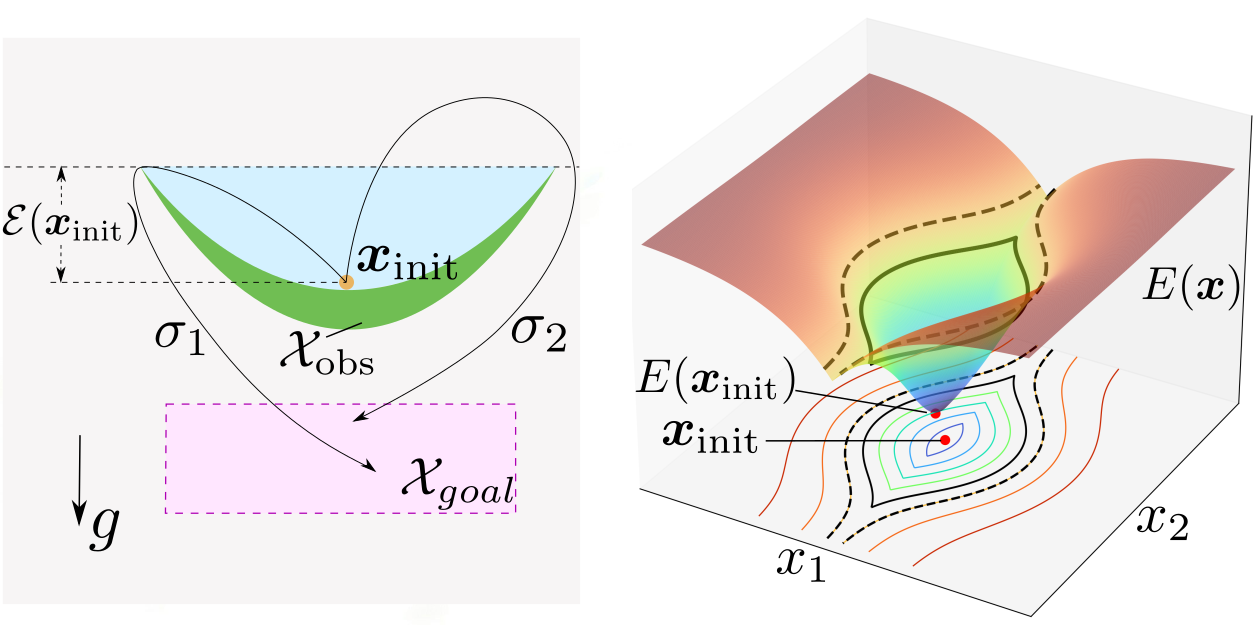}
    \caption{Left: escape energy $\mathcal{E}(\boldsymbol{x}_{\text{init}})$ of a point-mass object subject to gravity inside a bowl (green) in a 2D workspace. Right: a synthetic energy function $E(\boldsymbol{x})$ on a 2D C-space with its contour map to illustrate the concept of a soft fixture. We display the relative increase in energy as relative iso-lines below the energy surface. Note that the two subplots illustrate two different scenarios.}
    \vspace{-5pt}
    \label{fig-illu}
\end{figure}
}{}

\paragraph*{Caging and Soft Fixtures}
We introduce the concept of a soft fixture as an extension of caging to include both
rigid and deformable objects subject to potential energy defined above. Let us first recall the
classical notion of a cage.

\begin{definition}
An object in configuration $\boldsymbol{x}_{\text{init}}\in \mathcal{X}_{\text{free}}$ is \emph{caged} if $\boldsymbol{x}_{\text{init}}$ lies in a bounded path-component of $\mathcal{X}_{\text{free}}$. 
\label{def1}
\end{definition}

A cage thus restricts the movement of an object from escaping arbitrarily far away from its initial configuration.

\begin{definition}
We consider an object in configuration $\boldsymbol{x}_{\text{init}}\in\mathcal{X}_{\text{free}}$ to be in a \emph{soft fixture} configuration with respect to an energy function $E$ if, for some $e\ge 0$, there exists a bounded path-component containing $\boldsymbol{x}_{\text{init}}$ within the sublevel set
\begin{equation} \label{eq3}
\mathcal{X}_{e}(\boldsymbol{x}_{\text{init}}) = \{ \boldsymbol{x} \in \mathcal{X}_{\text{free}} : E(\boldsymbol{x})\le E(\boldsymbol{x}_{\text{init}})+e \}.
\end{equation}
We call the supremum of $e$, such that the condition above is satisfied, the \emph{escape energy} $\mathcal{E}(\boldsymbol{x}_{\text{init}})$ of the fixture.
\label{def2}
\end{definition}

The notion of a soft fixture provides a generalized form of energy-bounded cages introduced for 2D polygonal objects~\cite{mahler2018synthesis}. A simple real-world soft fixture example is that of an object that lies inside a deep bowl under gravity (Fig. \ref{fig-illu}, left). Escape energy $\mathcal{E}(\boldsymbol{x}_{\text{init}})$ subject to potential energy then intuitively corresponds to determining the minimal lift of the object required to remove it fully from the bowl (along path $\sigma_1$). Note that the light blue region corresponds to the largest bounded path-component containing $\boldsymbol{x}_{\text{init}}$, confined by an energy sublevel set.
The right of Fig. \ref{fig-illu} further illustrates the concept with a synthetic 2D C-space and energy function. Consider an $\boldsymbol{x}_{\text{init}}$ corresponding to the minimum of the displayed energy function. As the energy threshold increases, the sublevel set $\mathcal{X}_{e_1}(\boldsymbol{x}_{\text{init}})$ (denoted by solid lines) is still bounded, but for
$\mathcal{X}_{e_2}(\boldsymbol{x}_{\text{init}})$ (dashed lines), the sublevel set becomes unbounded along the $x_2$ axis direction. $e_1$, the supremum of $e$ providing boundedness, thus corresponds to the escape energy in this scenario.  
With respect to our definition of the energy function above, also note that a soft fixture with infinite escape energy corresponds to the classical concept of a cage as in Def. \ref{def1}.

\paragraph*{Problem Statement}
The central challenge addressed in this paper is the following:
Given a 3D rigid or deformable object in a soft fixture configuration $\boldsymbol{x}_{\text{init}}$, which is modeled as above and subject to potential energy as in Eq. \ref{eq1} or \ref{eq2}, determine the escape energy $\mathcal{E}(\boldsymbol{x}_{\text{init}})$ under fixed environmental constraints. While several efforts have tackled energy-bounded cages of 2D rigid objects, the analysis of fixturing 3D deformable objects under energy and environmental constraints remains unaddressed.

\paragraph*{Approximating Escape Energy with Escape Paths}
\begin{definition}
For any feasible path $\sigma:[0, 1]\to \mathcal{X}_{\text{free}}$, we denote 
the \emph{energy cost} along $\sigma$ by
\begin{equation} \label{eq4}
C(\sigma)= -E(\sigma(0)) + \max_{t\in[0, 1]} E(\sigma(t)).
\end{equation}
\label{def3}
\vspace{-15pt}
\end{definition}
For a given $\boldsymbol{x}_{\text{init}}$, and feasable path $\sigma$ leading arbitrarily far away from $\boldsymbol{x}_{\text{init}}$, $C(\sigma)$ establishes an upper bound of $\mathcal{E}(\boldsymbol{x}_{\text{init}})$. We are thus inspired to attempt to find ``best escape paths'' that lead infinitely far away from $\boldsymbol{x}_{\text{init}}$ but which encounter the lowest maximal potential energy values along the way to find tight upper bounds of escape energy.  If no such escape path can be determined (for example due to classical caging by obstacles), no finite upper bound estimate can be established (indicated by an upper bound of $\infty$ in experiments).
Since we are working with bounded obstacles and due to the form of our energy function, we furthermore simplify the problem to consider only escape paths from $\boldsymbol{x}_{\text{init}}$ to some bounded goal region $\mathcal{X}_{\text{goal}}$ that is sufficiently far away from obstacles and has lower energy values than
$\boldsymbol{x}_{\text{init}}$. Once any path reaches this goal region, we can extend the path to lead infinitely far away from the initial configuration by simply reducing the configuration's z-coordinate indefinitely while keeping the rest of the rigid or deformable object state static. This ensures that the supremum of energy values over the infinite extended path is the same as that over the initial bounded path from $\boldsymbol{x}_{\text{init}}$ to $\mathcal{X}_{\text{goal}}$ and makes the problem more amenable to standard motion planning algorithms expecting a finite goal region. The key question we address next is how to determine suitable candidate escape paths that minimize $C(\sigma)$ to provide suitable estimates for $\mathcal{E}(\boldsymbol{x}_{\text{init}})$. The left of Fig. \ref{fig-illu} provides an illustration in 2D under gravitational potential energy with a sub-optimal escape path $\sigma_2$ and another escape path $\sigma_1$ that provides a tight upper bound estimate.



\section{Escape Energy Estimation}
\paragraph*{Iterative Search}
Our first set of algorithms for approximating soft fixture escape energy focuses on searching for feasible escape paths from $\boldsymbol{x}_{\text{{init}}}$ to
$\mathcal{X}_{\text{{goal}}}$ subject to being constrainted to within sublevel sets $\mathcal{X}_e(\boldsymbol{x}_{\text{{init}}})$ with increasingly lower threshold $e$.
Specifically, we run an RRT planner constrained to $\mathcal{X}_e(\boldsymbol{x}_{\text{{init}}})$, initiating at $\boldsymbol{x}_{\text{{init}}}$, with the goal to reach $\mathcal{X}_{\text{{goal}}}$, positioned sufficiently distant from obstacles. In our experimental setup, we designate a cubic region $\mathcal{H} \subset \mathbb{R}^3$, situated lower than the workspace obstacles in the direction of gravity. A path is considered to have adequately escaped from an initial configuration $\boldsymbol{x}_{\text{{init}}}$ if it achieves a configuration $\boldsymbol{x}_{\text{{goal}}} = (\boldsymbol{r}_g, \boldsymbol{q}_g, \boldsymbol{\alpha}_g)$, where $\boldsymbol{r}_g \in \mathcal{H}$, $\boldsymbol{q}_g$ is arbitrary, and $\boldsymbol{\alpha}_g = 0$ (in the general formulation). This approach ensures a goal space consisting of configurations with reduced total potential energy compared to $\boldsymbol{x}_{\text{{init}}}$.

To approximate the upper bounds of escape energy, we first employ a simple conservative search (Algo. \ref{algo1}) as a baseline. In each iteration, we execute an RRT planner within a finite time budget to find a trajectory entirely within $\mathcal{X}_{\hat{e}}(\boldsymbol{x}_{\text{{init}}})$. This search initially sets $\hat{e} \gg 0$, begins at $\boldsymbol{x}_{\text{{init}}}$ and connects to $\mathcal{X}_{\text{{goal}}}$ (Line 4). If a feasible path $\sigma$ is found, it determines an upper bound $\hat{e} = C(\sigma)$ for escape energy, and the search recommences within the reduced sublevel set $\mathcal{X}_{\hat{e}}(\boldsymbol{x}_{\text{{init}}})$ (Lines 5-6). Each discovered path successively refines this upper bound in practice. If no path is located, the search repeats within the previous region. The process concludes when $\hat{e} \approx 0$, or upon reaching a preset maximum iteration or time limit for the motion planner, thus defining an upper bound $\hat{e}$ for escape energy $\mathcal{E}(\boldsymbol{x}_{\text{init}})$.

\begin{algorithm}[tp]
\text{Initialize}, $\hat{e} \gets \infty$, $t \gets 0$\\
\text{Set stopping criteria}, $t_{\text{max}}$\\

\Repeat{$t \ge t_{\textup{max}}$ or $\hat{e} \approx 0$}
{
$\Delta t, \sigma \gets \textsc{RRTSearch} (\mathcal{X}_{\hat{e}}(\boldsymbol{x}_{\text{init}}))$ \\ 
\If{$\sigma$ exists}{
    $\hat{e} \gets C(\sigma)$ \\
    }
$t \gets t + \Delta t$ \\
} 
\textbf{Return} \textit{{$\hat{e}$, $\sigma$}}
\caption{Conservative Search}
\label{algo1}
\end{algorithm}

We have also implemented a binary search approach (see Algo. \ref{algo2}), conceptually similar to the bisection method described in \cite{welle2021partial}.
This binary search adheres to a divide-and-conquer principle to locate the escape energy, maintaining both its lower bound $e_{-}$ and upper bounds $e_{+}$, while progressively narrowing the range. In each iteration, we execute RRT within the sublevel set $\mathcal{X}_{\hat{e}}(\boldsymbol{x}_{\text{{init}}})$ with $\hat{e}$ given by the average of the current upper and lower bounds (Lines 4-5). If a path is found, the upper bound for escape energy is updated (Lines 6-7), similar to Algo. \ref{algo1}. Otherwise, the lower bound is updated with $\hat{e}$ (Lines 11-12).
It is worth noting that in practice, $e_-$ does not necessarily represent a strict lower bound for escape energy. Given the finite time budget allocated to RRT, failure to find a path may only imply a high probability that the range does not contain the target. Consequently, if energy cost $C(\sigma)$ lower than the lower bound $e_-$ is discovered, we reset the lower bound to $0$ (Lines 8-9).

\begin{algorithm}[h]
\text{Initialize}, $\hat{e}\gets \infty$, $t \gets 0$, $e_{-} \gets 0$, $e_{+}$\\
\text{Set stopping criteria}, $t_{\text{max}}$, $e_{\epsilon}$\\
\Repeat{$t \ge t_{\textup{max}}$ or $e_+\approx0$ or $e_{+}-e_{-} < e_{\epsilon}$}
{
$\hat{e} = (e_{+} + e_{-}) / 2$ \\
$\Delta t, \sigma \gets \textsc{RRTSearch} (\mathcal{X}_{\hat{e}}(\boldsymbol{x}_{\text{init}}))$  \\
\If{$\sigma$ exists}{
    $e_{+} \gets C(\sigma)$ \\
    \If{$C(\sigma) < e_{-}$}{
        $e_{-} \gets 0$ \\
        }
    }
\Else{
    $e_{-} \gets \hat{e}$ \\
    }
$t \gets t + \Delta t$ \\
} 
\textbf{Return} \textit{{$e_{+}$, $\sigma$}}
\caption{Binary Search}
\label{algo2}
\end{algorithm}
\vspace{-3pt}

\ifthenelse{\boolean{includeFigures}}{
\begin{figure*}
    \centering
    \input{icra2024/tex/quantitative-tests/sum}
    \vspace{-2pt}
    \caption{Quasi-static analysis of escape energy in two scenarios with 27 and 1 frame, respectively. (i): a ring dangling on a hook and swinging back and forth (i-1, A, C), and an elastic band trapped by a conical frustum (i-2). (ii): estimated escape energy of the three algorithms w.r.t a reference as the ring swings downward (dashed lines A, C corresponding to configurations in (i-1)). (iii): convergence of the algorithms at frame B of (ii). (iv): escape energy of an elastic band in (i-2) with varying numbers of modeling control points and reference escape energy (green line) corresponding to ring configuration $\boldsymbol{x}^{*}_{max}$ in (i-2). }
    \label{fig-quan}
\end{figure*}
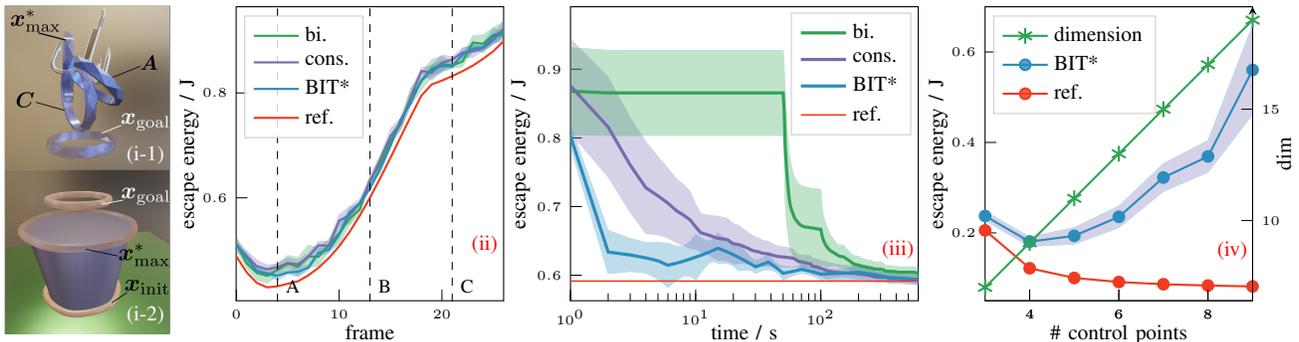
}{}

\ifthenelse{\boolean{includeFigures}}{
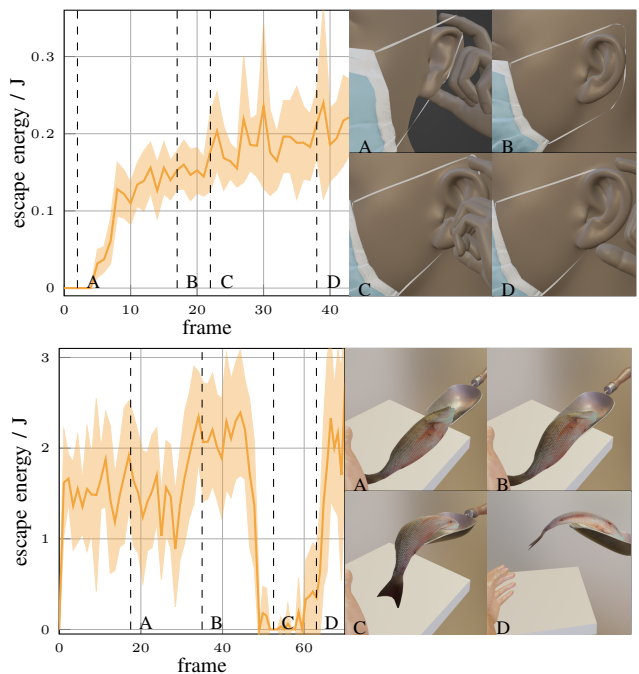
\begin{figure}
    \centering
    \input{icra2024/tex/band-ear-mosaic}
    \input{icra2024/tex/fish-shovel-mosaic}
    \vspace{-2pt}
    \caption{Quasi-static soft fixture analysis of the mask-wearing and fish-scooping scenarios in 44 and 71 frames, respectively. We run 5-12 independent runs of a 5-minute execution of the BIT* planner for each frame, with 4 illustrated frames. }
    \label{fig-qual}
\end{figure}
}{}

\paragraph*{Energy-biased Optimality Search}
\label{sect-rrt*}
Building upon the intuitive algorithms previously discussed, we introduce a configuration-cost function to prioritize states with lower potential energy within the framework of a standard RRT* algorithm\footnote{A full algorithm is on the project website we provided.}.
Let $\mathcal{X}_{v} \subset \mathcal{X}_{\text{{free}}}$ denote the set of vertices in the RRT* graph. We define a cost-to-come function $c:\mathcal{X}_{v} \rightarrow \mathbb{R}_{\geq 0}$, mapping each vertex in $\mathcal{X}_{v}$ to the cost of the unique path from the root of the tree to that vertex. This function is employed during vertex expansion and rewiring, following the standard sample-and-steer approach typical in RRT*-like methods.
The relationship between a child vertex $\boldsymbol{x}_{\text{{ch}}}$ and its parent vertex $\boldsymbol{x}_{\text{{pa}}}$ is captured by the recurrence relation of the cost functions,
\begin{equation} \label{eq6}
c(\boldsymbol{x}_{\text{{ch}}}) = \max \{c(\boldsymbol{x}_{\text{{pa}}}), E(\boldsymbol{x}_{\text{{ch}}}) - E(\boldsymbol{x}_{\text{{init}}})\},
\end{equation}
with $c(\boldsymbol{x}_{\text{{init}}})=0$.
Taking vertex expansion as an example, we define the cost-to-come $c(\boldsymbol{x}_{\text{new}})$ of a new sample $\boldsymbol{x}_{\text{new}} \in \mathcal{X}_{\text{free}}$ as 
$c(\boldsymbol{x}_{\text{new}}) = \max \{c(\boldsymbol{x}_{\text{near}}), E(\boldsymbol{x}_{\text{new}}) - E(\boldsymbol{x}_{\text{init}})\}$, if it connects to a nearby vertex $\boldsymbol{x}_{\text{near}} \in \mathcal{X}_{v}$ already in the graph.
This expression for the cost of $\boldsymbol{x}_{\text{new}}$ represents the energy cost from $\boldsymbol{x}_{\text{init}}$ to $\boldsymbol{x}_{\text{new}}$ via $\boldsymbol{x}_{\text{near}}$. 
The cost function steers exploration to expand vertices and rewire edges based on their potential energy-informed cost-to-come. This approach aims to prune branches of the search tree that are unlikely to lead to an escape path with minimal energy. Other asymptotically optimal algorithms besides RRT*, such as BIT*, Informed RRT*, AIT*, etc., share the cost function formulation. We specifically employ BIT* in scenarios, leveraging its advantages of anytime and incremental techniques and increasing search density by reusing information. A cost-to-go heuristic is not defined for BIT* with this particular cost function. 

\section{Evaluation}
This section provides an evaluation of the algorithms proposed for quasi-static soft fixture analysis. The examination includes a variety of manipulation primitives depicted in Fig. \ref{fig-scenarios} and the accompanying video, with two specific examples illustrated in Fig. \ref{fig-qual}.

In these manually created simulation sequences, objects are subject to gravitational forces and their mobility is constrained by various obstacles, such as grippers, hands, in-hand tools, etc. The configurations of objects are recorded for each frame of the simulation during this process. Subsequently, we analyze the configuration of each frame using our proposed methods in a quasi-static manner and approximate escape energy per frame to evaluate our methods. Note that, while kinetic energy is disregarded in the per-frame soft fixture analysis here, it contributes to the simulation of the manipulation primitive sequence.

We utilized the Open Motion Planning Library (OMPL) \cite{sucan2012open} to implement the proposed motion planners. Pybullet~\cite{coumans2016pybullet} served as the platform for collision detection, while forward simulation of bodies under gravity was jointly conducted in Pybullet and Blender. All experiments are performed on an Intel Core i9-12900H processor with 14 cores and speeds up to 5.0GHz. 

\ifthenelse{\boolean{includeFigures}}{
\begin{figure*}
    \centering
    \input{icra2024/tex/phys-experiment/side-by-side-plots}
    \vspace{-5pt}
    \caption{Physical experiment setup and results. (i-1): the arm's initial configuration.
    (i-2): the replication of (i-1) in simulation.
    (i-3): 3 bowl-like caging tools of various depths, objects (tennis ball, tape, glue stick, banana) for (ii) and toroidal objects with a section cut off, forming angles of $ [45^\circ,60^\circ,90^\circ,120^\circ]$ for (iii).
    The objects are respectively placed inside one of the bowls as in (i-1).
    (ii): observed perturbation level at which objects escape from bowls and estimated normalized escape energy obtained using BIT* with simple objects (Square, triangle and diamond markers denote small, medium and large bowls, respectively). 8 runs in simulation and physical experiments respectively for each data point. Ellipses showcasing standard deviation. (iii): the perturbation level in terms of different tori.
    }
    \label{fig-phys}
 \end{figure*}
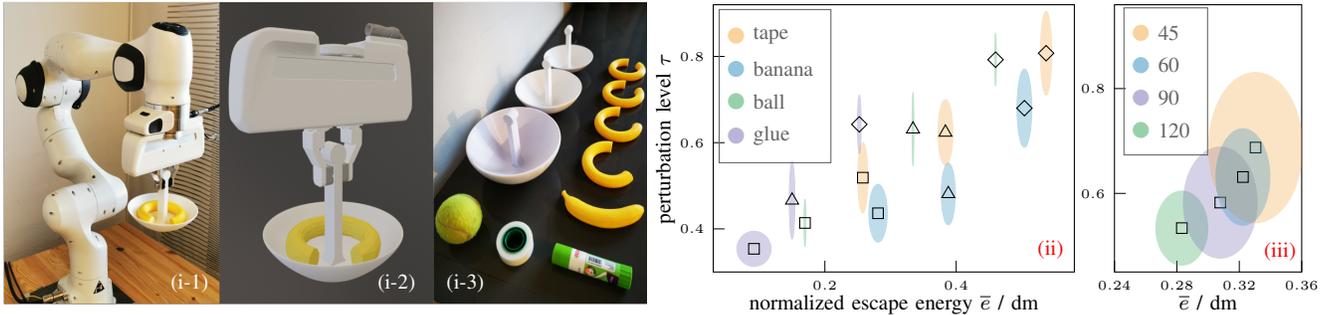
}{}

\paragraph*{Quantitative Analysis}
We conducted an analysis to verify both the accuracy and efficiency of our proposed approaches, with a focus on a rigid ring scenario (Fig. \ref{fig-quan}, i-1). Intuitively we believe an optimal escape path of the ring 
attains maximal energy at configurations where it is just barely lifted above the fish hook to escape with the ring in a vertical plan, as $\boldsymbol{x}_{\text{max}}^*$ illustrates. We display the resulting escape energy of the optimal escape path via $\boldsymbol{x}_{\text{max}}^*$ for each frame of the simulation as a reference curve in part (ii) of the figure.
As can be seen in (ii), BIT* demonstrated a similar level of performance to the iterative methods across the frames of the simulation, all of which were closely aligned with the empirical reference escape energy. 
Further observation revealed an enhanced convergence speed for BIT* over the other two methods within a 10-minute runtime (Fig. \ref{fig-quan}, iii). 

Convergence in binary search is contingent on the location of $e^*$ within the interval of the initial energy bounds $[e_-,e_+]$. When $e^*$ is closer to $e_-$, the algorithm converges more rapidly, as the target resides within the search range in the initial iterations.
A comparative analysis of different motion planning algorithms for frame B in the hook scenario in Fig. \ref{fig-quan} under our energy cost function was conducted in Table \ref{tab}. The anytime planners, BIT* and AIT*, exhibited superior performance in this specific problem relative to RRT*-based planners.
Moreover, we assessed the accuracy concerning state-space dimensionality within a simple scenario: an elastic hair tie escaping from a human wrist (modeled by a conical frustum) as depicted in Fig. \ref{fig-quan} (i-2). The observed deviation (Fig. \ref{fig-quan} iv) between the estimated escape energy and the expected reference escape energy corresponding to $\boldsymbol{x}_{\text{max}}^*$ displayed in the figure escalated with higher dimensions and a quadratic energy function, as articulated in Eq. \ref{eq2}, illustrating the expected curse of dimensionality effect affecting sampling-based motion planners in general.

\begin{table}
\centering
\caption{deviation from reference escape energy}
\begin{tabular}{|l|l|l|l|l|l|l|}
\hline
planners        & BIT* & bi. & cons. & AIT*  & inf. RRT* & RRT*  \\
\hline
error mean & 4.02 & 4.53   & 5.54   & 10.33 & 15.43     & 18.71 \\
\hline
error std. & 0.61 & 1.01   & 0.62   & 0.90  & 2.53      & 1.97 \\
\hline
\end{tabular}
\label{tab}
\end{table}

\paragraph*{Qualitative Evaluation}

In the mask example (Fig. \ref{fig-qual}), we focus on a motion primitive where a person manipulates the mask strap by securing it around the ear with two fingers. The mask strap around the ear is represented as a spring-loaded open loop, consisting of 8 control points—two fixed at the filter's boundary, with the remaining control points being freely movable (comprising 18 degrees of freedom). We consider elasticity but disregard the mass of the mask components, and assume the wearer is holding the central part of the mask with another hand until the deformation energy of the strap is zero. Consequently, we model a goal state for soft fixture analysis corresponding to a configuration in $\mathcal{X}_{\text{free}}$ with zero elastic potential energy from which the mask can be removed arbitrarily far from the face without increasing deformation energy. Note that the illustrated hand is not considered an obstacle during the soft fixture analysis, as we are primarily examining the interaction between the ear and the mask loop. Our initial estimated escape energy is zero, as the loop has not yet engaged the auricle (subfigure A) and a mere contraction offers a viable escape path. However, as the loop progresses behind the ear (B), begins to contract (C), and is snugly fits around the back of the ear (D), an increased escape energy is estimated, indicating the stability of this configuration. 

In a fish scooping scenario, we approximate an articulated fish model with $n_l = 10, n_j = 9$ (Fig. \ref{fig-qual}) and consider primarily its bending deformation, following the natural curvature of the spine. Initially, a hand, tabletop and shovel collectively act as obstacles, constraining the fish's movement during scooping, and thus the fish needs to lift and increase the gravitational component of its potential energy slightly to escape (A, B). As it settles into the shovel, the shovel lifts from the desk (C). The fish's elastic potential energy and the shovel's mouth opening facilitate an effortless leftward and downward slide. By rotating outward (D), the shovel incline angle is changed, resulting in an increase in estimated escape energy corresponding to the intuition that the fish requires an addition of energy to escape this configuration.

\paragraph*{Physical Experiment}
Here we conducted evaluation experiments using a Franka Emika Panda robot arm \cite{haddadin2022franka}. 
Bowl-like tools with three varying depths were attached to the end-effector of the arm (Fig. \ref{fig-phys}, i-1) and we positioned four different objects of various shapes within the bowls to create soft fixture candidates. 
Increasing levels of perturbation motions, denoted as $\tau$, were applied to simulate increasing external disturbance to the initial configuration of the arm. The tool, influenced by this perturbation, rotates and translates randomly, resulting in the object's release upon reaching a critical perturbation level $\tau$ that was recorded. A positive correlation between perturbation level and mass-normalized escape energy $\overline{e} = \hat{e} / (mg)$, was observed. Here, $m$ denotes the object's mass. Normalized escape energy is computed by replicating the real-world geometry and poses in simulation (Fig. \ref{fig-phys}, i-2) and employing our upper bound escape energy estimation approach. 

In a second experiment, we examined toroidal objects with cutouts (Fig. \ref{fig-phys}, i) to explore the empirical escape perturbation levels in relation to normalized escape energy, specifically with narrow C-space tunnels. Tori with a smaller cutout demonstrated resistance to higher perturbation levels and exhibited slower convergence in escape energy estimation, as is to be expected as this behavior stems from the necessity for denser sampling to find a feasible path through the narrow C-space tunnel (Fig. \ref{fig-phys}, iii). The observed exponential trend underscores that $\overline{e}$ only partially characterizes the escape probability in complex scenarios exhibiting low-clearance escape paths since escape energy for all these cut-out objects intuitively is identical and corresponds to a small lift and sidewards escape motion.

\section{Conclusion}
We presented steps towards extending caging-based manipulation to novel scenarios involving both rigid and deformable objects. The energy-function based sublevel set analysis approach investigated here can, we believe, provide a fruitful long-term research direction, with many interesting challenges, such as the creation of algorithms for soft fixture synthesis (rather than analysis) and the inclusion of dynamic rather than quasi-static settings in the analysis. 

\bibliography{main}


\ifthenelse{\boolean{includeAppendix}}{
\clearpage
\section{Appendix}
\subsection{Energy-biased Optimality Search}
We display the complete algorithm of RRT* with a hybrid cost function in Algo. \ref{algo3}, as is discussed in Section \ref{sect-rrt*}.

\begin{algorithm}[h]
$\mathcal{X}_{v} \gets \{\boldsymbol{x}_{\text{init}}\}$; $\text{c}(\boldsymbol{x}_{\text{init}}) \gets 0$; $\mathcal{W} \gets \emptyset$;\\
\For{$i = 1$ to $n$}
{
    $\boldsymbol{x}_{\text{rand}} \gets \textsc{SampleFree}_i$\\
    $\boldsymbol{x}_{\text{nearest}} \gets \textsc{Nearest}(G = (\mathcal{X}_{v}, \mathcal{W}), \boldsymbol{x}_{\text{rand}})$\\
    $\boldsymbol{x}_{\text{new}} \gets \textsc{Steer}(\boldsymbol{x}_{\text{nearest}}, \boldsymbol{x}_{\text{rand}})$\\
    \If{$\textsc{ObstacleFree}(\boldsymbol{x}_{\textup{nearest}}, \boldsymbol{x}_{\textup{new}})$}
    {
        $\mathcal{X}_{\text{near}} \gets \textsc{Near}(G = (\mathcal{X}_{v}, \mathcal{W}), \boldsymbol{x}_{\text{new}}, r(\textsc{card}(V)))$\\
        $\mathcal{X}_{v} \gets \mathcal{X}_{v} \cup \{\boldsymbol{x}_{\text{new}}\}$\\
        $\boldsymbol{x}_{\text{min}} \gets \boldsymbol{x}_{\text{nearest}}$\\
        $c_{e,\text{min}} \gets \max \{c_e(\boldsymbol{x}_{\text{{nearest}}}), E(\boldsymbol{x}_{\text{{new}}}) - E(\boldsymbol{x}_{\text{{init}}})\}$\\
        $c_{l,\text{min}} \gets c_l(\boldsymbol{x}_{\text{{nearest}}}) + \lVert \boldsymbol{x}_{\text{{new}}} - \boldsymbol{x}_{\text{{nearest}}} \rVert $\\
        $c_{\text{min}} \gets  c_{e,\text{min}} + \gamma c_{l,\text{min}}$\\
        \ForEach{$\boldsymbol{x}_{\textup{near}} \in \mathcal{X}_{\textup{near}}$}  
        {    
            $c_{e,\text{que}} \gets \max \{c_e(\boldsymbol{x}_{\text{{near}}}), E(\boldsymbol{x}_{\text{{new}}}) - E(\boldsymbol{x}_{\text{{init}}})\}$ \\
            $c_{l,\text{que}} \gets c_l(\boldsymbol{x}_{\text{{near}}}) + \lVert \boldsymbol{x}_{\text{{new}}} - \boldsymbol{x}_{\text{{near}}} \rVert $\\
            $c_{\text{que}} = c_{e,\text{que}} + \gamma c_{l,\text{que}}$\\
            \If{$\textsc{CollisionFree}(\boldsymbol{x}_{\textup{near}}, \boldsymbol{x}_{\textup{new}}) \wedge c_{\textup{que}} < c_{\textup{min}}$}
            {
                $\boldsymbol{x}_{\text{min}} \gets \boldsymbol{x}_{\text{near}}$\\
                $c_{\text{min}} \gets c_{\text{que}}$\\
            }
        }
        \ForEach{$\boldsymbol{x}_{\textup{near}} \in \mathcal{X}_{\textup{near}}$}
        {
            $c_{e,\text{que}} \gets \max \{c_e(\boldsymbol{x}_{\text{{new}}}), E(\boldsymbol{x}_{\text{{near}}}) - E(\boldsymbol{x}_{\text{{init}}})\}$ \\
            $c_{l,\text{que}} \gets c_l(\boldsymbol{x}_{\text{{new}}}) + \lVert \boldsymbol{x}_{\text{{near}}} - \boldsymbol{x}_{\text{{new}}} \rVert $\\
            $c_{\text{que}} = c_{e,\text{que}} + \gamma c_{l,\text{que}}$\\
            \If{$\textsc{CollisionFree}(\boldsymbol{x}_{\textup{new}}, \boldsymbol{x}_{\textup{near}}) \wedge c_{\textup{que}} < c(\boldsymbol{x}_{\textup{near}})$}
            {
                $\boldsymbol{x}_{\text{pa}} \gets \textsc{Parent}(\boldsymbol{x}_{\text{near}})$\\
                $\mathcal{W} \gets (\mathcal{W} \setminus \{(\boldsymbol{x}_{\text{pa}}, \boldsymbol{x}_{\text{near}})\}) \cup \{(\boldsymbol{x}_{\text{new}}, \boldsymbol{x}_{\text{near}})\}$\\
            }
        }
    }     
}
\textbf{Return} $\mathcal{G} = (\mathcal{X}_{v}, \mathcal{W})$
\caption{RRT* with a Hybrid Cost Function}
\label{algo3}
\end{algorithm}

The algorithm follows the framework defined in \cite{karaman2011sampling}. It first samples a new point $\boldsymbol{x}_{\text{new}}$, finds its nearest neighbor $\boldsymbol{x}_{\text{nearest}}$ in the tree, and steers towards it (Lines 3-5) as in RRT. Instead of directly connecting the two, RRT* expands and rewires nodes to create a minimum-cost path. It establishes a set of nodes $\mathcal{X}_{\text{near}}$ nearby $\boldsymbol{x}_{\text{new}}$ within a radius 
\begin{equation}
r(\textsc{card}(V))) = \min\left\{\gamma_{\text{RRT}^*} \left( \frac{\log(\text{card}(\mathcal{X}_{v}))}{\text{card}(V)} \right)^{\frac{1}{d}}, \eta \right\},
\end{equation}
where $\textsc{card}(\cdot)$ denotes the cardinality of a set, i.e. the number of nodes in the set. $\gamma_{\text{RRT}^*}$ and $\eta$ are tuning hyperparameters so as to guarantee the radius decreases in a reasonable way as the tree grows. $d$ denotes the dimension of the state space. We introduce a hybrid cost function, mixing path energy and path length cost (Lines 10-12), rather than an additive cost function only. Starting from $\boldsymbol{x}_{\text{nearest}}$, the algorithm exhausts $\mathcal{X}_{\text{near}}$ and creates an edge from the vertex in $\mathcal{X}_{\text{near}}$ to $\mathcal{X}_{\text{new}}$ along a path with minimum cost (Lines 13-19). Subsequently, new edges are created from $\mathcal{X}_{\text{new}}$ to $\mathcal{X}_{\text{near}}$, if the current path through $\mathcal{X}_{\text{pa}}$ costs higher than a path through $\mathcal{X}_{\text{new}}$; the suboptimal edge connecting $\mathcal{X}_{\text{pa}}$ and $\mathcal{X}_{\text{near}}$ is removed from the tree in this case, since it is not a part of the minimum-cost path from the root of the tree (Lines 20-26).

Furthermore, we place the definitions of probabilistic completeness and asymptotic optimality in \cite{karaman2011sampling} here to help readers better understand the proof in Section \ref{sect-rrt*}.

\textit{Definition 4}: An algorithm is \emph{probabilistically complete}, if, for any robustly feasible path planning problem \( (\mathcal{X}_{\text{free}}, \boldsymbol{x}_{\text{init}}, \mathcal{X}_{\text{goal}}) \), 
\[
\begin{multlined}
\liminf_{n \rightarrow \infty} \mathbb{P}(\{\exists \boldsymbol{x}_{\text{goal}} \in \mathcal{X}_v^{n} \cap \mathcal{X}_{\text{goal}} |  \\
 \boldsymbol{x}_{\text{init}} \text{ is connected to } \boldsymbol{x}_{\text{goal}} \text{ in } \mathcal{G}^n \} ) = 1,
\end{multlined}
\]
where $\mathcal{G}^n =  (\mathcal{X}_{v}^n, \mathcal{W}^n)$ denotes the tree at step $n$ with a node set $\mathcal{X}_{v}^n$ and an edge set $\mathcal{W}^n$.

\textit{Definition 5}: An algorithm is \emph{asymptotically optimal} if, for any path planning problem \( (\mathcal{X}_{\text{free}}, \boldsymbol{x}_{\text{init}}, \mathcal{X}_{\text{goal}}) \) and cost function $c: \Sigma \rightarrow \mathbb{R}_{\geq 0}$ that admit a robustly optimal solution with finite cost $c^*$,
\[
\mathbb{P}\left(\left\{\limsup _{n \rightarrow \infty} C(\sigma_n^*)=c^*\right\}\right)=1,
\]
where $\sigma_n^*$ denotes the optimal path found by the planner at iteration $n$. Since an asymptotically optimal algorithm is able to find an optimal path given infinite runtime, it clearly satisfies probabilistic completeness as well.

\subsection{Hybrid Cost Function}

In this section, we empirically clarify that $\gamma=0$ is a practical and valid simplification of the original hybrid cost function in Algo. \ref{algo3}.
We define normalized total cost $\overline{C}(\hat{\sigma}^*)$ along $\hat{\sigma}^*$ as follows,
\begin{equation}
\overline{C}(\hat{\sigma}^*) = \frac{C_e(\hat{\sigma}^*) + \gamma C_l(\hat{\sigma}^*)}{e^*},
\end{equation}
with a path length cost
\begin{equation}
{C}_l(\hat{\sigma}^*) = \sup_{\{n \in \mathbb{N}, 0<\tau_0<\tau_1<...<\tau_n =1\}} \sum_{i=1}^{n} \lVert \hat{\sigma}^*(\tau_i) - \hat{\sigma}^*(\tau_{i-1}) \rVert,
\end{equation}
which is in line with the definition of total variation in \cite{karaman2011sampling}. $\hat{\sigma}^*$ denotes the estimated optimal path given by Algo. \ref{algo3}.
For example, in a simple 2D $1 \times 1$ workspace, a point-mass object with $mg=1$ and gravitational potential energy traverses from $(0,0)$ to $(1,0)$ while avoiding obstacles (Fig. \ref{app-varying-gammas}, \ref{app-varying-obstacles}). Note that the negative direction of the $y$ axis aligns with the direction of gravity.

\begin{table*}
\centering
\caption{Technical details of all soft fixture scenarios.\protect\footnotemark}
\begin{tabular}{|l|l|l|l|l|l|l|l|l|}
\hline
scenarios & radish   & mask     & hook     & starfish & shovel   & rope     & cube       & lock  \\
\hline
w.s. dim. & 3        & 3        & 3        & 3        & 3        & 3        & 3          & 2     \\
\hline
\# DoF    & 18       & 18       & 6        & 16       & 15       & 18       & 6          & 4     \\
\hline
rigid     &          &          &\checkmark&          &          &          & \checkmark &       \\
\hline
$E_g$     &          &          &\checkmark&\checkmark&\checkmark&\checkmark&\checkmark  &       \\
\hline
$E_e$     &\checkmark&\checkmark&          &          &\checkmark&          &            &\checkmark \\
\hline
planner   &BIT*      &BIT*      &BIT*      &BIT*      &BIT*      &RRT*      &BIT*        &BIT*      \\
\hline
runtime /s&300       &300       &300       &200       &300       &200       &200         &200    \\
\hline
\# runs   &-         &12        &5         &6         &12        &16        &10          &-      \\
\hline
s.p.b     &1000      &1000      &2000      &2000      &5000      &-         &5000        &1000   \\
\hline
\end{tabular}
\label{tab-scenarios}
\end{table*}
\footnotetext{1. w.s. dim.: workspace dimension. 2. \# DoF: degrees of freedom of a fixtured object. 3. rigid: a rigid fixtured object, deformable or articulated otherwise. 4. $E_g$: gravitational potential energy is considered. 5. $E_e$: elastic potential energy is considered. 6. planner: asymptotically optimal algorithm used. 7. runtime: per iteration per run. 8. \# runs: number of runs per iteration. 9. s.p.b.: samples per batch of BIT*.}

Depending on the mixture parameter $\gamma$, the planner takes different strategies to balance the path length cost $C_l(\hat{\sigma}^*)$ and energy cost $C_e(\hat{\sigma}^*)$, as is illustrated in Fig. \ref{app-varying-gammas}. When $\gamma = 0$, the path goes along the lower border of the free space to minimize $C_e(\hat{\sigma}^*)$. A higher weight on the length cost ($\gamma = 10.0$, Fig. \ref{app-varying-gammas} right) encourages a path taking a shortcut around the top of the middle ellipsoid at the cost of higher $C_e(\hat{\sigma}^*)$. A compromise is made when $\gamma=1$, taking both length and energy costs into consideration.

\ifthenelse{\boolean{includeFigures}}{
\begin{figure}
    \centering
    \hspace{-3pt}
    \input{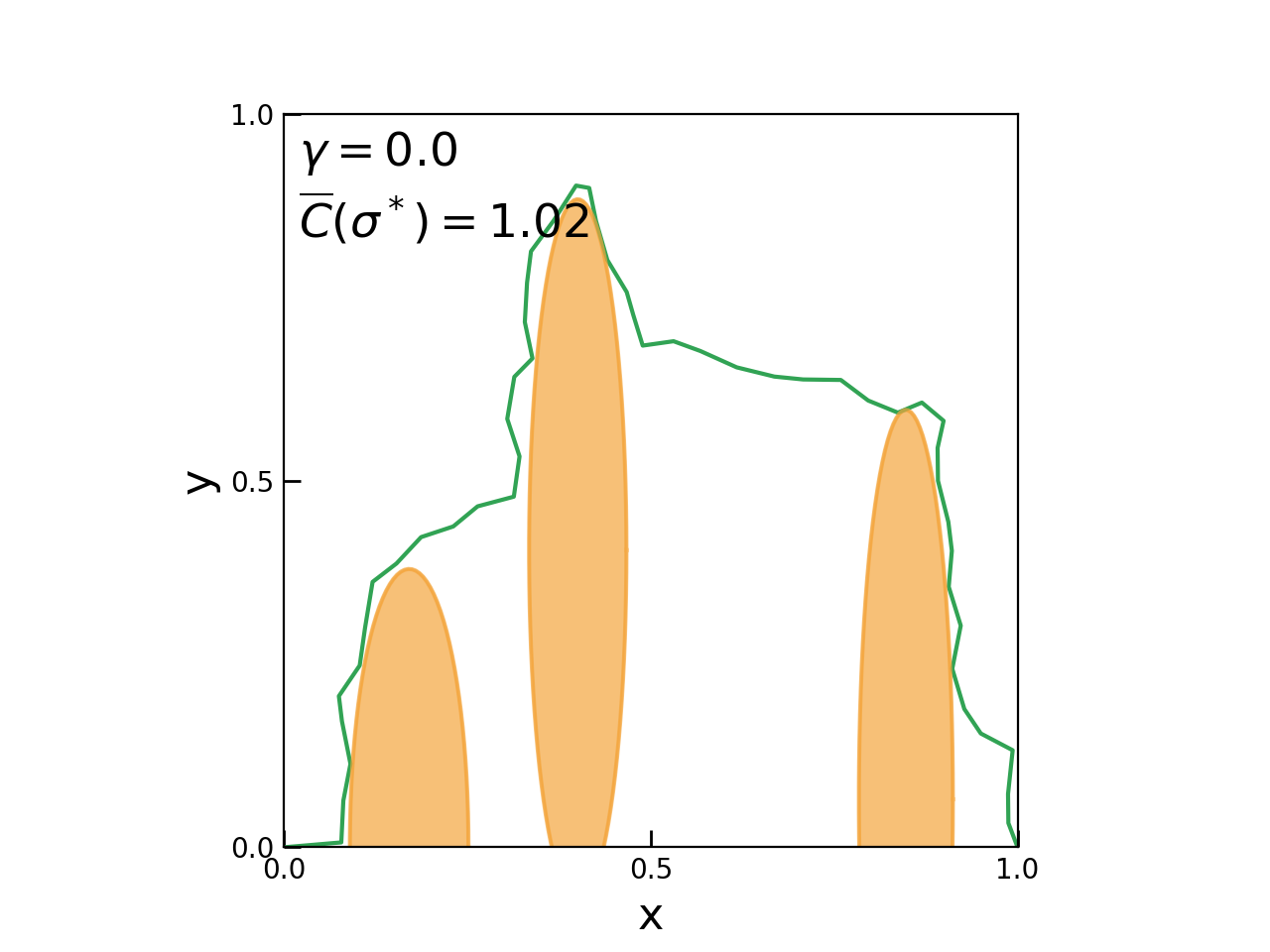}
    \input{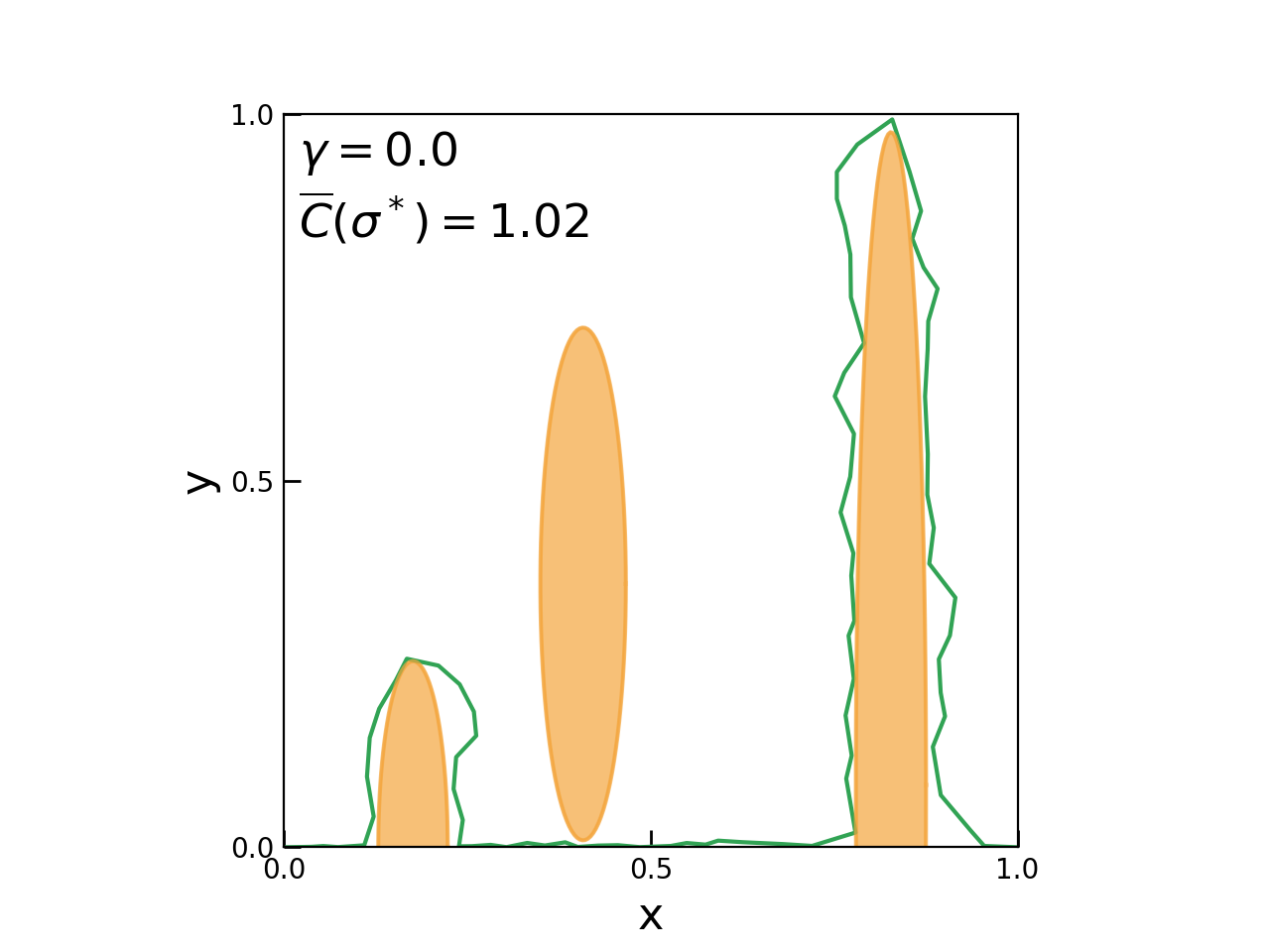}         
    \vspace{5pt}

    \input{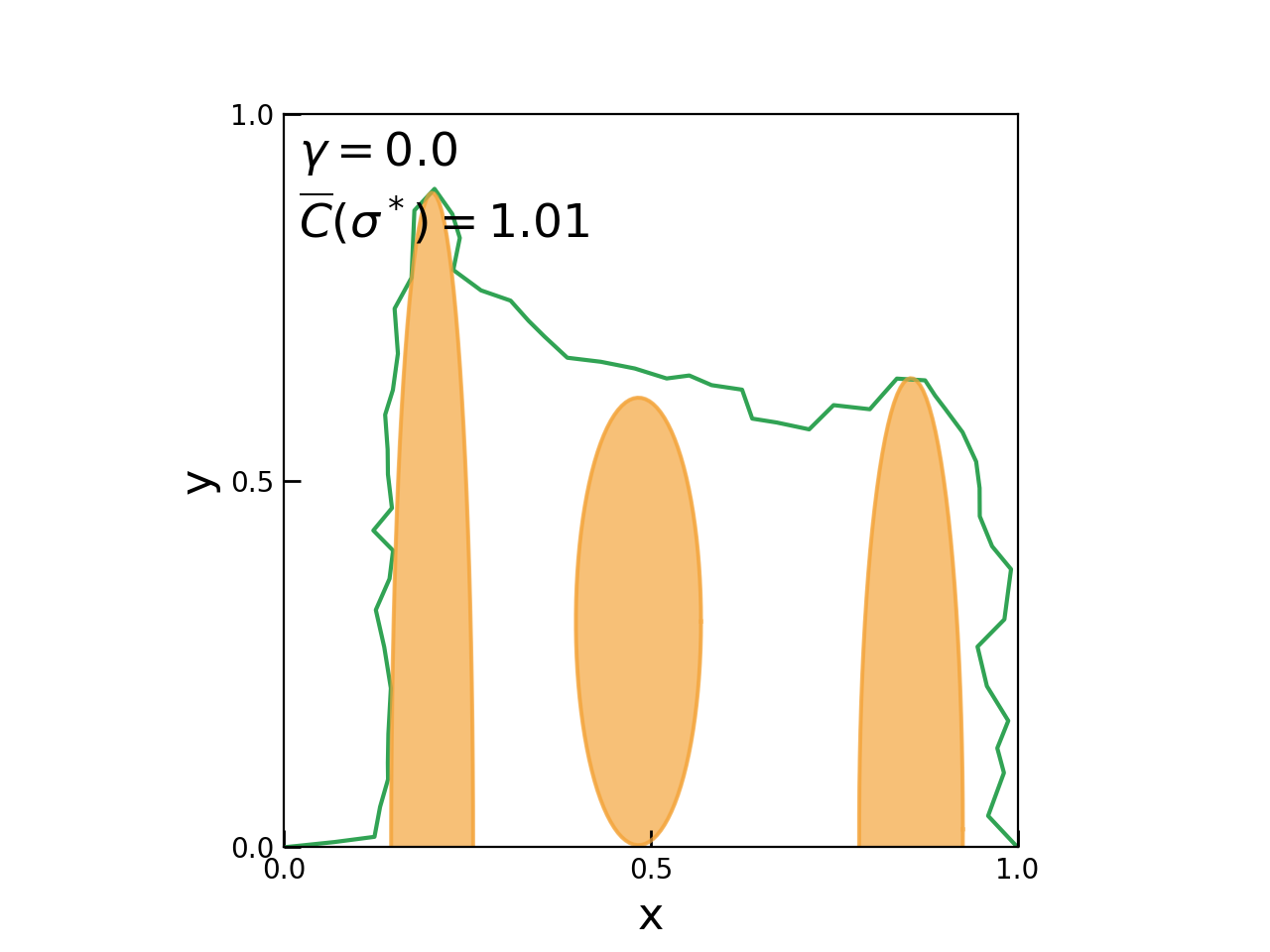}
    \input{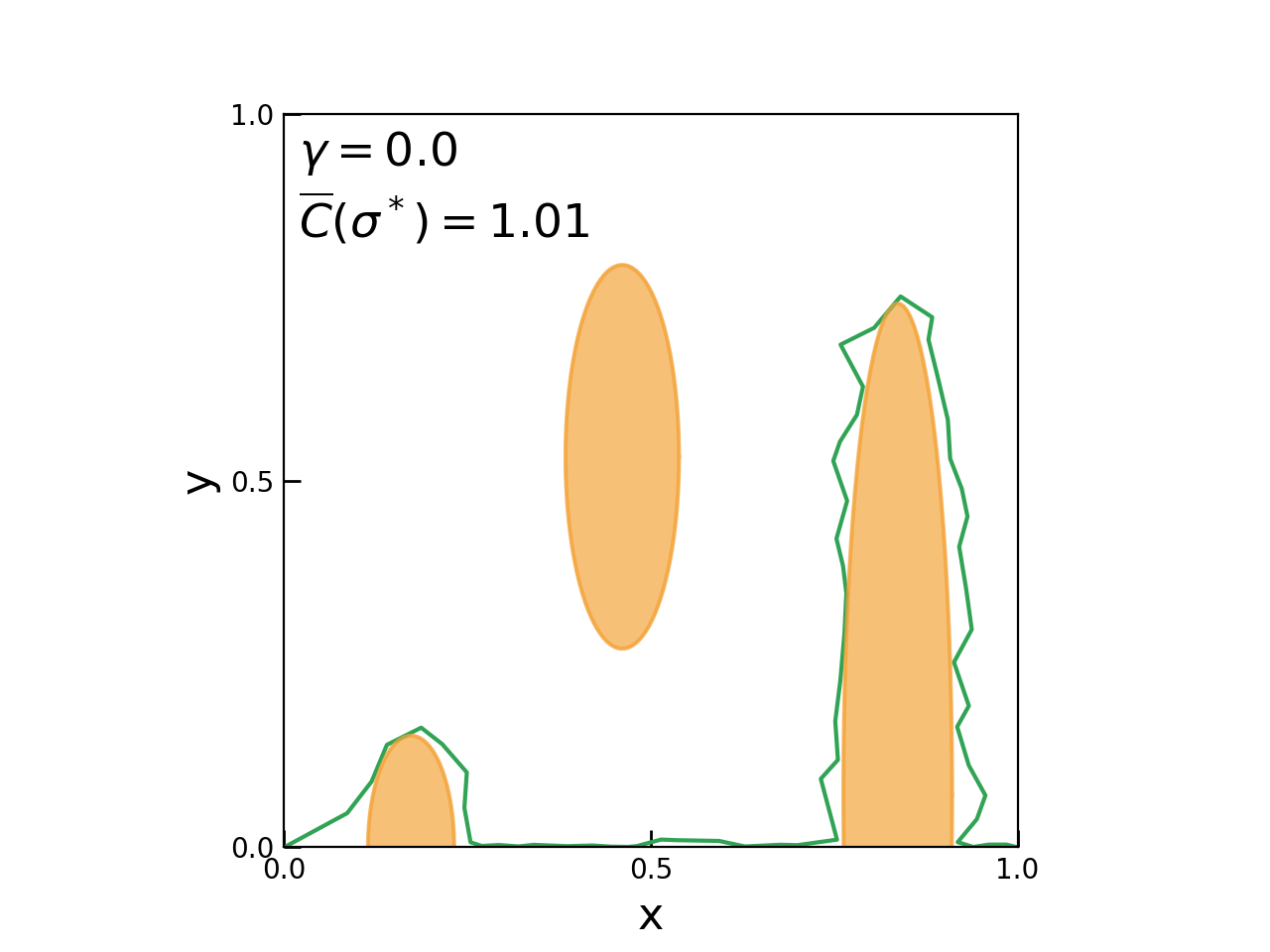}
    \vspace{2pt}
    \caption{4 examples of randomly generated obstacle spaces. Estimated optimal paths are displayed with $\gamma=0$, i.e. the planner tends to minimize the energy cost only. $\overline{C}(\hat{\sigma}^*)$ of the 4 scenarios above lies within $1.01$ and $1.02$.}
    \label{app-varying-obstacles}
\end{figure}
}{}

\ifthenelse{\boolean{includeFigures}}{
\begin{figure}
    \centering
    \input{icra2024/tex/appendix/gamma-normalized-cost}
    \caption{Normalized total cost with respect to the mixture parameter $\gamma$. Each data point denotes an average of results over 8 10-sec runs with varying obstacle space composition (Fig. \ref{app-varying-obstacles}). BIT* rewiring factor is set to 0.5, with 2000 samples per batch.}
    \label{app-gamma}
\end{figure}
}{}

To evaluate a degenerated form of the hybrid cost function, i.e. $\gamma=0$, we consider the environment in Fig. \ref{app-varying-obstacles} with a limited level of randomness of obstacle space compositions. A normalized cost $\overline{C}(\hat{\sigma}^*)$ is considered, instead of ${C}(\hat{\sigma}^*)$, to create a unified measure under various scenarios. Presented in Fig. \ref{app-gamma} and Tab. \ref{tab-gamma}, $\overline{C}(\hat{\sigma}^*)$ tends to $1.0$ as $\gamma$ decreases to $0.0$. $\overline{C}(\hat{\sigma}^*)$ slightly larger than $1.0$ indicates that the planner provides a convincing upper bound to the escape energy. Combining with the paths in Fig. \ref{app-varying-obstacles}, we take $\gamma=0$ as an empirically valid simplification of the hybrid cost function, although it slightly violates the strict positivity in the definition of cost function in \cite{karaman2011sampling}.

\begin{table}
\centering
\caption{Normalized total cost w.r.t the mixture parameter.}
\begin{tabular}{|l|l|l|l|l|}
\hline
$\gamma$        & $0.0$ & $10^{-3}$ & $10^{-2}$ & $10^{-1}$  \\
\hline
$\overline{C}(\hat{\sigma}^*)$ mean & 1.021 & 1.032   & 1.041 & 1.393 \\
\hline
$\overline{C}(\hat{\sigma}^*)$ std. & 0.012 & 0.016   & 0.009 & 0.065 \\
\hline
\end{tabular}
\label{tab-gamma}
\end{table}

\subsection{Qualitative Evaluation of Other Scenarios}

\ifthenelse{\boolean{includeFigures}}{
\begin{figure*}
    \centering
    \input{icra2024/tex/appendix/varying-gammas/gamma-0.0}
    \input{icra2024/tex/appendix/varying-gammas/gamma-1.0}
    \input{icra2024/tex/appendix/varying-gammas/gamma-10.0}
    \caption{Estimated optimal paths found by an asymptotically optimal algorithm (BIT*) with a hybrid cost function. Varying values of $\gamma \in \{0.0,1.0,10.0\}$ are used in 3 scenarios with fixed obstacles. Paths are denoted in green and ellipsoidal obstacles in orange. The runtime and BIT* setup stay the same as Fig. \ref{app-gamma}.}
    \label{app-varying-gammas}
\end{figure*}
}{}

\ifthenelse{\boolean{includeFigures}}{
\begin{figure*}
    \centering
    \includegraphics[width=0.47\linewidth]{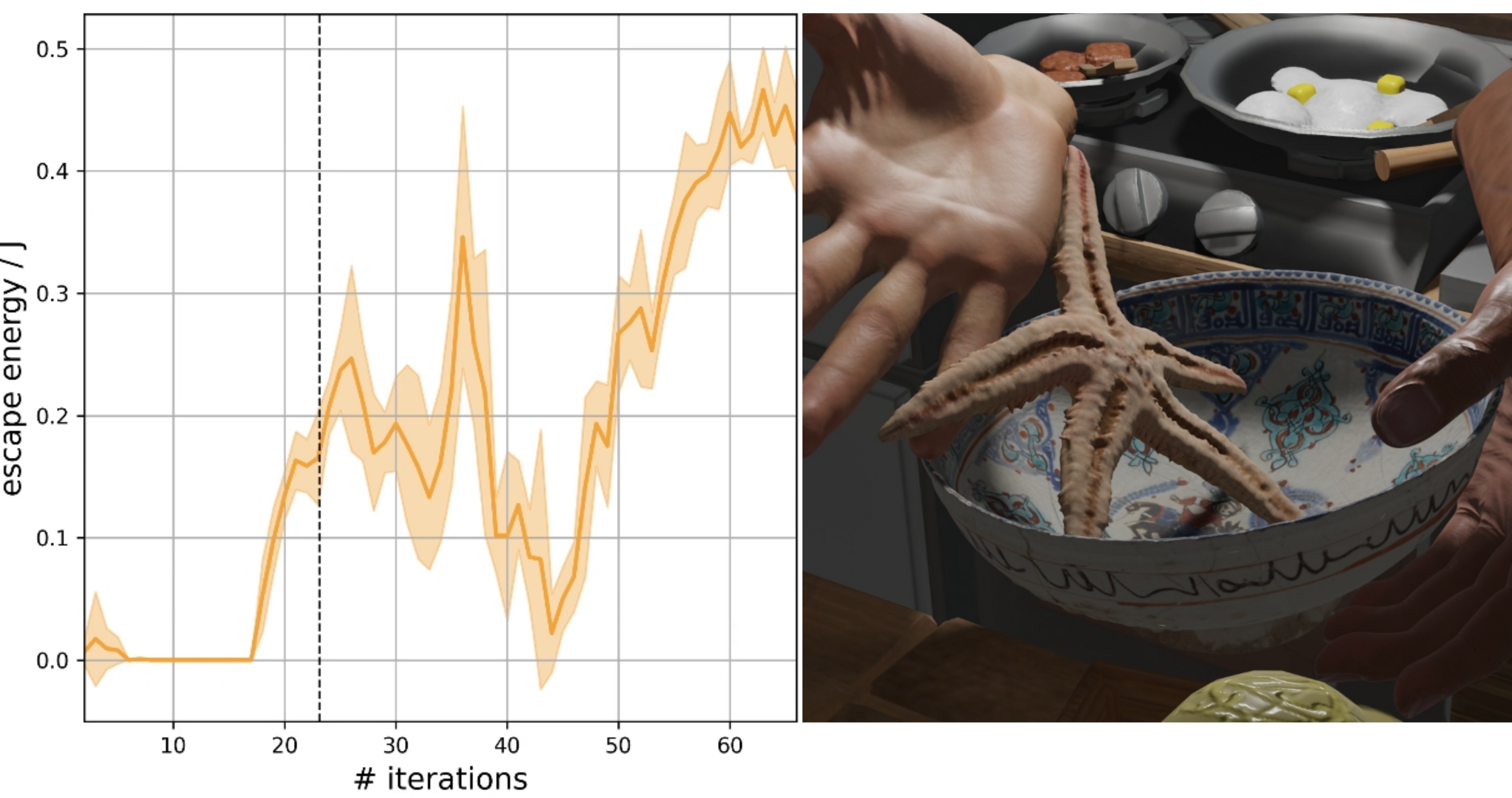}
    \vspace{2pt}
    \includegraphics[width=0.47\linewidth]{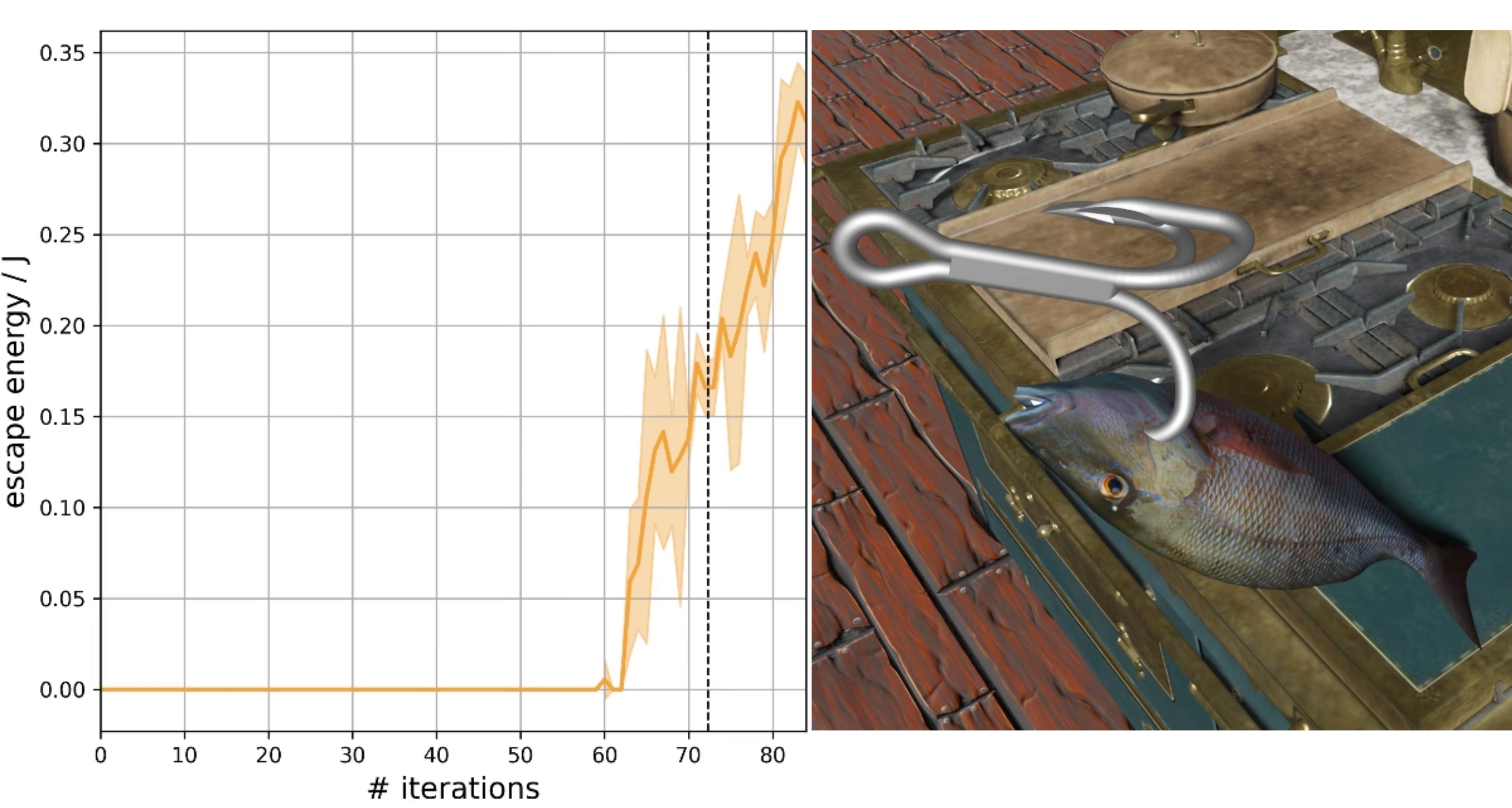}
    
    \centering
    \includegraphics[width=0.47\linewidth]{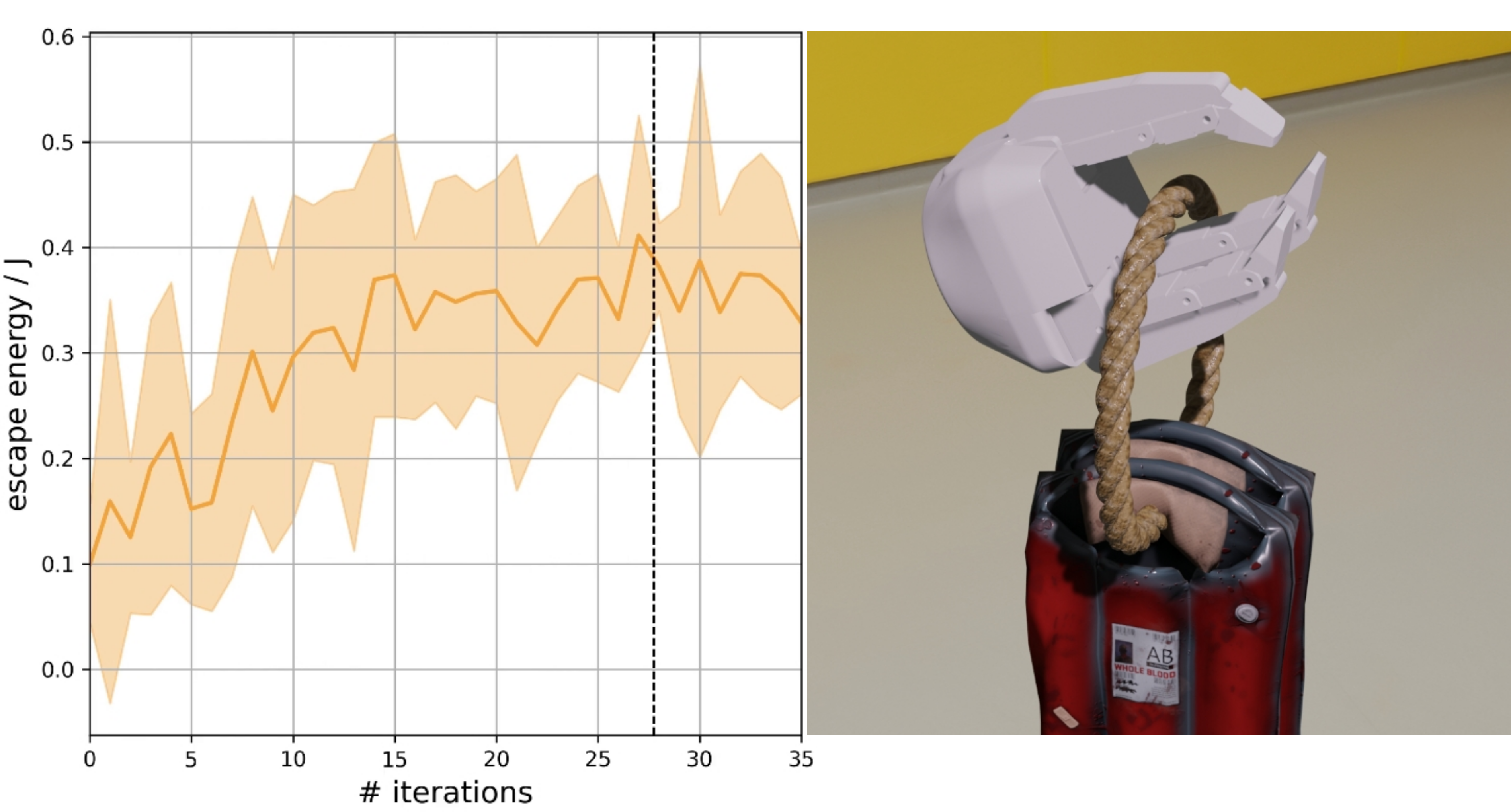}
    \vspace{2pt}
    \includegraphics[width=0.47\linewidth]{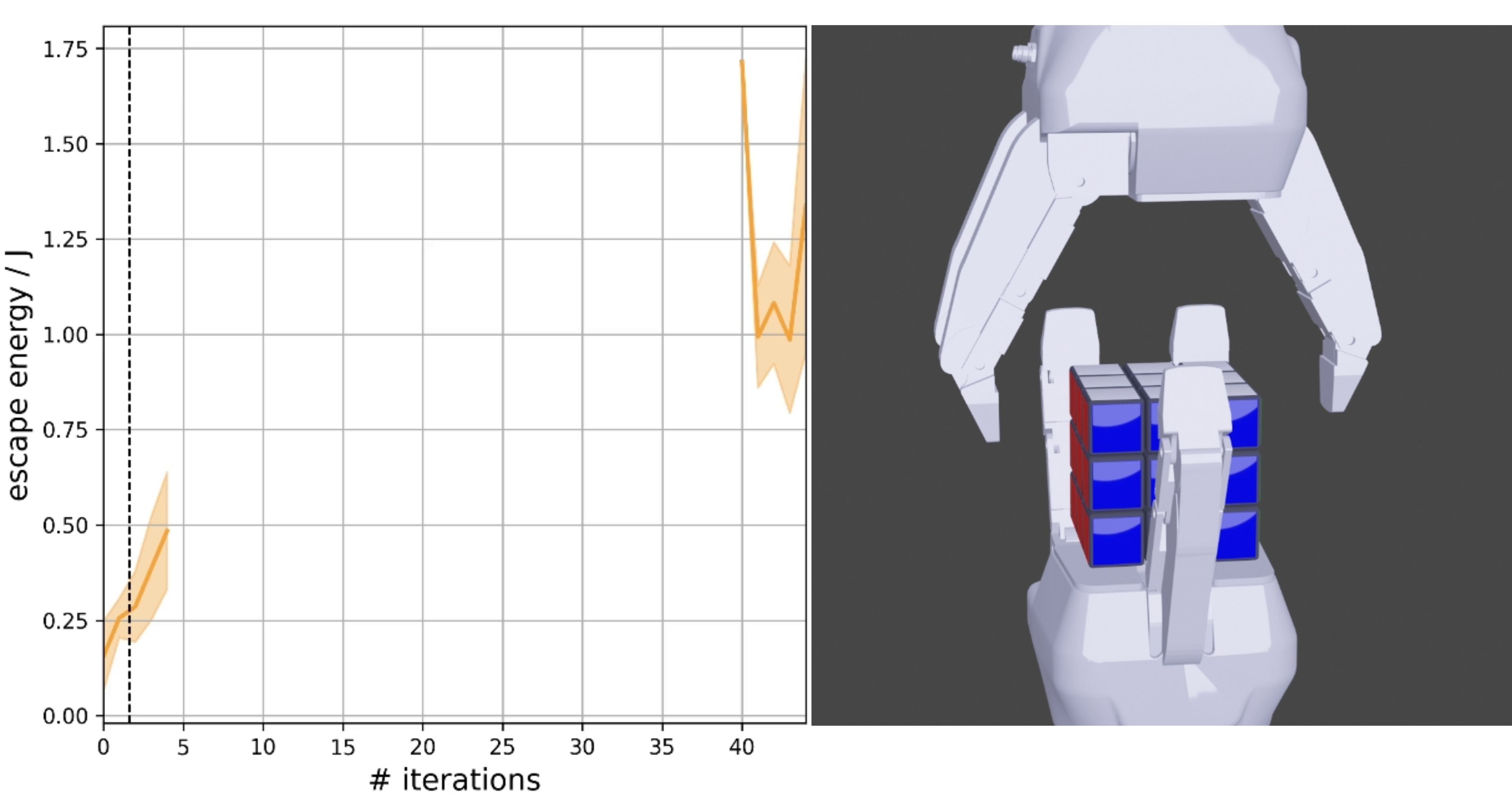}
    
    \caption{From top to bottom: (1) catching a starfish with a bowl, (2) hooking a frozen fish with a fishing hook, (3) grasping a rope loop of blood bags, (4) manipulating a Rubik cube with two grippers. Escape energy is plotted over iterations. One of the iterations is illustrated to the right of each plot, corresponding to the dashed verticle lines in the plots.}
    \label{fig-additional-scenarios}
\end{figure*}
}{}

4 additional scenarios of soft fixtures are presented in Fig. \ref{fig-additional-scenarios}. In the first example, a starfish, modeled by kinematic links, falls into a bowl from a tilting human palm. The other hand, holding the rim of the bowl, wobbles the bowl back and forth to make the starfish slide down to the bottom of the bowl. In the meantime, the escape energy estimated by BIT* goes up and down. In a second scenario (Fig. \ref{fig-additional-scenarios}, top right), a frozen fish, modeled as a rigid object, lies initially on the table. It is pierced by a fish hook through its gill and lifted. A rope loop with blood bags, as presented in the bottom left, is fixtured by a Robotiq gripper. The loop is modeled by kinematic links and only the gravitational potential energy of the blood bags is considered. To firmly grasp the loop in hand, the wrist rotates and the palm turns upward, as the fingers further close. The last scenario emulates the process that a Rubik cube player examines the cube by holding it in hand. The other gripper covers it from the top and forms a cage (infinite escape energy in the middle of the plot). The two grippers rotate, centering at the cube, ending up being aligned horizontally. As the two fingers covering the cube open up, the cage is broken up (the last few iterations in the plot). Please refer to the complementary video for more details. 

We observe from above that real-life fixturing motions tend to increase the escape energy of objects so that they won't readily escape, even in the existence of external disturbances. This indicates the capability of escape energy in measuring the stability of caging-based grasps. It also inspires us to explore the correlation between escape energy and the probability of escape in real-world experiments. Technical details of all 8 soft fixtures scenarios in Fig. \ref{fig-scenarios} are listed in Tab. \ref{tab-scenarios}.

\subsection{Details of Real World Experiments}
The initial configuration of the 7-axis robot arm with joint angles in Fig. \ref{fig-phys} (i-1) could be expressed as $\boldsymbol{\beta}_{\text{init}} = [0, -\pi/4, 0, -3\pi/4, 0, \pi/2, \pi/4]$.
For a perturbation level $\tau$, the arm is commanded to reach a perturbed configuration $\boldsymbol{\beta}_p $ with joint $\beta_p^k, k \in \{2,3,5,6,7\} $ uniformly sampled from $ [\beta_{\text{init}}^k-\pi\tau/4,\beta_{\text{init}}^k+\pi\tau/4]$, which repeats 10 times. Therefore, the arm reaches 10 perturbed configurations consecutively under $\tau$. Joint $k \in \{1,4\}$ stay fixed since their aggressive and random movement might cause a collision. 
The perturbation level $\tau$ increases from $0$ with a fixed resolution, for instance, $\Delta\tau = 0.03$ until objects escape from the caging tool. 
We placed an object to the bottom of the caging tool, marking its configuration w.r.t. the tool within the inner side of the bowl-like tool. This is a valid way since most of the objects are regular in shape and 3D printed, a perception system with visual markers (i.e. April tag) attached to the objects is better but too expensive for our demands.
}{}

\end{document}

%% file: icra2024/tex/quantitative-tests/sum.tex
\begin{tikzpicture}[font=\footnotesize]

\definecolor{darkgray176}{RGB}{176,176,176}
\definecolor{lightgray204}{RGB}{204,204,204}
\definecolor{orangered2405932}{RGB}{240,59,32}
\definecolor{seagreen4916384}{RGB}{49,163,84}
\definecolor{slateblue117107177}{RGB}{117,107,177}
\definecolor{steelblue43140190}{RGB}{43,140,190}
\definecolor{sandybrown24416662}{RGB}{244,166,62}

\begin{axis}[name=zero,
width=0.1225\linewidth, height=0.245\linewidth, scale only axis,enlargelimits=false,
yticklabels={,,}, xticklabels={,,},
tick style={draw=none}
]
\addplot graphics [includegraphics, xmin=0, xmax=1023.5, ymin=1023.5, ymax=0] {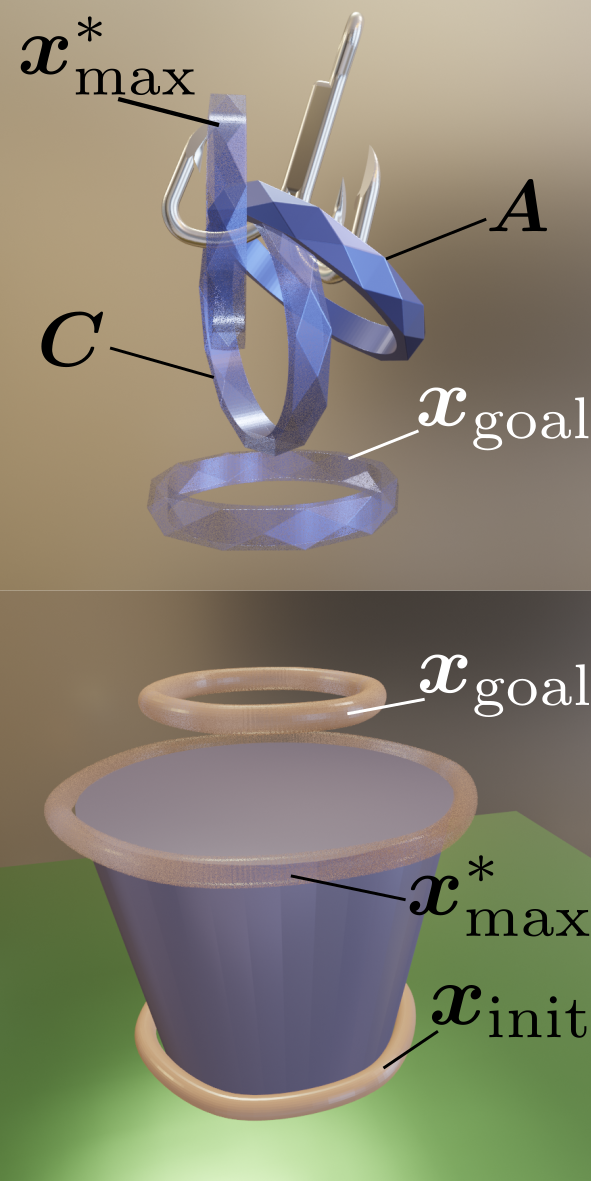};
\draw (axis cs:700,40) node[
  scale=1,
  anchor=base west,
  text=white,
  rotate=0.0
]{(i-2)};
\draw (axis cs:700,540) node[
  scale=1,
  anchor=base west,
  text=white,
  rotate=0.0
]{(i-1)};
\end{axis}

\input{icra2024/tex/quantitative-tests/test01/test01-3basic}

\input{icra2024/tex/quantitative-tests/test05/test05}

\input{icra2024/tex/quantitative-tests/test04/test04}

\end{tikzpicture}

%% file: icra2024/tex/band-ear-mosaic.tex
\begin{tikzpicture}[font=\footnotesize]

\definecolor{darkgray176}{RGB}{176,176,176}
\definecolor{lightgray204}{RGB}{204,204,204}
\definecolor{sandybrown24416662}{RGB}{244,166,62}
\definecolor{seagreen4916384}{RGB}{49,163,84}

\begin{axis}[name=one, width=0.44\linewidth, height=0.44\linewidth, scale only axis,
legend cell align={left},
legend style={fill opacity=0.8, draw opacity=1, text opacity=1, draw=lightgray204, at={(0.02,0.98)}, anchor=north west,},
tick align=inside,
tick pos=left,
x grid style={darkgray176},
xmajorgrids,
xmin=0, xmax=43,
x label style={at={(axis description cs:0.5,-0.05)}, anchor=north},
xlabel={frame},
y grid style={darkgray176},
y label style={at={(axis description cs:-0.09,.5)},anchor=south},
ylabel={escape energy / J},
ymajorgrids,
ymin=-0.01, ymax=0.36,
]
\path [draw=sandybrown24416662, fill=sandybrown24416662, opacity=0.4]
(axis cs:0,0)
--(axis cs:0,0)
--(axis cs:1,0)
--(axis cs:2,0)
--(axis cs:3,0)
--(axis cs:4,0)
--(axis cs:5,0.0137389372326945)
--(axis cs:6,0.020433520439641)
--(axis cs:7,0.0315536793244082)
--(axis cs:8,0.0934668578469165)
--(axis cs:9,0.0907325846030632)
--(axis cs:10,0.0827693215900094)
--(axis cs:11,0.105906676612597)
--(axis cs:12,0.0997425408408701)
--(axis cs:13,0.123081128665719)
--(axis cs:14,0.0898872227121655)
--(axis cs:15,0.122497393210673)
--(axis cs:16,0.114181950260943)
--(axis cs:17,0.127719871929026)
--(axis cs:18,0.12849071263176)
--(axis cs:19,0.113252799948205)
--(axis cs:20,0.123201108758737)
--(axis cs:21,0.119179998083261)
--(axis cs:22,0.12531893531763)
--(axis cs:23,0.152720449248032)
--(axis cs:24,0.126663388014935)
--(axis cs:25,0.141603583239722)
--(axis cs:26,0.130898307627834)
--(axis cs:27,0.136860995662585)
--(axis cs:28,0.152152928849076)
--(axis cs:29,0.144364897618197)
--(axis cs:30,0.129419842811071)
--(axis cs:31,0.120033141599821)
--(axis cs:32,0.127489251877648)
--(axis cs:33,0.149711993251803)
--(axis cs:34,0.149584805106791)
--(axis cs:35,0.115315814189935)
--(axis cs:36,0.140945777218533)
--(axis cs:37,0.139574023450548)
--(axis cs:38,0.168568669522392)
--(axis cs:39,0.115593158082403)
--(axis cs:40,0.138304383288974)
--(axis cs:41,0.147031116387738)
--(axis cs:42,0.158425150643829)
--(axis cs:43,0.173272749480949)
--(axis cs:43,0.270337930797018)
--(axis cs:43,0.270337930797018)
--(axis cs:42,0.275459000233042)
--(axis cs:41,0.24126377864993)
--(axis cs:40,0.233158452759217)
--(axis cs:39,0.364403473986316)
--(axis cs:38,0.25737333953273)
--(axis cs:37,0.22574241786789)
--(axis cs:36,0.235798068400229)
--(axis cs:35,0.261052213337835)
--(axis cs:34,0.242592349287338)
--(axis cs:33,0.243121154418057)
--(axis cs:32,0.201762512831504)
--(axis cs:31,0.230644586271129)
--(axis cs:30,0.341368866004472)
--(axis cs:29,0.224459411455895)
--(axis cs:28,0.218820808848878)
--(axis cs:27,0.300882694802772)
--(axis cs:26,0.178864047639683)
--(axis cs:25,0.187107450123109)
--(axis cs:24,0.210383860139938)
--(axis cs:23,0.254064188210879)
--(axis cs:22,0.228658852934527)
--(axis cs:21,0.169677845258336)
--(axis cs:20,0.181822262402849)
--(axis cs:19,0.179968713122355)
--(axis cs:18,0.191521482465677)
--(axis cs:17,0.178037962704814)
--(axis cs:16,0.165361422127051)
--(axis cs:15,0.189518362769285)
--(axis cs:14,0.162712329868706)
--(axis cs:13,0.188954970813868)
--(axis cs:12,0.178093949517313)
--(axis cs:11,0.162207410113351)
--(axis cs:10,0.136766188604093)
--(axis cs:9,0.153044870173467)
--(axis cs:8,0.162780954595588)
--(axis cs:7,0.0915604638995122)
--(axis cs:6,0.0536329112211946)
--(axis cs:5,0.0501616890728072)
--(axis cs:4,0)
--(axis cs:3,0)
--(axis cs:2,0)
--(axis cs:1,0)
--(axis cs:0,0)
--cycle;

\addplot [thick, sandybrown24416662]
table {%
0 0
1 0
2 0
3 0
4 0
5 0.0319503131527508
6 0.0370332158304178
7 0.0615570716119602
8 0.128123906221252
9 0.121888727388265
10 0.109767755097051
11 0.134057043362974
12 0.138918245179092
13 0.156018049739794
14 0.126299776290436
15 0.156007877989979
16 0.139771686193997
17 0.15287891731692
18 0.160006097548718
19 0.14661075653528
20 0.152511685580793
21 0.144428921670799
22 0.176988894126078
23 0.203392318729455
24 0.168523624077437
25 0.164355516681416
26 0.154881177633759
27 0.218871845232678
28 0.185486868848977
29 0.184412154537046
30 0.235394354407771
31 0.175338863935475
32 0.164625882354576
33 0.19641657383493
34 0.196088577197064
35 0.188184013763885
36 0.188371922809381
37 0.182658220659219
38 0.212971004527561
39 0.23999831603436
40 0.185731418024096
41 0.194147447518834
42 0.216942075438435
43 0.221805340138983
};
\addplot [black, dashed, forget plot]
table {%
2.0 0
2.0 0.45
};
\addplot [black, dashed, forget plot]
table {%
17 0
17 0.45
};
\addplot [black, dashed, forget plot]
table {%
22 0
22 0.45
};
\addplot [black, dashed, forget plot]
table {%
38 0
38 0.45
};
\draw (axis cs:2.1,-0.0) node[
  scale=0.9,
  anchor=base west,
  text=black,
  rotate=0.0
]{A};
\draw (axis cs:17.1,-0.0) node[
  scale=0.9,
  anchor=base west,
  text=black,
  rotate=0.0
]{B};
\draw (axis cs:22.3,-0.0) node[
  scale=0.9,
  anchor=base west,
  text=black,
  rotate=0.0
]{C};
\draw (axis cs:38.3,-0.0) node[
  scale=0.9,
  anchor=base west,
  text=black,
  rotate=0.0
]{D};
\end{axis}


\begin{axis}[name=two,
width=0.22\linewidth, height=0.22\linewidth, scale only axis,enlargelimits=false,
at=(one.north east), anchor=north west,
yticklabels={,,}, xticklabels={,,},
tick style={draw=none}
]
\addplot graphics [includegraphics, xmin=0, xmax=1023.5, ymin=1023.5, ymax=0] {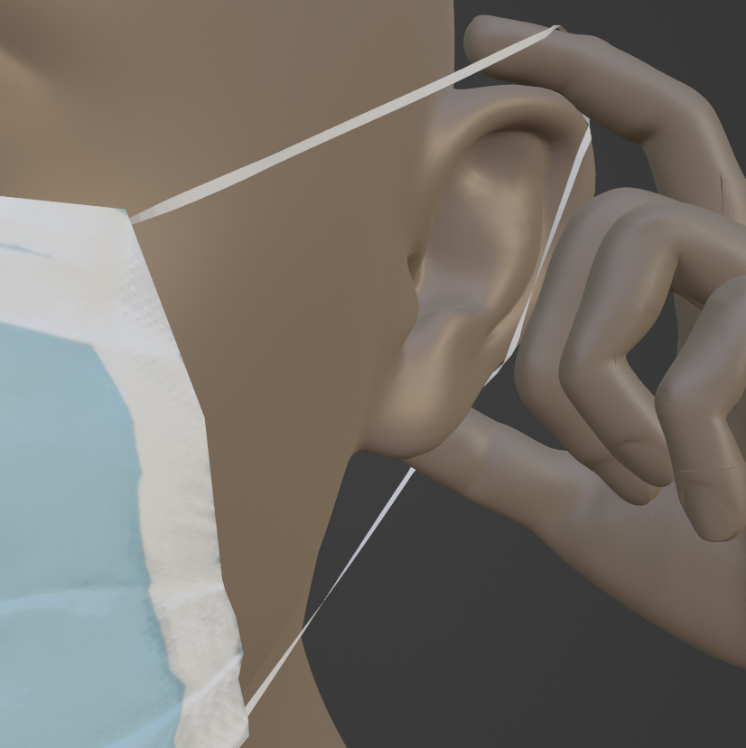};
\draw (axis cs:0,-0.1) node[
  scale=0.9,
  anchor=base west,
  text=black,
  rotate=0.0
]{A};
\end{axis}

\begin{axis}[name=three,
width=0.22\linewidth, height=0.22\linewidth, scale only axis,enlargelimits=false,
at=(two.north east), anchor=north west,
yticklabel pos=right,
yticklabels={,,}, xticklabels={,,},
tick style={draw=none},
]
\addplot graphics [includegraphics cmd=\pgfimage,xmin=-0.5, xmax=1023.5, ymin=1023.5, ymax=-0.5] {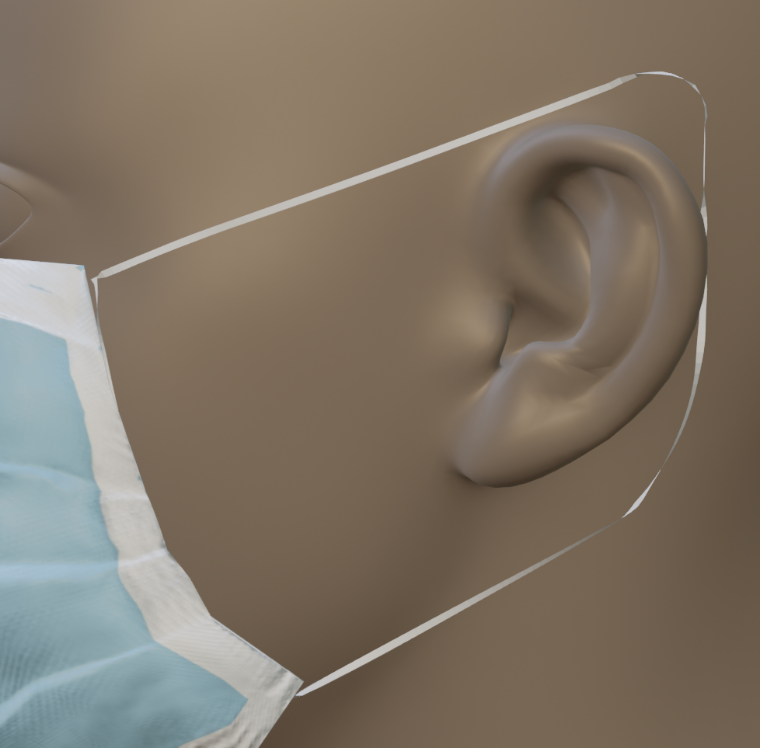};
\draw (axis cs:0,-0.1) node[
  scale=0.9,
  anchor=base west,
  text=black,
  rotate=0.0
]{B};
\end{axis}

\begin{axis}[name=four,
width=0.22\linewidth, height=0.22\linewidth, scale only axis,enlargelimits=false,
at=(one.east), anchor=north west,
yticklabel pos=right,
yticklabels={,,}, xticklabels={,,},
tick style={draw=none},
]
\addplot graphics [includegraphics cmd=\pgfimage,xmin=-0.5, xmax=1023.5, ymin=1023.5, ymax=-0.5] {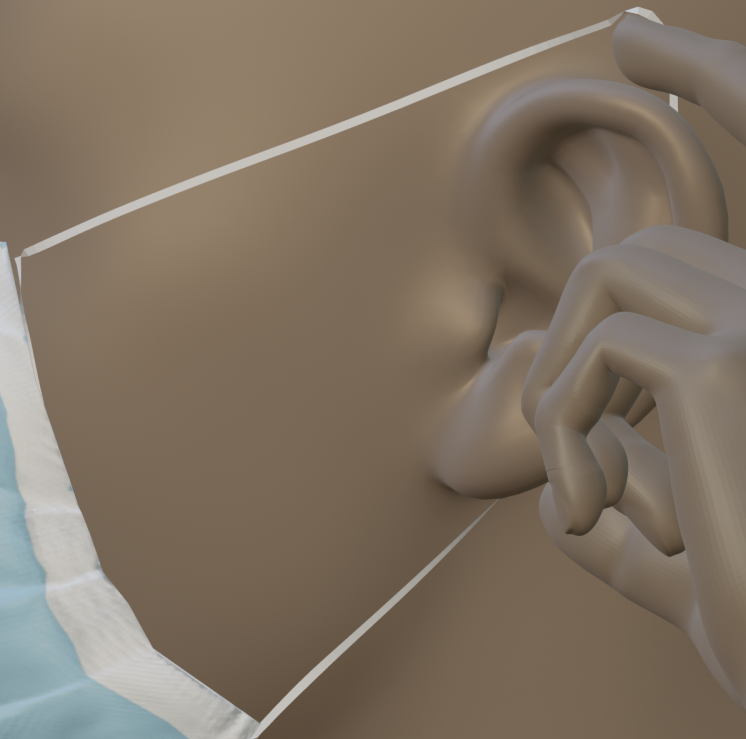};
\draw (axis cs:0,-0.1) node[
  scale=0.9,
  anchor=base west,
  text=black,
  rotate=0.0
]{C};
\end{axis}

\begin{axis}[name=five,
width=0.22\linewidth, height=0.22\linewidth, scale only axis,enlargelimits=false,
at=(four.south east), anchor=south west,
yticklabel pos=right,
yticklabels={,,}, xticklabels={,,},
tick style={draw=none},
]
\addplot graphics [includegraphics cmd=\pgfimage,xmin=-0.5, xmax=1023.5, ymin=1023.5, ymax=-0.5] {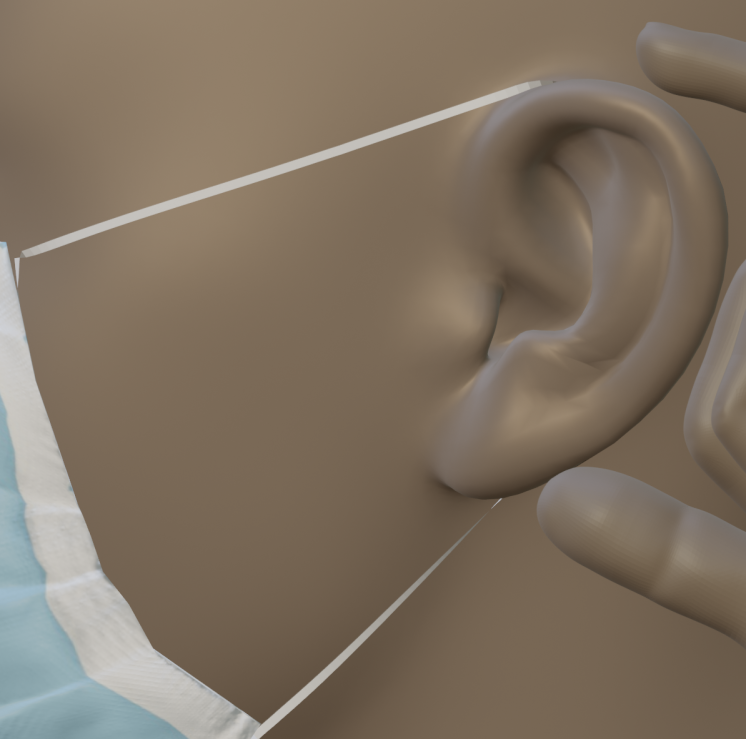};
\draw (axis cs:0,-0.1) node[
  scale=0.9,
  anchor=base west,
  text=black,
  rotate=0.0
]{D};
\end{axis}

\end{tikzpicture}

%% file: icra2024/tex/fish-shovel-mosaic.tex
\begin{tikzpicture}[font=\footnotesize]

\definecolor{darkgray176}{RGB}{176,176,176}
\definecolor{lightgray204}{RGB}{204,204,204}
\definecolor{sandybrown24416662}{RGB}{244,166,62}
\definecolor{seagreen4916384}{RGB}{49,163,84}

\begin{axis}[name=one, width=0.44\linewidth, height=0.44\linewidth, scale only axis,
legend cell align={left},
legend style={fill opacity=0.8, draw opacity=1, text opacity=1, draw=lightgray204, at={(0.02,0.98)}, anchor=north west,},
tick align=inside,
tick pos=left,
x grid style={darkgray176},
xmajorgrids,
xmin=0, xmax=70,
x label style={at={(axis description cs:0.5,-0.05)}, anchor=north},
xlabel={frame},
y grid style={darkgray176},
y label style={at={(axis description cs:-0.066,.5)},anchor=south},
ylabel={escape energy / J},
ymajorgrids,
ymin=-0.05, ymax=3.1,
]
\path [draw=sandybrown24416662, fill=sandybrown24416662, opacity=0.4]
(axis cs:0,0)
--(axis cs:0,0)
--(axis cs:1.13755687784389,1.03823386769232)
--(axis cs:2.27511375568778,1.39598528438367)
--(axis cs:3.41267063353168,0.877791737708901)
--(axis cs:4.55022751137557,0.959816736504835)
--(axis cs:5.68778438921946,1.05613892533137)
--(axis cs:6.82534126706335,0.864153071767281)
--(axis cs:7.96289814490725,1.05902093391642)
--(axis cs:9.10045502275114,0.421425081327182)
--(axis cs:10.238011900595,1.33333048045528)
--(axis cs:11.3755687784389,1.35244101752027)
--(axis cs:12.5131256562828,0.620373014703291)
--(axis cs:13.6506825341267,0.651513695405539)
--(axis cs:14.7882394119706,0.908071447863422)
--(axis cs:15.9257962898145,0.947661597166829)
--(axis cs:17.0633531676584,1.31963581606956)
--(axis cs:18.2009100455023,0.920451531077932)
--(axis cs:19.3384669233462,0.84398244882185)
--(axis cs:20.4760238011901,0.509360246315535)
--(axis cs:21.613580679034,0.867801033820097)
--(axis cs:22.7511375568778,0.904339826660183)
--(axis cs:23.8886944347217,1.17421007060075)
--(axis cs:25.0262513125656,0.54861514150086)
--(axis cs:26.1638081904095,1.04971418013112)
--(axis cs:27.3013650682534,0.900283759144947)
--(axis cs:28.4389219460973,0.251284116717893)
--(axis cs:29.5764788239412,0.759680652258656)
--(axis cs:30.7140357017851,0.953819685008166)
--(axis cs:31.851592579629,1.22481886108371)
--(axis cs:32.9891494574729,1.38825412076067)
--(axis cs:34.1267063353168,1.84989367131452)
--(axis cs:35.2642632131607,1.41435055693679)
--(axis cs:36.4018200910046,1.43758324931322)
--(axis cs:37.5393769688484,1.57970691102016)
--(axis cs:38.6769338466923,1.42799277775716)
--(axis cs:39.8144907245362,1.23187229903729)
--(axis cs:40.9520476023801,1.54124071402398)
--(axis cs:42.089604480224,1.45371410192215)
--(axis cs:43.2271613580679,1.65421532880323)
--(axis cs:44.3647182359118,1.69918488233666)
--(axis cs:45.5022751137557,1.65624141429496)
--(axis cs:46.6398319915996,0.910128875857797)
--(axis cs:47.7773888694435,0.675680229741845)
--(axis cs:48.9149457472874,-0.160718581718112)
--(axis cs:49.0024501225061,0)
--(axis cs:49.8774938746937,-0.114511804153429)
--(axis cs:50.7525376268813,-0.151545855848094)
--(axis cs:51.627581379069,0)
--(axis cs:52.5026251312566,0)
--(axis cs:53.3776688834442,-0.0021762392439075)
--(axis cs:54.2527126356318,-0.0739678561805089)
--(axis cs:55.1277563878194,0)
--(axis cs:56.002800140007,-0.14982386328065)
--(axis cs:56.8778438921946,-0.00167586874102439)
--(axis cs:57.7528876443822,-0.00355489040505396)
--(axis cs:58.6279313965698,-0.181418535353383)
--(axis cs:59.5029751487574,0)
--(axis cs:60.3780189009451,-0.0959019440148102)
--(axis cs:61.2530626531327,-0.127337319883415)
--(axis cs:62.1281064053203,-0.113272047984218)
--(axis cs:63.0031501575079,-0.207897629120442)
--(axis cs:63.8781939096955,-0.0596850410003882)
--(axis cs:64.7532376618831,0.478106027765717)
--(axis cs:65.6282814140707,0.638472023045862)
--(axis cs:66.5033251662583,1.41652292403708)
--(axis cs:67.3783689184459,1.32559064450562)
--(axis cs:68.2534126706335,1.25712478080384)
--(axis cs:69.1284564228211,0.542202066268613)
--(axis cs:70.0035001750088,1.71346840175277)
--(axis cs:70.0035001750088,3.7188249265225)
--(axis cs:70.0035001750088,3.7188249265225)
--(axis cs:69.1284564228211,2.86485458208415)
--(axis cs:68.2534126706335,3.07839920713433)
--(axis cs:67.3783689184459,2.67793875743735)
--(axis cs:66.5033251662583,3.24750457654806)
--(axis cs:65.6282814140707,3.0100343028104)
--(axis cs:64.7532376618831,2.313262334369)
--(axis cs:63.8781939096955,1.33330637433691)
--(axis cs:63.0031501575079,0.908733401326761)
--(axis cs:62.1281064053203,0.952042822222398)
--(axis cs:61.2530626531327,0.854189897959494)
--(axis cs:60.3780189009451,0.765861530786107)
--(axis cs:59.5029751487574,0)
--(axis cs:58.6279313965698,0.565333538034773)
--(axis cs:57.7528876443822,0.00662391601494445)
--(axis cs:56.8778438921946,0.0031226880516019)
--(axis cs:56.002800140007,0.279170543765463)
--(axis cs:55.1277563878194,0)
--(axis cs:54.2527126356318,0.137826152516154)
--(axis cs:53.3776688834442,0.00405504089790658)
--(axis cs:52.5026251312566,0)
--(axis cs:51.627581379069,0)
--(axis cs:50.7525376268813,0.456082574748656)
--(axis cs:49.8774938746937,0.473699473978519)
--(axis cs:49.0024501225061,0)
--(axis cs:48.9149457472874,0.520086597270931)
--(axis cs:47.7773888694435,2.08111952629304)
--(axis cs:46.6398319915996,2.53830413802223)
--(axis cs:45.5022751137557,2.75098009720459)
--(axis cs:44.3647182359118,3.08272225305486)
--(axis cs:43.2271613580679,2.94421561522182)
--(axis cs:42.089604480224,2.75666257371757)
--(axis cs:40.9520476023801,3.04505550229309)
--(axis cs:39.8144907245362,2.54410508516196)
--(axis cs:38.6769338466923,2.59101682229736)
--(axis cs:37.5393769688484,2.8258452769329)
--(axis cs:36.4018200910046,2.70489095614315)
--(axis cs:35.2642632131607,2.71906580327554)
--(axis cs:34.1267063353168,2.84371293007826)
--(axis cs:32.9891494574729,2.90168307788075)
--(axis cs:31.851592579629,2.60324707018411)
--(axis cs:30.7140357017851,2.40024777691663)
--(axis cs:29.5764788239412,1.97193893781591)
--(axis cs:28.4389219460973,1.52493516243522)
--(axis cs:27.3013650682534,2.04004271373263)
--(axis cs:26.1638081904095,2.08284467934622)
--(axis cs:25.0262513125656,1.51411147811519)
--(axis cs:23.8886944347217,2.22585351390888)
--(axis cs:22.7511375568778,2.15028264964033)
--(axis cs:21.613580679034,1.94995041500253)
--(axis cs:20.4760238011901,1.96920599652563)
--(axis cs:19.3384669233462,2.16896831339511)
--(axis cs:18.2009100455023,2.3652243224755)
--(axis cs:17.0633531676584,2.52046028639598)
--(axis cs:15.9257962898145,2.44680275019142)
--(axis cs:14.7882394119706,2.03716546339831)
--(axis cs:13.6506825341267,1.71818949494806)
--(axis cs:12.5131256562828,2.24590280519261)
--(axis cs:11.3755687784389,2.4015205427758)
--(axis cs:10.238011900595,2.00117492448616)
--(axis cs:9.10045502275114,2.54751899837534)
--(axis cs:7.96289814490725,1.9170742420101)
--(axis cs:6.82534126706335,2.23584602659115)
--(axis cs:5.68778438921946,1.65071551336737)
--(axis cs:4.55022751137557,2.14641759039566)
--(axis cs:3.41267063353168,1.84904039617877)
--(axis cs:2.27511375568778,1.93188440026696)
--(axis cs:1.13755687784389,2.20124234267544)
--(axis cs:0,0)
--cycle;

\addplot [thick, sandybrown24416662]
table {%
0 0
1.13755687784389 1.61973810518388
2.27511375568778 1.66393484232531
3.41267063353168 1.36341606694384
4.55022751137557 1.55311716345024
5.68778438921946 1.35342721934937
6.82534126706335 1.54999954917922
7.96289814490725 1.48804758796326
9.10045502275114 1.48447203985126
10.238011900595 1.66725270247072
11.3755687784389 1.87698078014804
12.5131256562828 1.43313790994795
13.6506825341267 1.1848515951768
14.7882394119706 1.47261845563087
15.9257962898145 1.69723217367912
17.0633531676584 1.92004805123277
18.2009100455023 1.64283792677672
19.3384669233462 1.50647538110848
20.4760238011901 1.23928312142058
21.613580679034 1.40887572441131
22.7511375568778 1.52731123815026
23.8886944347217 1.70003179225481
25.0262513125656 1.03136330980803
26.1638081904095 1.56627942973867
27.3013650682534 1.47016323643879
28.4389219460973 0.888109639576557
29.5764788239412 1.36580979503728
30.7140357017851 1.6770337309624
31.851592579629 1.91403296563391
32.9891494574729 2.14496859932071
34.1267063353168 2.34680330069639
35.2642632131607 2.06670818010617
36.4018200910046 2.07123710272818
37.5393769688484 2.20277609397653
38.6769338466923 2.00950480002726
39.8144907245362 1.88798869209963
40.9520476023801 2.29314810815854
42.089604480224 2.10518833781986
43.2271613580679 2.29921547201252
44.3647182359118 2.39095356769576
45.5022751137557 2.20361075574977
46.6398319915996 1.72421650694001
47.7773888694435 1.37839987801744
48.9149457472874 0.179684007776409
49.0024501225061 0
49.8774938746937 0.179593834912545
50.7525376268813 0.152268359450281
51.627581379069 0
52.5026251312566 0
53.3776688834442 0.000939400826999541
54.2527126356318 0.0319291481678228
55.1277563878194 0
56.002800140007 0.064673340242407
56.8778438921946 0.000723409655288757
57.7528876443822 0.00153451280494525
58.6279313965698 0.191957501340695
59.5029751487574 0
60.3780189009451 0.334979793385648
61.2530626531327 0.36342628903804
62.1281064053203 0.41938538711909
63.0031501575079 0.35041788610316
63.8781939096955 0.636810666668259
64.7532376618831 1.39568418106736
65.6282814140707 1.82425316292813
66.5033251662583 2.33201375029257
67.3783689184459 2.00176470097148
68.2534126706335 2.16776199396909
69.1284564228211 1.70352832417638
70.0035001750088 2.71614666413763
};
\addplot [black, dashed, forget plot]
table {%
17.5 0
17.5 4
};
\addplot [black, dashed, forget plot]
table {%
35 0
35 4
};
\addplot [black, dashed, forget plot]
table {%
52.5 0
52.5 4
};
\addplot [black, dashed, forget plot]
table {%
63 0
63 4
};
\draw (axis cs:17.5,-0.0) node[
  scale=0.9,
  anchor=base west,
  text=black,
  rotate=0.0
]{A};
\draw (axis cs:35,-0.0) node[
  scale=0.9,
  anchor=base west,
  text=black,
  rotate=0.0
]{B};
\draw (axis cs:52.5,-0.0) node[
  scale=0.9,
  anchor=base west,
  text=black,
  rotate=0.0
]{C};
\draw (axis cs:63,-0.0) node[
  scale=0.9,
  anchor=base west,
  text=black,
  rotate=0.0
]{D};
\end{axis}


\begin{axis}[name=two,
width=0.22\linewidth, height=0.22\linewidth, scale only axis,enlargelimits=false,
at=(one.north east), anchor=north west,
yticklabels={,,}, xticklabels={,,},
tick style={draw=none}
]
\addplot graphics [includegraphics, xmin=0, xmax=1023.5, ymin=1023.5, ymax=0] {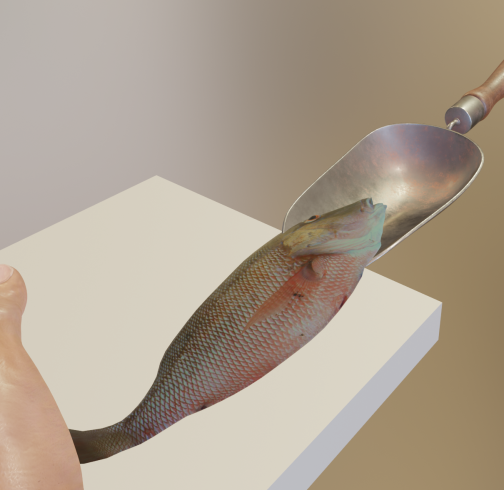};
\draw (axis cs:0,-0.1) node[
  scale=0.9,
  anchor=base west,
  text=black,
  rotate=0.0
]{A};
\end{axis}

\begin{axis}[name=three,
width=0.22\linewidth, height=0.22\linewidth, scale only axis,enlargelimits=false,
at=(two.north east), anchor=north west,
yticklabel pos=right,
yticklabels={,,}, xticklabels={,,},
tick style={draw=none},
]
\addplot graphics [includegraphics cmd=\pgfimage,xmin=-0.5, xmax=1023.5, ymin=1023.5, ymax=-0.5] {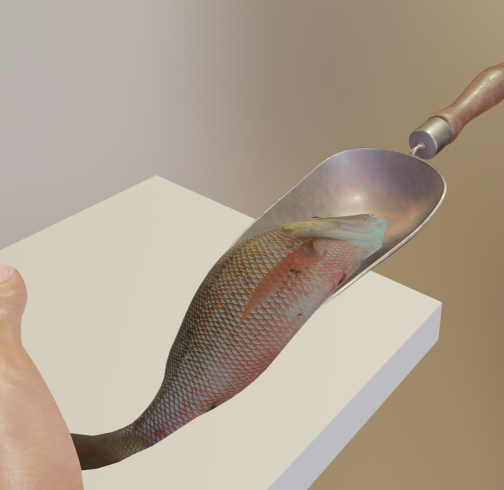};
\draw (axis cs:0,-0.1) node[
  scale=0.9,
  anchor=base west,
  text=black,
  rotate=0.0
]{B};
\end{axis}

\begin{axis}[name=four,
width=0.22\linewidth, height=0.22\linewidth, scale only axis,enlargelimits=false,
at=(one.east), anchor=north west,
yticklabel pos=right,
yticklabels={,,}, xticklabels={,,},
tick style={draw=none},
]
\addplot graphics [includegraphics cmd=\pgfimage,xmin=-0.5, xmax=1023.5, ymin=1023.5, ymax=-0.5] {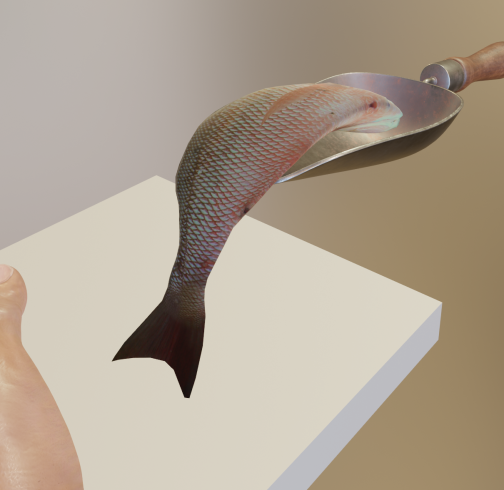};
\draw (axis cs:0,-0.1) node[
  scale=0.9,
  anchor=base west,
  text=black,
  rotate=0.0
]{C};
\end{axis}

\begin{axis}[name=five,
width=0.22\linewidth, height=0.22\linewidth, scale only axis,enlargelimits=false,
at=(four.south east), anchor=south west,
yticklabel pos=right,
yticklabels={,,}, xticklabels={,,},
tick style={draw=none},
]
\addplot graphics [includegraphics cmd=\pgfimage,xmin=-0.5, xmax=1023.5, ymin=1023.5, ymax=-0.5] {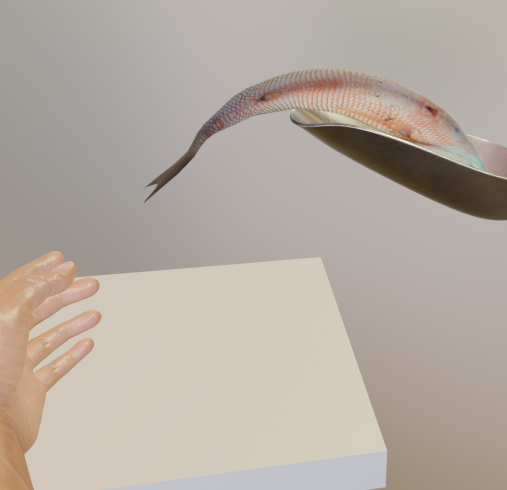};
\draw (axis cs:0,-0.1) node[
  scale=0.9,
  anchor=base west,
  text=black,
  rotate=0.0
]{D};
\end{axis}

\end{tikzpicture}

%% file: icra2024/tex/phys-experiment/side-by-side-plots.tex
\definecolor{darkgray176}{RGB}{100,100,100}
\definecolor{lightgray204}{RGB}{204,204,204}
\definecolor{sandybrown24416662}{RGB}{244,166,62}
\definecolor{seagreen4916384}{RGB}{49,163,84}
\definecolor{slateblue117107177}{RGB}{117,107,177}
\definecolor{steelblue43140190}{RGB}{43,140,190}

\begin{tikzpicture}[font=\footnotesize,
  dia/.style={draw,diamond, minimum size=1mm,inner sep=1.5pt},
  rec/.style={draw,rectangle,minimum size=1mm,inner sep=2pt},
  cir/.style={draw,regular polygon, regular polygon sides=3, minimum size=1mm,inner sep=1pt},
  cirL/.style={draw,circle, minimum size=2mm,inner sep=2pt},
  labelOpacity/.style={opacity=1}
]

\begin{axis}[name=zero,
width=0.48\linewidth, height=0.225\linewidth, scale only axis,
yticklabels={,,}, xticklabels={,,}, enlargelimits=false,
tick style={draw=none}
]
\addplot graphics [includegraphics cmd=\pgfimage,xmin=-0.5, xmax=1023.5, ymin=1023.5, ymax=-0.5] {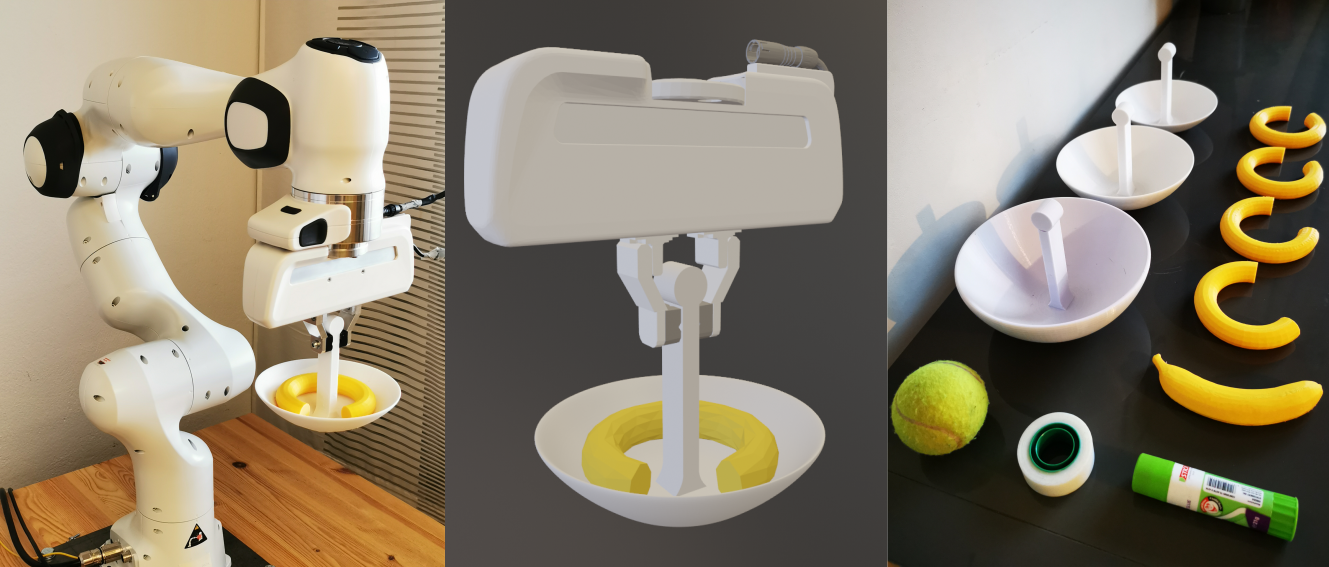};
\draw (axis cs:250,50) node[
  scale=1,
  anchor=base west,
  text=white,
  rotate=0.0
]{(i-1)};
\draw (axis cs:580,50) node[
  scale=1,
  anchor=base west,
  text=white,
  rotate=0.0
]{(i-2)};
\draw (axis cs:690,50) node[
  scale=1,
  anchor=base west,
  text=white,
  rotate=0.0
]{(i-3)};
\end{axis}

\begin{axis}[
name=one,
width = .27\linewidth,
height = .2\linewidth,
scale only axis,
at={($(zero.east)+(.05\linewidth,-.089\linewidth)$)},
tick align=inside,
tick pos=left,
x grid style={darkgray176},
xlabel={normalized escape energy $\overline{e}$ / dm},
x label style={at={(axis description cs:0.5,-0.05)}, anchor=north},
xmin=0.03, xmax=0.58,
xtick style={color=black},
y grid style={darkgray176},
ylabel={perturbation level $\tau$},
y label style={at={(axis description cs:-0.08,.5)},anchor=south},
ymin=0.3, ymax=0.92,
ytick style={color=black}
]


\draw[draw=seagreen4916384,fill=seagreen4916384,opacity=0.3] (axis cs:0.1697260452,0.41375) ellipse (0.001479323377 and 0.05423164601);
\node [rec] at (0.1697260452,0.41375) {};

\draw[draw=seagreen4916384,fill=seagreen4916384,opacity=0.3] (axis cs:0.3342074684,0.63125) ellipse (0.001297597143 and 0.08322731351);
\node [cir] at (0.3342074684,0.63125) {};

\draw[draw=seagreen4916384,fill=seagreen4916384,opacity=0.3] (axis cs:0.4599809861,0.7925) ellipse (0.001244198743 and 0.05946187254);
\node [dia] at (0.4599809861,0.7925) {};

\draw[draw=slateblue117107177,fill=slateblue117107177,opacity=0.3] (axis cs:0.09200999502,0.35375) ellipse (0.02601387858 and 0.04068608046);
\node [rec] at (0.09200999502,0.35375) {};

\draw[draw=slateblue117107177,fill=slateblue117107177,opacity=0.3] (axis cs:0.1500594512,0.46625) ellipse (0.004287236082 and 0.08991066995);
\node [cir] at (0.1500594512,0.46625) {};

\draw[draw=slateblue117107177,fill=slateblue117107177,opacity=0.3] (axis cs:0.2524673438,0.6425) ellipse (0.00276152771 and 0.06755949759);
\node [dia] at (0.2524673438,0.6425) {};

\draw[draw=steelblue43140190,fill=steelblue43140190,opacity=0.3] (axis cs:0.2809728784,0.43625) ellipse (0.0140091211 and 0.06696214282);
\node [rec] at (0.2809728784,0.43625) {};

\draw[draw=steelblue43140190,fill=steelblue43140190,opacity=0.3] (axis cs:0.3879901447,0.48125) ellipse (0.0107376311 and 0.0716015762);
\node [cir] at (0.3879901447,0.48125) {};

\draw[draw=steelblue43140190,fill=steelblue43140190,opacity=0.3] (axis cs:0.5039077638,0.68) ellipse (0.01129383459 and 0.09071147352);
\node [dia] at (0.5039077638,0.68) {};

\draw[draw=sandybrown24416662,fill=sandybrown24416662,opacity=0.3] (axis cs:0.2578059871,0.51875) ellipse (0.007770725066 and 0.08166788143);
\node [rec] at (0.2578059871,0.51875) {};

\draw[draw=sandybrown24416662,fill=sandybrown24416662,opacity=0.3] (axis cs:0.3836056399,0.62375) ellipse (0.0114331104 and 0.07595816518);
\node [cir] at (0.3836056399,0.62375) {};

\draw[draw=sandybrown24416662,fill=sandybrown24416662,opacity=0.3] (axis cs:0.5369652167,0.8075) ellipse (0.008659683695 and 0.09721111048);
\node [dia] at (0.5369652167,0.8075) {};

\matrix [draw,below left,color=darkgray176] at (0.21,0.91) {
  \node [cirL, draw=sandybrown24416662, fill=sandybrown24416662, opacity=0.5, label=right:tape] {}; \\
  \node [cirL, draw=steelblue43140190, fill=steelblue43140190, opacity=0.5, label=right:banana] {}; \\
  \node [cirL, draw=seagreen4916384, fill=seagreen4916384, opacity=0.5, label=right:ball] {}; \\
  \node [cirL, draw=slateblue117107177, fill=slateblue117107177, opacity=0.5, label=right:glue] {}; \\
};
\draw (axis cs:0.51,0.34) node[
  scale=1,
  anchor=base west,
  text=red,
  rotate=0.0
]{(ii)};

\end{axis}

\begin{axis}[
name=two,
width = .14\linewidth,
height = .2\linewidth,
scale only axis,
at={($(one.east)+(.03\linewidth,-.1\linewidth)$)},
legend cell align={left},
legend style={fill opacity=0.8, draw opacity=1, text opacity=1, draw=lightgray204},
tick align=inside,
tick pos=left,
x grid style={darkgray176},
xlabel={$\overline{e}$ / dm},
x label style={at={(axis description cs:0.5,-0.05)}, anchor=north},
xmin=0.24, xmax=0.36,
xtick style={color=black},
xtick={0.24,0.28,0.32,0.36},
y grid style={darkgray176},
ymin=0.45, ymax=0.96,
ytick style={color=black}
]
\draw[draw=seagreen4916384,fill=seagreen4916384,opacity=0.3] (axis cs:0.2831818684,0.53375) ellipse (0.01666203107 and 0.07069805008);
\node [rec] at (0.2831818684,0.53375) {};

\draw[draw=slateblue117107177,fill=slateblue117107177,opacity=0.3] (axis cs:0.3078113936,0.5825) ellipse (0.02364518268 and 0.1060660172);
\node [rec] at (0.3078113936,0.5825) {};

\draw[draw=sandybrown24416662,fill=sandybrown24416662,opacity=0.3] (axis cs:0.3304189876,0.6875) ellipse (0.03004986464 and 0.1432031524);
\node [rec] at (0.3304189876,0.6875) {};

\draw[draw=steelblue43140190,fill=steelblue43140190,opacity=0.3] (axis cs:0.322336502633477,0.63125) ellipse (0.01723913978 and 0.09203066259);
\node [rec] at (0.322336502633477,0.63125) {};

\matrix [draw,below left,color=darkgray176] at (0.3,0.955) {
  \node [cirL, draw=sandybrown24416662, fill=sandybrown24416662, opacity=0.5, label=right:45] {}; \\
  \node [cirL, draw=steelblue43140190, fill=steelblue43140190, opacity=0.5, label={right:60}] {}; \\
  \node [cirL, draw=slateblue117107177, fill=slateblue117107177, opacity=0.5, label=right:90] {}; \\
  \node [cirL, draw=seagreen4916384, fill=seagreen4916384, opacity=0.5, label=right:120] {}; \\
};
\draw (axis cs:0.33,0.48) node[
  scale=1,
  anchor=base west,
  text=red,
  rotate=0.0
]{(iii)};
\end{axis}

\end{tikzpicture}

%% file: icra2024/tex/appendix/varying-obstacle-spaces/gamma-0.0-0.tex
\begin{tikzpicture}

\definecolor{sandybrown24416662}{RGB}{244,166,62}
\definecolor{seagreen4916384}{RGB}{49,163,84}

\begin{axis}[
width=0.6\linewidth, 
height=0.6\linewidth,  
tick pos=left,
xmin=0, xmax=1,
ymin=0, ymax=1,
xtick=\empty,
ytick=\empty,
xticklabels={},
yticklabels={},
]
\path [draw=sandybrown24416662, fill=sandybrown24416662, opacity=0.7]
(axis cs:0.251270777971036,0)
--(axis cs:0.251108053629402,0.0240799257265772)
--(axis cs:0.25062053583799,0.0480628900911132)
--(axis cs:0.249810187658881,0.0718523221607498)
--(axis cs:0.248680272078184,0.0953524302888745)
--(axis cs:0.247235338867147,0.118468587834272)
--(axis cs:0.245481206261829,0.141107714189211)
--(axis cs:0.243424937535122,0.163178649582198)
--(axis cs:0.241074812555442,0.184592522146205)
--(axis cs:0.238440294446619,0.20526310577431)
--(axis cs:0.235531991483241,0.225107167321796)
--(axis cs:0.232361614374875,0.244044801756649)
--(axis cs:0.228941929111171,0.261999753908921)
--(axis cs:0.225286705557728,0.278899725523381)
--(axis cs:0.221410662009709,0.294676666379079)
--(axis cs:0.217329405926456,0.309267048303562)
--(axis cs:0.213059371085763,0.322612120978404)
--(axis cs:0.208617751410849,0.334658148506004)
--(axis cs:0.204022431736506,0.345356625785078)
--(axis cs:0.19929191579317,0.354664473823585)
--(axis cs:0.194445251698942,0.36254421320262)
--(axis cs:0.189501955259526,0.368964114992809)
--(axis cs:0.184481931384981,0.373898328515508)
--(axis cs:0.179405393939662,0.37732698543436)
--(axis cs:0.174292784348129,0.379236279758076)
--(axis cs:0.169164689284744,0.379618523432288)
--(axis cs:0.164041757778398,0.37847217729663)
--(axis cs:0.15894461806616,0.375801857282396)
--(axis cs:0.153893794530648,0.371618315825809)
--(axis cs:0.148909625055582,0.365938398571761)
--(axis cs:0.144012179132306,0.358784976542341)
--(axis cs:0.139221177047032,0.350186854043308)
--(axis cs:0.134555910474202,0.34017865267931)
--(axis cs:0.130035164795728,0.328800671944908)
--(axis cs:0.125677143458889,0.316098726952733)
--(axis cs:0.12149939467747,0.302123963952195)
--(axis cs:0.117518740771305,0.286932654381592)
--(axis cs:0.113751210428727,0.270585968282886)
--(axis cs:0.110211974164699,0.253149727991537)
--(axis cs:0.106915283234504,0.234694143093185)
--(axis cs:0.103874412248973,0.215293527714432)
--(axis cs:0.101101605722307,0.195026001286082)
--(axis cs:0.0986080287677447,0.173973173983765)
--(axis cs:0.0964037221395894,0.152219818112563)
--(axis cs:0.0944975618026366,0.129853526758858)
--(axis cs:0.0928972231917956,0.106964361083889)
--(axis cs:0.0916091503058229,0.0836444876792389)
--(axis cs:0.0906385297596129,0.0599878074445054)
--(axis cs:0.0899892698995307,0.0360895774815158)
--(axis cs:0.0896639850658807,0.0120460275275997)
--(axis cs:0.0896639850658807,-0.0120460275275998)
--(axis cs:0.0899892698995307,-0.0360895774815159)
--(axis cs:0.0906385297596129,-0.0599878074445055)
--(axis cs:0.0916091503058229,-0.0836444876792388)
--(axis cs:0.0928972231917956,-0.106964361083889)
--(axis cs:0.0944975618026366,-0.129853526758858)
--(axis cs:0.0964037221395894,-0.152219818112563)
--(axis cs:0.0986080287677447,-0.173973173983765)
--(axis cs:0.101101605722307,-0.195026001286082)
--(axis cs:0.103874412248973,-0.215293527714432)
--(axis cs:0.106915283234504,-0.234694143093185)
--(axis cs:0.110211974164699,-0.253149727991537)
--(axis cs:0.113751210428727,-0.270585968282886)
--(axis cs:0.117518740771305,-0.286932654381591)
--(axis cs:0.12149939467747,-0.302123963952195)
--(axis cs:0.125677143458889,-0.316098726952733)
--(axis cs:0.130035164795728,-0.328800671944908)
--(axis cs:0.134555910474202,-0.34017865267931)
--(axis cs:0.139221177047032,-0.350186854043308)
--(axis cs:0.144012179132306,-0.358784976542341)
--(axis cs:0.148909625055582,-0.365938398571761)
--(axis cs:0.153893794530648,-0.371618315825809)
--(axis cs:0.15894461806616,-0.375801857282396)
--(axis cs:0.164041757778398,-0.37847217729663)
--(axis cs:0.169164689284744,-0.379618523432288)
--(axis cs:0.174292784348129,-0.379236279758076)
--(axis cs:0.179405393939662,-0.37732698543436)
--(axis cs:0.184481931384981,-0.373898328515508)
--(axis cs:0.189501955259526,-0.368964114992809)
--(axis cs:0.194445251698942,-0.36254421320262)
--(axis cs:0.19929191579317,-0.354664473823585)
--(axis cs:0.204022431736506,-0.345356625785078)
--(axis cs:0.208617751410849,-0.334658148506004)
--(axis cs:0.213059371085763,-0.322612120978404)
--(axis cs:0.217329405926456,-0.309267048303562)
--(axis cs:0.221410662009709,-0.294676666379079)
--(axis cs:0.225286705557728,-0.278899725523381)
--(axis cs:0.228941929111171,-0.261999753908921)
--(axis cs:0.232361614374875,-0.24404480175665)
--(axis cs:0.235531991483241,-0.225107167321796)
--(axis cs:0.238440294446619,-0.20526310577431)
--(axis cs:0.241074812555442,-0.184592522146205)
--(axis cs:0.243424937535122,-0.163178649582198)
--(axis cs:0.245481206261829,-0.141107714189211)
--(axis cs:0.247235338867147,-0.118468587834272)
--(axis cs:0.248680272078184,-0.0953524302888746)
--(axis cs:0.249810187658881,-0.0718523221607496)
--(axis cs:0.25062053583799,-0.0480628900911131)
--(axis cs:0.251108053629402,-0.0240799257265772)
--(axis cs:0.251270777971036,-9.29914269709016e-17)
--cycle;
\path [draw=sandybrown24416662, fill=sandybrown24416662, opacity=0.7]
(axis cs:0.466663834563755,0.405336756767039)
--(axis cs:0.466530244300152,0.435699156388856)
--(axis cs:0.466130011430184,0.465939297343616)
--(axis cs:0.46546474755036,0.495935413256764)
--(axis cs:0.464537131443522,0.525566720356458)
--(axis cs:0.463350898292337,0.554713903827195)
--(axis cs:0.46191082463903,0.583259598248453)
--(axis cs:0.460222709151939,0.611088860183824)
--(axis cs:0.45829334927632,0.638089631017658)
--(axis cs:0.456130513863436,0.664153188175562)
--(axis cs:0.453742911888137,0.689174582911831)
--(axis cs:0.451140157380887,0.713053062901006)
--(axis cs:0.448332730715462,0.735692477931933)
--(axis cs:0.445331936408178,0.757001667070722)
--(axis cs:0.442149857598601,0.776894825733657)
--(axis cs:0.438799307395,0.795291851191968)
--(axis cs:0.435293777280491,0.812118665117251)
--(axis cs:0.431647382787589,0.827307511868747)
--(axis cs:0.427874806659945,0.840797231321398)
--(axis cs:0.423991239730117,0.852533505136079)
--(axis cs:0.420012319751453,0.862469075480384)
--(axis cs:0.415954068430369,0.870563935319233)
--(axis cs:0.411832826912603,0.876785489509087)
--(axis cs:0.40766518998318,0.88110868604708)
--(axis cs:0.403467939245072,0.883516116946601)
--(axis cs:0.399257975545609,0.883998088333105)
--(axis cs:0.395052250922728,0.882552659477926)
--(axis cs:0.390867700345101,0.87918565061291)
--(axis cs:0.386721173520993,0.87391061949439)
--(axis cs:0.382629367050434,0.866748806810893)
--(axis cs:0.378608757193904,0.857729050654377)
--(axis cs:0.374675533528245,0.846887670399418)
--(axis cs:0.370845533756952,0.834268320457905)
--(axis cs:0.36713417993733,0.819921814498131)
--(axis cs:0.363556416381311,0.803905920836087)
--(axis cs:0.360126649479985,0.78628512982284)
--(axis cs:0.35685868969415,0.767130394164652)
--(axis cs:0.353765695944458,0.746518843221473)
--(axis cs:0.350860122625084,0.724533472434234)
--(axis cs:0.348153669454283,0.701262809131495)
--(axis cs:0.345657234363748,0.676800556061141)
--(axis cs:0.343380869616486,0.651245214082485)
--(axis cs:0.341333741329906,0.624699685538078)
--(axis cs:0.339524092567097,0.597270859902288)
--(axis cs:0.337959210144924,0.569069183375119)
--(axis cs:0.336645395292583,0.540208214154339)
--(axis cs:0.335587938278783,0.510804165176701)
--(axis cs:0.334791097109696,0.480975436169461)
--(axis cs:0.33425808038347,0.450842136896468)
--(axis cs:0.333991034370335,0.420525603518535)
--(axis cs:0.333991034370335,0.390147910015542)
--(axis cs:0.33425808038347,0.359831376637609)
--(axis cs:0.334791097109696,0.329698077364616)
--(axis cs:0.335587938278783,0.299869348357377)
--(axis cs:0.336645395292583,0.270465299379738)
--(axis cs:0.337959210144924,0.241604330158958)
--(axis cs:0.339524092567097,0.213402653631789)
--(axis cs:0.341333741329906,0.185973827996)
--(axis cs:0.343380869616486,0.159428299451592)
--(axis cs:0.345657234363748,0.133872957472937)
--(axis cs:0.348153669454283,0.109410704402582)
--(axis cs:0.350860122625084,0.0861400410998431)
--(axis cs:0.353765695944458,0.0641546703126039)
--(axis cs:0.35685868969415,0.0435431193694253)
--(axis cs:0.360126649479985,0.024388383711237)
--(axis cs:0.363556416381311,0.00676759269798993)
--(axis cs:0.36713417993733,-0.0092483009640541)
--(axis cs:0.370845533756952,-0.0235948069238276)
--(axis cs:0.374675533528245,-0.0362141568653407)
--(axis cs:0.378608757193904,-0.0470555371202997)
--(axis cs:0.382629367050434,-0.0560752932768158)
--(axis cs:0.386721173520993,-0.063237105960313)
--(axis cs:0.390867700345101,-0.0685121370788322)
--(axis cs:0.395052250922728,-0.0718791459438485)
--(axis cs:0.399257975545609,-0.0733245747990274)
--(axis cs:0.403467939245072,-0.072842603412524)
--(axis cs:0.40766518998318,-0.0704351725130031)
--(axis cs:0.411832826912603,-0.0661119759750093)
--(axis cs:0.415954068430369,-0.0598904217851557)
--(axis cs:0.420012319751453,-0.0517955619463062)
--(axis cs:0.423991239730117,-0.0418599916020015)
--(axis cs:0.427874806659945,-0.0301237177873203)
--(axis cs:0.431647382787589,-0.0166339983346699)
--(axis cs:0.435293777280491,-0.00144515158317349)
--(axis cs:0.438799307395,0.0153816623421094)
--(axis cs:0.442149857598601,0.033778687800421)
--(axis cs:0.445331936408178,0.0536718464633557)
--(axis cs:0.448332730715462,0.0749810356021446)
--(axis cs:0.451140157380887,0.0976204506330712)
--(axis cs:0.453742911888137,0.121498930622247)
--(axis cs:0.456130513863436,0.146520325358515)
--(axis cs:0.45829334927632,0.172583882516419)
--(axis cs:0.460222709151939,0.199584653350253)
--(axis cs:0.46191082463903,0.227413915285624)
--(axis cs:0.463350898292337,0.255959609706883)
--(axis cs:0.464537131443522,0.285106793177619)
--(axis cs:0.46546474755036,0.314738100277314)
--(axis cs:0.466130011430184,0.344734216190461)
--(axis cs:0.466530244300152,0.374974357145221)
--(axis cs:0.466663834563755,0.405336756767039)
--cycle;
\path [draw=sandybrown24416662, fill=sandybrown24416662, opacity=0.7]
(axis cs:0.91173274368321,0.0654475360020926)
--(axis cs:0.911603818112721,0.0991395183592101)
--(axis cs:0.911217560539023,0.132695834994751)
--(axis cs:0.910575526285038,0.165981366465116)
--(axis cs:0.909680300596104,0.198862083682996)
--(axis cs:0.908535488230107,0.231205587605198)
--(axis cs:0.90714569894239,0.262881642356865)
--(axis cs:0.905516528923898,0.29376269964538)
--(axis cs:0.903654538267289,0.323724412352317)
--(axis cs:0.90156722455176,0.352646135235406)
--(axis cs:0.899262992652938,0.380411410724339)
--(axis cs:0.896751120899416,0.406908437854302)
--(axis cs:0.894041723712196,0.432030522449004)
--(axis cs:0.891145710877486,0.455676506740465)
--(axis cs:0.888074743616842,0.477751176695646)
--(axis cs:0.884841187631543,0.498165645409755)
--(axis cs:0.881458063310279,0.51683771102244)
--(axis cs:0.877938993300639,0.533692187715667)
--(axis cs:0.874298147655518,0.548661208460476)
--(axis cs:0.870550186775314,0.56168449829355)
--(axis cs:0.866710202375675,0.572709617023219)
--(axis cs:0.862793656718477,0.581692170387608)
--(axis cs:0.858816320350762,0.588595988814652)
--(axis cs:0.854794208602307,0.593393273064207)
--(axis cs:0.850743517097543,0.596064706165771)
--(axis cs:0.846680556541495,0.59659953120111)
--(axis cs:0.842621687042324,0.594995594618569)
--(axis cs:0.83858325223495,0.591259354904667)
--(axis cs:0.834581513471004,0.58540585657805)
--(axis cs:0.830632584340099,0.577458669610523)
--(axis cs:0.826752365786094,0.567449794519093)
--(axis cs:0.822956482079608,0.555419533511185)
--(axis cs:0.819260217904593,0.541416328201871)
--(axis cs:0.815678456812309,0.525496564556587)
--(axis cs:0.812225621290514,0.507724345844753)
--(axis cs:0.808915614689197,0.488171234518527)
--(axis cs:0.805761765236696,0.46691596405607)
--(axis cs:0.802776772371625,0.44404412192962)
--(axis cs:0.799972655606716,0.419647804974945)
--(axis cs:0.797360706130487,0.393825248549895)
--(axis cs:0.794951441341606,0.366680430975282)
--(axis cs:0.792754562499044,0.338322654850885)
--(axis cs:0.790778915658525,0.308866106932458)
--(axis cs:0.789032456052583,0.278429398341961)
--(axis cs:0.78752221605765,0.247135086962443)
--(axis cs:0.786254276877155,0.215109183940696)
--(axis cs:0.785233744054667,0.182480646284847)
--(axis cs:0.784464726915671,0.149380857599996)
--(axis cs:0.783950322020767,0.115943099052813)
--(axis cs:0.78369260069691,0.08230201269532)
--(axis cs:0.78369260069691,0.0485930593088652)
--(axis cs:0.783950322020767,0.0149519729513717)
--(axis cs:0.784464726915671,-0.0184857855958112)
--(axis cs:0.785233744054667,-0.0515855742806618)
--(axis cs:0.786254276877155,-0.0842141119365111)
--(axis cs:0.78752221605765,-0.116240014958257)
--(axis cs:0.789032456052583,-0.147534326337776)
--(axis cs:0.790778915658525,-0.177971034928272)
--(axis cs:0.792754562499044,-0.2074275828467)
--(axis cs:0.794951441341606,-0.235785358971097)
--(axis cs:0.797360706130487,-0.26293017654571)
--(axis cs:0.799972655606716,-0.28875273297076)
--(axis cs:0.802776772371625,-0.313149049925434)
--(axis cs:0.805761765236696,-0.336020892051885)
--(axis cs:0.808915614689197,-0.357276162514341)
--(axis cs:0.812225621290514,-0.376829273840568)
--(axis cs:0.815678456812309,-0.394601492552402)
--(axis cs:0.819260217904593,-0.410521256197685)
--(axis cs:0.822956482079608,-0.424524461507)
--(axis cs:0.826752365786094,-0.436554722514908)
--(axis cs:0.830632584340099,-0.446563597606337)
--(axis cs:0.834581513471004,-0.454510784573865)
--(axis cs:0.83858325223495,-0.460364282900482)
--(axis cs:0.842621687042324,-0.464100522614384)
--(axis cs:0.846680556541495,-0.465704459196925)
--(axis cs:0.850743517097543,-0.465169634161586)
--(axis cs:0.854794208602307,-0.462498201060022)
--(axis cs:0.858816320350762,-0.457700916810467)
--(axis cs:0.862793656718477,-0.450797098383422)
--(axis cs:0.866710202375675,-0.441814545019034)
--(axis cs:0.870550186775314,-0.430789426289364)
--(axis cs:0.874298147655518,-0.417766136456291)
--(axis cs:0.877938993300639,-0.402797115711482)
--(axis cs:0.881458063310279,-0.385942639018254)
--(axis cs:0.884841187631543,-0.36727057340557)
--(axis cs:0.888074743616842,-0.346856104691461)
--(axis cs:0.891145710877486,-0.32478143473628)
--(axis cs:0.894041723712196,-0.301135450444819)
--(axis cs:0.896751120899416,-0.276013365850117)
--(axis cs:0.899262992652938,-0.249516338720153)
--(axis cs:0.90156722455176,-0.221751063231221)
--(axis cs:0.903654538267289,-0.192829340348132)
--(axis cs:0.905516528923898,-0.162867627641195)
--(axis cs:0.90714569894239,-0.13198657035268)
--(axis cs:0.908535488230107,-0.100310515601012)
--(axis cs:0.909680300596104,-0.0679670116788104)
--(axis cs:0.910575526285038,-0.0350862944609304)
--(axis cs:0.911217560539023,-0.00180076299056529)
--(axis cs:0.911603818112721,0.0317555536449752)
--(axis cs:0.91173274368321,0.0654475360020925)
--cycle;
\addplot [semithick, seagreen4916384]
table {%
0 0
0.0778870023732268 0.00666794757527299
0.0807014309598316 0.0636549848960973
0.0904097225721177 0.113759655389135
0.0787980650021983 0.171929706033058
0.0746123071351113 0.206188387843743
0.102990051299115 0.248007014020168
0.110217169778164 0.298943307332903
0.120581260684259 0.362033220306615
0.153162641578227 0.38712897616678
0.186690301852034 0.423109802506518
0.230603986803209 0.437693651052566
0.263792406457923 0.465102903296372
0.313136447725651 0.478287496100314
0.320928291932746 0.533241454479968
0.304174972978193 0.584427846523702
0.313806958466127 0.641727380762031
0.338556631016509 0.66695594872156
0.328140243946688 0.716450509620737
0.33116243838952 0.770250110838058
0.336544746262371 0.812953126416446
0.367712950243834 0.855290551661422
0.39798071249994 0.902667739047231
0.415945628999324 0.899427536205647
0.425200394173346 0.85422671583444
0.440635562246462 0.80117346578153
0.467086053041925 0.757343778008465
0.476044592505197 0.72605238854091
0.488890406072036 0.684395833761384
0.530988550650457 0.690357050015447
0.567555463503873 0.676976428919826
0.617058390914183 0.654710960880697
0.668329765513102 0.642122906891937
0.707652518776293 0.638191495083538
0.758995359575643 0.637522430629099
0.796335872874243 0.609431135578555
0.836748145149773 0.592941860222218
0.869507092009435 0.606658501168502
0.899107954815953 0.581809715940789
0.890954323231005 0.544819393372776
0.891428392737914 0.500094200105633
0.905586717594952 0.443856793263453
0.910381034906306 0.404513203314496
0.906105584780952 0.355401644715469
0.92249469126329 0.302421652494932
0.910945470418751 0.24352994956759
0.927369045850556 0.188659708696318
0.949852592149132 0.155240987164844
0.993293764327773 0.132399855606185
0.986982148679319 0.0726536715756888
0.987442624618232 0.0330649942331653
1 0
};
\addplot [semithick, sandybrown24416662, opacity=0.7]
table {%
0.251270777971036 0
0.251108053629402 0.0240799257265772
0.25062053583799 0.0480628900911132
0.249810187658881 0.0718523221607498
0.248680272078184 0.0953524302888745
0.247235338867147 0.118468587834272
0.245481206261829 0.141107714189211
0.243424937535122 0.163178649582198
0.241074812555442 0.184592522146205
0.238440294446619 0.20526310577431
0.235531991483241 0.225107167321796
0.232361614374875 0.244044801756649
0.228941929111171 0.261999753908921
0.225286705557728 0.278899725523381
0.221410662009709 0.294676666379079
0.217329405926456 0.309267048303562
0.213059371085763 0.322612120978404
0.208617751410849 0.334658148506004
0.204022431736506 0.345356625785078
0.19929191579317 0.354664473823585
0.194445251698942 0.36254421320262
0.189501955259526 0.368964114992809
0.184481931384981 0.373898328515508
0.179405393939662 0.37732698543436
0.174292784348129 0.379236279758076
0.169164689284744 0.379618523432288
0.164041757778398 0.37847217729663
0.15894461806616 0.375801857282396
0.153893794530648 0.371618315825809
0.148909625055582 0.365938398571761
0.144012179132306 0.358784976542341
0.139221177047032 0.350186854043308
0.134555910474202 0.34017865267931
0.130035164795728 0.328800671944908
0.125677143458889 0.316098726952733
0.12149939467747 0.302123963952195
0.117518740771305 0.286932654381592
0.113751210428727 0.270585968282886
0.110211974164699 0.253149727991537
0.106915283234504 0.234694143093185
0.103874412248973 0.215293527714432
0.101101605722307 0.195026001286082
0.0986080287677447 0.173973173983765
0.0964037221395894 0.152219818112563
0.0944975618026366 0.129853526758858
0.0928972231917956 0.106964361083889
0.0916091503058229 0.0836444876792389
0.0906385297596129 0.0599878074445054
0.0899892698995307 0.0360895774815158
0.0896639850658807 0.0120460275275997
0.0896639850658807 -0.0120460275275998
0.0899892698995307 -0.0360895774815159
0.0906385297596129 -0.0599878074445055
0.0916091503058229 -0.0836444876792388
0.0928972231917956 -0.106964361083889
0.0944975618026366 -0.129853526758858
0.0964037221395894 -0.152219818112563
0.0986080287677447 -0.173973173983765
0.101101605722307 -0.195026001286082
0.103874412248973 -0.215293527714432
0.106915283234504 -0.234694143093185
0.110211974164699 -0.253149727991537
0.113751210428727 -0.270585968282886
0.117518740771305 -0.286932654381591
0.12149939467747 -0.302123963952195
0.125677143458889 -0.316098726952733
0.130035164795728 -0.328800671944908
0.134555910474202 -0.34017865267931
0.139221177047032 -0.350186854043308
0.144012179132306 -0.358784976542341
0.148909625055582 -0.365938398571761
0.153893794530648 -0.371618315825809
0.15894461806616 -0.375801857282396
0.164041757778398 -0.37847217729663
0.169164689284744 -0.379618523432288
0.174292784348129 -0.379236279758076
0.179405393939662 -0.37732698543436
0.184481931384981 -0.373898328515508
0.189501955259526 -0.368964114992809
0.194445251698942 -0.36254421320262
0.19929191579317 -0.354664473823585
0.204022431736506 -0.345356625785078
0.208617751410849 -0.334658148506004
0.213059371085763 -0.322612120978404
0.217329405926456 -0.309267048303562
0.221410662009709 -0.294676666379079
0.225286705557728 -0.278899725523381
0.228941929111171 -0.261999753908921
0.232361614374875 -0.24404480175665
0.235531991483241 -0.225107167321796
0.238440294446619 -0.20526310577431
0.241074812555442 -0.184592522146205
0.243424937535122 -0.163178649582198
0.245481206261829 -0.141107714189211
0.247235338867147 -0.118468587834272
0.248680272078184 -0.0953524302888746
0.249810187658881 -0.0718523221607496
0.25062053583799 -0.0480628900911131
0.251108053629402 -0.0240799257265772
0.251270777971036 -9.29914269709016e-17
};
\addplot [semithick, sandybrown24416662, opacity=0.7]
table {%
0.466663834563755 0.405336756767039
0.466530244300152 0.435699156388856
0.466130011430184 0.465939297343616
0.46546474755036 0.495935413256764
0.464537131443522 0.525566720356458
0.463350898292337 0.554713903827195
0.46191082463903 0.583259598248453
0.460222709151939 0.611088860183824
0.45829334927632 0.638089631017658
0.456130513863436 0.664153188175562
0.453742911888137 0.689174582911831
0.451140157380887 0.713053062901006
0.448332730715462 0.735692477931933
0.445331936408178 0.757001667070722
0.442149857598601 0.776894825733657
0.438799307395 0.795291851191968
0.435293777280491 0.812118665117251
0.431647382787589 0.827307511868747
0.427874806659945 0.840797231321398
0.423991239730117 0.852533505136079
0.420012319751453 0.862469075480384
0.415954068430369 0.870563935319233
0.411832826912603 0.876785489509087
0.40766518998318 0.88110868604708
0.403467939245072 0.883516116946601
0.399257975545609 0.883998088333105
0.395052250922728 0.882552659477926
0.390867700345101 0.87918565061291
0.386721173520993 0.87391061949439
0.382629367050434 0.866748806810893
0.378608757193904 0.857729050654377
0.374675533528245 0.846887670399418
0.370845533756952 0.834268320457905
0.36713417993733 0.819921814498131
0.363556416381311 0.803905920836087
0.360126649479985 0.78628512982284
0.35685868969415 0.767130394164652
0.353765695944458 0.746518843221473
0.350860122625084 0.724533472434234
0.348153669454283 0.701262809131495
0.345657234363748 0.676800556061141
0.343380869616486 0.651245214082485
0.341333741329906 0.624699685538078
0.339524092567097 0.597270859902288
0.337959210144924 0.569069183375119
0.336645395292583 0.540208214154339
0.335587938278783 0.510804165176701
0.334791097109696 0.480975436169461
0.33425808038347 0.450842136896468
0.333991034370335 0.420525603518535
0.333991034370335 0.390147910015542
0.33425808038347 0.359831376637609
0.334791097109696 0.329698077364616
0.335587938278783 0.299869348357377
0.336645395292583 0.270465299379738
0.337959210144924 0.241604330158958
0.339524092567097 0.213402653631789
0.341333741329906 0.185973827996
0.343380869616486 0.159428299451592
0.345657234363748 0.133872957472937
0.348153669454283 0.109410704402582
0.350860122625084 0.0861400410998431
0.353765695944458 0.0641546703126039
0.35685868969415 0.0435431193694253
0.360126649479985 0.024388383711237
0.363556416381311 0.00676759269798993
0.36713417993733 -0.0092483009640541
0.370845533756952 -0.0235948069238276
0.374675533528245 -0.0362141568653407
0.378608757193904 -0.0470555371202997
0.382629367050434 -0.0560752932768158
0.386721173520993 -0.063237105960313
0.390867700345101 -0.0685121370788322
0.395052250922728 -0.0718791459438485
0.399257975545609 -0.0733245747990274
0.403467939245072 -0.072842603412524
0.40766518998318 -0.0704351725130031
0.411832826912603 -0.0661119759750093
0.415954068430369 -0.0598904217851557
0.420012319751453 -0.0517955619463062
0.423991239730117 -0.0418599916020015
0.427874806659945 -0.0301237177873203
0.431647382787589 -0.0166339983346699
0.435293777280491 -0.00144515158317349
0.438799307395 0.0153816623421094
0.442149857598601 0.033778687800421
0.445331936408178 0.0536718464633557
0.448332730715462 0.0749810356021446
0.451140157380887 0.0976204506330712
0.453742911888137 0.121498930622247
0.456130513863436 0.146520325358515
0.45829334927632 0.172583882516419
0.460222709151939 0.199584653350253
0.46191082463903 0.227413915285624
0.463350898292337 0.255959609706883
0.464537131443522 0.285106793177619
0.46546474755036 0.314738100277314
0.466130011430184 0.344734216190461
0.466530244300152 0.374974357145221
0.466663834563755 0.405336756767039
};
\addplot [semithick, sandybrown24416662, opacity=0.7]
table {%
0.91173274368321 0.0654475360020926
0.911603818112721 0.0991395183592101
0.911217560539023 0.132695834994751
0.910575526285038 0.165981366465116
0.909680300596104 0.198862083682996
0.908535488230107 0.231205587605198
0.90714569894239 0.262881642356865
0.905516528923898 0.29376269964538
0.903654538267289 0.323724412352317
0.90156722455176 0.352646135235406
0.899262992652938 0.380411410724339
0.896751120899416 0.406908437854302
0.894041723712196 0.432030522449004
0.891145710877486 0.455676506740465
0.888074743616842 0.477751176695646
0.884841187631543 0.498165645409755
0.881458063310279 0.51683771102244
0.877938993300639 0.533692187715667
0.874298147655518 0.548661208460476
0.870550186775314 0.56168449829355
0.866710202375675 0.572709617023219
0.862793656718477 0.581692170387608
0.858816320350762 0.588595988814652
0.854794208602307 0.593393273064207
0.850743517097543 0.596064706165771
0.846680556541495 0.59659953120111
0.842621687042324 0.594995594618569
0.83858325223495 0.591259354904667
0.834581513471004 0.58540585657805
0.830632584340099 0.577458669610523
0.826752365786094 0.567449794519093
0.822956482079608 0.555419533511185
0.819260217904593 0.541416328201871
0.815678456812309 0.525496564556587
0.812225621290514 0.507724345844753
0.808915614689197 0.488171234518527
0.805761765236696 0.46691596405607
0.802776772371625 0.44404412192962
0.799972655606716 0.419647804974945
0.797360706130487 0.393825248549895
0.794951441341606 0.366680430975282
0.792754562499044 0.338322654850885
0.790778915658525 0.308866106932458
0.789032456052583 0.278429398341961
0.78752221605765 0.247135086962443
0.786254276877155 0.215109183940696
0.785233744054667 0.182480646284847
0.784464726915671 0.149380857599996
0.783950322020767 0.115943099052813
0.78369260069691 0.08230201269532
0.78369260069691 0.0485930593088652
0.783950322020767 0.0149519729513717
0.784464726915671 -0.0184857855958112
0.785233744054667 -0.0515855742806618
0.786254276877155 -0.0842141119365111
0.78752221605765 -0.116240014958257
0.789032456052583 -0.147534326337776
0.790778915658525 -0.177971034928272
0.792754562499044 -0.2074275828467
0.794951441341606 -0.235785358971097
0.797360706130487 -0.26293017654571
0.799972655606716 -0.28875273297076
0.802776772371625 -0.313149049925434
0.805761765236696 -0.336020892051885
0.808915614689197 -0.357276162514341
0.812225621290514 -0.376829273840568
0.815678456812309 -0.394601492552402
0.819260217904593 -0.410521256197685
0.822956482079608 -0.424524461507
0.826752365786094 -0.436554722514908
0.830632584340099 -0.446563597606337
0.834581513471004 -0.454510784573865
0.83858325223495 -0.460364282900482
0.842621687042324 -0.464100522614384
0.846680556541495 -0.465704459196925
0.850743517097543 -0.465169634161586
0.854794208602307 -0.462498201060022
0.858816320350762 -0.457700916810467
0.862793656718477 -0.450797098383422
0.866710202375675 -0.441814545019034
0.870550186775314 -0.430789426289364
0.874298147655518 -0.417766136456291
0.877938993300639 -0.402797115711482
0.881458063310279 -0.385942639018254
0.884841187631543 -0.36727057340557
0.888074743616842 -0.346856104691461
0.891145710877486 -0.32478143473628
0.894041723712196 -0.301135450444819
0.896751120899416 -0.276013365850117
0.899262992652938 -0.249516338720153
0.90156722455176 -0.221751063231221
0.903654538267289 -0.192829340348132
0.905516528923898 -0.162867627641195
0.90714569894239 -0.13198657035268
0.908535488230107 -0.100310515601012
0.909680300596104 -0.0679670116788104
0.910575526285038 -0.0350862944609304
0.911217560539023 -0.00180076299056529
0.911603818112721 0.0317555536449752
0.91173274368321 0.0654475360020925
};
\end{axis}

\end{tikzpicture}

%% file: icra2024/tex/appendix/varying-obstacle-spaces/gamma-0.0-1.tex
\begin{tikzpicture}

\definecolor{sandybrown24416662}{RGB}{244,166,62}
\definecolor{seagreen4916384}{RGB}{49,163,84}

\begin{axis}[
width=0.6\linewidth, 
height=0.6\linewidth,  
tick pos=left,
xmin=0, xmax=1,
ymin=0, ymax=1,
xtick=\empty,
ytick=\empty,
xticklabels={},
yticklabels={},
]
\path [draw=sandybrown24416662, fill=sandybrown24416662, opacity=0.7]
(axis cs:0.223014052111481,0)
--(axis cs:0.222918748763103,0.0161046718362932)
--(axis cs:0.222633222470915,0.0321444958431451)
--(axis cs:0.222158622948522,0.0480548853104363)
--(axis cs:0.221496861240694,0.0637717747152546)
--(axis cs:0.220650602028262,0.0792318776911452)
--(axis cs:0.219623252898395,0.094372941859971)
--(axis cs:0.21841895062344,0.109133999500268)
--(axis cs:0.217042544503609,0.123455613042743)
--(axis cs:0.215499576840551,0.137280114404385)
--(axis cs:0.213796260620466,0.15055183719748)
--(axis cs:0.211939454496597,0.163217340878515)
--(axis cs:0.209936635171856,0.175225625934374)
--(axis cs:0.207795867292782,0.186528339239389)
--(axis cs:0.205525770976053,0.197079968756302)
--(axis cs:0.20313548709832,0.206838026797181)
--(axis cs:0.200634640489122,0.215763221106339)
--(axis cs:0.198033301175098,0.223819613076387)
--(axis cs:0.195341943831537,0.230974762460315)
--(axis cs:0.192571405604565,0.237199857996917)
--(axis cs:0.189732842473777,0.242469833423578)
--(axis cs:0.186837684331049,0.246763468409259)
--(axis cs:0.183897588956392,0.250063474001284)
--(axis cs:0.180924395076192,0.252356562241851)
--(axis cs:0.177930074692835,0.253633499673947)
--(axis cs:0.174926684877677,0.253889144521218)
--(axis cs:0.171926319221477,0.253122467392095)
--(axis cs:0.168941059137771,0.25133655542478)
--(axis cs:0.165982925215288,0.248538599856442)
--(axis cs:0.163063828815279,0.244739867066631)
--(axis cs:0.160195524108669,0.239955653211555)
--(axis cs:0.157389560746162,0.234205222631848)
--(axis cs:0.154657237351871,0.227511730281876)
--(axis cs:0.152009556027749,0.219902128492907)
--(axis cs:0.149457178052005,0.211407058445583)
--(axis cs:0.147010380949901,0.202060726788702)
--(axis cs:0.144679017109787,0.191900767901118)
--(axis cs:0.142472474111006,0.180968092351377)
--(axis cs:0.140399636923428,0.169306722165297)
--(axis cs:0.138468852130813,0.156963613564809)
--(axis cs:0.136687894322062,0.143988467891826)
--(axis cs:0.135063934785693,0.130433531478475)
--(axis cs:0.133603512633592,0.116353385269562)
--(axis cs:0.13231250847032,0.101804725044371)
--(axis cs:0.131196120713989,0.0868461331227697)
--(axis cs:0.130258844664073,0.0715378424748844)
--(axis cs:0.129504454400419,0.0559414941841885)
--(axis cs:0.128935987586361,0.0401198892406147)
--(axis cs:0.128555733237122,0.0241367356631337)
--(axis cs:0.128365222502759,0.00805639197004795)
--(axis cs:0.128365222502759,-0.008056391970048)
--(axis cs:0.128555733237122,-0.0241367356631338)
--(axis cs:0.128935987586361,-0.0401198892406147)
--(axis cs:0.129504454400419,-0.0559414941841885)
--(axis cs:0.130258844664073,-0.0715378424748844)
--(axis cs:0.131196120713989,-0.0868461331227696)
--(axis cs:0.13231250847032,-0.101804725044371)
--(axis cs:0.133603512633592,-0.116353385269562)
--(axis cs:0.135063934785693,-0.130433531478475)
--(axis cs:0.136687894322062,-0.143988467891826)
--(axis cs:0.138468852130813,-0.156963613564809)
--(axis cs:0.140399636923428,-0.169306722165297)
--(axis cs:0.142472474111006,-0.180968092351377)
--(axis cs:0.144679017109787,-0.191900767901118)
--(axis cs:0.147010380949901,-0.202060726788703)
--(axis cs:0.149457178052005,-0.211407058445583)
--(axis cs:0.152009556027749,-0.219902128492907)
--(axis cs:0.154657237351871,-0.227511730281876)
--(axis cs:0.157389560746162,-0.234205222631848)
--(axis cs:0.160195524108669,-0.239955653211555)
--(axis cs:0.163063828815279,-0.244739867066631)
--(axis cs:0.165982925215288,-0.248538599856442)
--(axis cs:0.168941059137771,-0.25133655542478)
--(axis cs:0.171926319221477,-0.253122467392095)
--(axis cs:0.174926684877677,-0.253889144521218)
--(axis cs:0.177930074692835,-0.253633499673947)
--(axis cs:0.180924395076192,-0.252356562241851)
--(axis cs:0.183897588956392,-0.250063474001284)
--(axis cs:0.186837684331049,-0.246763468409259)
--(axis cs:0.189732842473777,-0.242469833423578)
--(axis cs:0.192571405604565,-0.237199857996917)
--(axis cs:0.195341943831537,-0.230974762460315)
--(axis cs:0.198033301175098,-0.223819613076387)
--(axis cs:0.200634640489122,-0.215763221106339)
--(axis cs:0.20313548709832,-0.206838026797181)
--(axis cs:0.205525770976053,-0.197079968756302)
--(axis cs:0.207795867292782,-0.186528339239389)
--(axis cs:0.209936635171856,-0.175225625934374)
--(axis cs:0.211939454496597,-0.163217340878515)
--(axis cs:0.213796260620466,-0.15055183719748)
--(axis cs:0.215499576840551,-0.137280114404385)
--(axis cs:0.217042544503609,-0.123455613042743)
--(axis cs:0.21841895062344,-0.109133999500269)
--(axis cs:0.219623252898395,-0.0943729418599709)
--(axis cs:0.220650602028262,-0.0792318776911451)
--(axis cs:0.221496861240694,-0.0637717747152547)
--(axis cs:0.222158622948522,-0.0480548853104362)
--(axis cs:0.222633222470915,-0.0321444958431451)
--(axis cs:0.222918748763103,-0.0161046718362932)
--(axis cs:0.223014052111481,-6.2192733979328e-17)
--cycle;
\path [draw=sandybrown24416662, fill=sandybrown24416662, opacity=0.7]
(axis cs:0.465876735370982,0.359203557852482)
--(axis cs:0.465759424774706,0.381382907086414)
--(axis cs:0.465407965354246,0.403472947909044)
--(axis cs:0.464823772312642,0.425384731522521)
--(axis cs:0.464009197989086,0.447030026907862)
--(axis cs:0.462967522386879,0.468321676100122)
--(axis cs:0.461702939966019,0.489173945142759)
--(axis cs:0.460220542753588,0.509502869308016)
--(axis cs:0.458526299839958,0.529226591193247)
--(axis cs:0.456627033343369,0.548265690331785)
--(axis cs:0.454530390939669,0.56654350299113)
--(axis cs:0.452244815067821,0.583986430870744)
--(axis cs:0.449779508935181,0.600524237456427)
--(axis cs:0.44714439945943,0.616090330837976)
--(axis cs:0.444350097296383,0.630622031851297)
--(axis cs:0.441407854114617,0.644060826465285)
--(axis cs:0.43832951728898,0.656352601397167)
--(axis cs:0.435127482195389,0.667447862007598)
--(axis cs:0.431814642299027,0.677301931598102)
--(axis cs:0.42840433723691,0.685875131308372)
--(axis cs:0.424910299103879,0.693132939889031)
--(axis cs:0.421346597158295,0.699046132706524)
--(axis cs:0.417727581170107,0.703590899420398)
--(axis cs:0.41406782363939,0.706748939859145)
--(axis cs:0.410382061118033,0.708507537708528)
--(axis cs:0.406685134870841,0.708859611715694)
--(axis cs:0.402991931115004,0.707803744202878)
--(axis cs:0.399317321078556,0.705344186775892)
--(axis cs:0.39567610111918,0.70149084320441)
--(axis cs:0.392082933144509,0.696259229542981)
--(axis cs:0.38855228557379,0.689670411653353)
--(axis cs:0.385098375078669,0.681750920379685)
--(axis cs:0.381735109337675,0.672532644718202)
--(axis cs:0.378476031034903,0.66205270341146)
--(axis cs:0.375334263328407,0.65035329548427)
--(axis cs:0.372322457007873,0.637481530323123)
--(axis cs:0.369452739554356,0.623489237983314)
--(axis cs:0.366736666307193,0.608432760487601)
--(axis cs:0.364185173934728,0.592372724956771)
--(axis cs:0.361808536396213,0.575373799485616)
--(axis cs:0.359616323572195,0.557504432747344)
--(axis cs:0.357617362729989,0.538836578374929)
--(axis cs:0.355819702979386,0.519445405229226)
--(axis cs:0.354230582861728,0.499408994720497)
--(axis cs:0.35285640120286,0.478808026402137)
--(axis cs:0.351702691347312,0.457725453102579)
--(axis cs:0.350774098877479,0.436246166903538)
--(axis cs:0.350074362907495,0.414456657309557)
--(axis cs:0.349606301027144,0.392444662985299)
--(axis cs:0.349371797956419,0.370298818462913)
--(axis cs:0.349371797956419,0.348108297242051)
--(axis cs:0.349606301027144,0.325962452719665)
--(axis cs:0.350074362907495,0.303950458395408)
--(axis cs:0.350774098877479,0.282160948801427)
--(axis cs:0.351702691347312,0.260681662602386)
--(axis cs:0.35285640120286,0.239599089302828)
--(axis cs:0.354230582861728,0.218998120984467)
--(axis cs:0.355819702979386,0.198961710475739)
--(axis cs:0.357617362729989,0.179570537330035)
--(axis cs:0.359616323572195,0.16090268295762)
--(axis cs:0.361808536396213,0.143033316219349)
--(axis cs:0.364185173934728,0.126034390748194)
--(axis cs:0.366736666307193,0.109974355217364)
--(axis cs:0.369452739554356,0.0949178777216509)
--(axis cs:0.372322457007873,0.080925585381841)
--(axis cs:0.375334263328407,0.068053820220694)
--(axis cs:0.378476031034903,0.0563544122935044)
--(axis cs:0.381735109337675,0.0458744709867622)
--(axis cs:0.385098375078669,0.0366561953252794)
--(axis cs:0.38855228557379,0.0287367040516119)
--(axis cs:0.392082933144509,0.0221478861619837)
--(axis cs:0.39567610111918,0.016916272500554)
--(axis cs:0.399317321078556,0.0130629289290721)
--(axis cs:0.402991931115004,0.0106033715020866)
--(axis cs:0.406685134870841,0.00954750398927062)
--(axis cs:0.410382061118033,0.00989957799643637)
--(axis cs:0.41406782363939,0.0116581758458192)
--(axis cs:0.417727581170107,0.014816216284566)
--(axis cs:0.421346597158295,0.0193609829984406)
--(axis cs:0.424910299103879,0.0252741758159332)
--(axis cs:0.42840433723691,0.0325319843965924)
--(axis cs:0.431814642299027,0.0411051841068622)
--(axis cs:0.435127482195389,0.0509592536973667)
--(axis cs:0.43832951728898,0.0620545143077974)
--(axis cs:0.441407854114617,0.0743462892396792)
--(axis cs:0.444350097296383,0.087785083853667)
--(axis cs:0.44714439945943,0.102316784866989)
--(axis cs:0.449779508935181,0.117882878248537)
--(axis cs:0.452244815067821,0.134420684834221)
--(axis cs:0.454530390939669,0.151863612713834)
--(axis cs:0.456627033343369,0.17014142537318)
--(axis cs:0.458526299839958,0.189180524511718)
--(axis cs:0.460220542753588,0.208904246396949)
--(axis cs:0.461702939966019,0.229233170562206)
--(axis cs:0.462967522386879,0.250085439604842)
--(axis cs:0.464009197989086,0.271377088797102)
--(axis cs:0.464823772312642,0.293022384182443)
--(axis cs:0.465407965354246,0.314934167795921)
--(axis cs:0.465759424774706,0.33702420861855)
--(axis cs:0.465876735370982,0.359203557852482)
--cycle;
\path [draw=sandybrown24416662, fill=sandybrown24416662, opacity=0.7]
(axis cs:0.875159308676694,0.0850077919941101)
--(axis cs:0.875063064046179,0.141489329132554)
--(axis cs:0.874774717697794,0.197743435055748)
--(axis cs:0.874295430700512,0.253543594333676)
--(axis cs:0.873627132973911,0.308665119419243)
--(axis cs:0.872772515517063,0.362886055385729)
--(axis cs:0.871735019572837,0.415988073660939)
--(axis cs:0.870518822771221,0.467757351159305)
--(axis cs:0.869128822307487,0.517985431271984)
--(axis cs:0.86757061522291,0.566470063248051)
--(axis cs:0.865850475867459,0.613016016586887)
--(axis cs:0.863975330635208,0.657435867162501)
--(axis cs:0.861952730074186,0.699550751914334)
--(axis cs:0.859790818482987,0.739191089065676)
--(axis cs:0.857498301116555,0.776197260969629)
--(axis cs:0.855084409133197,0.810420256833035)
--(axis cs:0.852558862423961,0.841722272730343)
--(axis cs:0.849931830474068,0.869977266491387)
--(axis cs:0.847213891413987,0.895071465228715)
--(axis cs:0.844415989425033,0.916903823460853)
--(axis cs:0.841549390671013,0.935386429986795)
--(axis cs:0.838625637933362,0.950444861873383)
--(axis cs:0.835656504132433,0.962018484130187)
--(axis cs:0.832653944922106,0.970060693865218)
--(axis cs:0.829630050548587,0.974539107938321)
--(axis cs:0.826596997167252,0.975435693356652)
--(axis cs:0.823566997813571,0.972746839887176)
--(axis cs:0.820552253225522,0.966483374593799)
--(axis cs:0.817564902715526,0.956670518240607)
--(axis cs:0.81461697528972,0.943347783736749)
--(axis cs:0.811720341211394,0.926568817031894)
--(axis cs:0.808886664203634,0.906401181102932)
--(axis cs:0.806127354483621,0.882926083901708)
--(axis cs:0.803453522817716,0.85623805135926)
--(axis cs:0.800875935782325,0.826444546763264)
--(axis cs:0.798404972410698,0.793665538041294)
--(axis cs:0.796050582400219,0.758033014692329)
--(axis cs:0.793822246048495,0.719690456311616)
--(axis cs:0.791728936079536,0.678792254848988)
--(axis cs:0.789779081513766,0.635503092926969)
--(axis cs:0.787980533727333,0.589997280721981)
--(axis cs:0.786340534837387,0.542458054078778)
--(axis cs:0.784865688540629,0.49307683668438)
--(axis cs:0.78356193352256,0.442052469272441)
--(axis cs:0.782434519544484,0.389590408961796)
--(axis cs:0.781487986304578,0.335901901953159)
--(axis cs:0.780726145158126,0.28120313291523)
--(axis cs:0.780152063770537,0.225714354485334)
--(axis cs:0.779768053764941,0.16965900038979)
--(axis cs:0.779575661414097,0.113262785755154)
--(axis cs:0.779575661414097,0.0567527982330659)
--(axis cs:0.779768053764941,0.000356583598429799)
--(axis cs:0.780152063770537,-0.0556987704971138)
--(axis cs:0.780726145158126,-0.111187548927009)
--(axis cs:0.781487986304578,-0.165886317964939)
--(axis cs:0.782434519544484,-0.219574824973576)
--(axis cs:0.78356193352256,-0.272036885284221)
--(axis cs:0.784865688540629,-0.32306125269616)
--(axis cs:0.786340534837387,-0.372442470090558)
--(axis cs:0.787980533727333,-0.41998169673376)
--(axis cs:0.789779081513766,-0.465487508938749)
--(axis cs:0.791728936079536,-0.508776670860767)
--(axis cs:0.793822246048495,-0.549674872323396)
--(axis cs:0.796050582400219,-0.588017430704108)
--(axis cs:0.798404972410698,-0.623649954053074)
--(axis cs:0.800875935782325,-0.656428962775043)
--(axis cs:0.803453522817716,-0.68622246737104)
--(axis cs:0.806127354483621,-0.712910499913488)
--(axis cs:0.808886664203634,-0.736385597114712)
--(axis cs:0.811720341211394,-0.756553233043674)
--(axis cs:0.81461697528972,-0.773332199748529)
--(axis cs:0.817564902715526,-0.786654934252387)
--(axis cs:0.820552253225522,-0.796467790605578)
--(axis cs:0.823566997813571,-0.802731255898955)
--(axis cs:0.826596997167252,-0.805420109368432)
--(axis cs:0.829630050548587,-0.804523523950101)
--(axis cs:0.832653944922106,-0.800045109876998)
--(axis cs:0.835656504132433,-0.792002900141967)
--(axis cs:0.838625637933362,-0.780429277885162)
--(axis cs:0.841549390671013,-0.765370845998575)
--(axis cs:0.844415989425033,-0.746888239472633)
--(axis cs:0.847213891413987,-0.725055881240495)
--(axis cs:0.849931830474068,-0.699961682503166)
--(axis cs:0.852558862423961,-0.671706688742122)
--(axis cs:0.855084409133197,-0.640404672844815)
--(axis cs:0.857498301116555,-0.606181676981409)
--(axis cs:0.859790818482987,-0.569175505077456)
--(axis cs:0.861952730074186,-0.529535167926113)
--(axis cs:0.863975330635208,-0.487420283174281)
--(axis cs:0.865850475867459,-0.443000432598667)
--(axis cs:0.86757061522291,-0.396454479259831)
--(axis cs:0.869128822307487,-0.347969847283764)
--(axis cs:0.870518822771221,-0.297741767171085)
--(axis cs:0.871735019572837,-0.245972489672719)
--(axis cs:0.872772515517063,-0.192870471397509)
--(axis cs:0.873627132973911,-0.138649535431023)
--(axis cs:0.874295430700512,-0.083528010345455)
--(axis cs:0.874774717697794,-0.0277278510675276)
--(axis cs:0.875063064046179,0.0285262548556662)
--(axis cs:0.875159308676694,0.0850077919941099)
--cycle;
\addplot [semithick, seagreen4916384]
table {%
0 0
0.037496293091515 0.000634539039349185
0.0533252309914267 0.00171354380317028
0.0733929133328653 0.000267692256786874
0.109189336049712 0.00275059648434714
0.122065396598222 0.0419689968010497
0.11290969774906 0.0963720995911638
0.116675446477857 0.149319861675762
0.129477382795203 0.188973561351336
0.152358956981549 0.227947592351947
0.167209567588023 0.257207471274667
0.210378159397903 0.247858943116905
0.239280674258946 0.222330153677558
0.258693709774853 0.185087874568434
0.261876700826594 0.152160030819255
0.234503176922471 0.122991663954148
0.231010491880464 0.0794927287202139
0.243570723453852 0.0371958995978765
0.238320865134699 0.00134572423601285
0.255929663200503 0.00153068585457643
0.281498947213094 0.00328257792066019
0.303107151464159 0.000190809475542356
0.33138820381095 0.0059146494323334
0.355502287050124 0.00244823073674717
0.383199135263016 0.00680406339652877
0.400324674212302 0.000458293151555957
0.42817989308727 0.00235763070294049
0.455127896214152 0.0028704439561888
0.484821640002806 0.000184884024987951
0.52733735132434 0.00184911481974859
0.548832562273934 0.00582466155324396
0.574595974094954 0.00364684127300616
0.591696074999786 0.00906458892959148
0.625149366788299 0.00694884914895657
0.680294614890607 0.0043631204496986
0.719781103473537 0.00188753497616572
0.778960774622299 0.0200950821923051
0.766336526361109 0.09371040156491
0.773810445255726 0.125221282472394
0.765461500749083 0.179802502430792
0.776718794154425 0.229861440592123
0.769561222948628 0.288670218764933
0.777412758512293 0.309006445456842
0.773228795876464 0.37052376557435
0.776094700353515 0.401199415962217
0.758633697806266 0.456821888351026
0.771827594065182 0.505600876210136
0.776913255523105 0.564082717787861
0.750737130846097 0.615335331951552
0.763682936949989 0.647423552002511
0.790740690022597 0.687770911377167
0.772747543442002 0.750257199204754
0.771999385395291 0.808655822598221
0.76438262829981 0.848112870460408
0.753581022057803 0.884540623808424
0.753569231023508 0.921247897011896
0.781468083846895 0.958750011553269
0.829092302071497 0.99294152299124
0.85258609876124 0.922030385717513
0.868408312427707 0.868076276795138
0.856467575994592 0.830425902069228
0.871799768266396 0.790927209556489
0.890983057736126 0.761962482560448
0.878191749678281 0.717635234228825
0.877453174619977 0.679919203834358
0.87382400355738 0.615014197893005
0.87760872445262 0.536292489331166
0.876942231077571 0.480922252737948
0.88576165598272 0.435916117000039
0.879601207650251 0.38683559159828
0.915196086991316 0.340584743696062
0.907764211817186 0.288996203788662
0.892604654225725 0.256617536217284
0.895155999938938 0.210913407918952
0.901077015809694 0.178664094077454
0.884414265593791 0.137017862045939
0.895256659876369 0.071199860454335
0.935597422641381 0.0233219415553266
0.954501943381832 0.00180476053378228
1 0
};
\addplot [semithick, sandybrown24416662, opacity=0.7]
table {%
0.223014052111481 0
0.222918748763103 0.0161046718362932
0.222633222470915 0.0321444958431451
0.222158622948522 0.0480548853104363
0.221496861240694 0.0637717747152546
0.220650602028262 0.0792318776911452
0.219623252898395 0.094372941859971
0.21841895062344 0.109133999500268
0.217042544503609 0.123455613042743
0.215499576840551 0.137280114404385
0.213796260620466 0.15055183719748
0.211939454496597 0.163217340878515
0.209936635171856 0.175225625934374
0.207795867292782 0.186528339239389
0.205525770976053 0.197079968756302
0.20313548709832 0.206838026797181
0.200634640489122 0.215763221106339
0.198033301175098 0.223819613076387
0.195341943831537 0.230974762460315
0.192571405604565 0.237199857996917
0.189732842473777 0.242469833423578
0.186837684331049 0.246763468409259
0.183897588956392 0.250063474001284
0.180924395076192 0.252356562241851
0.177930074692835 0.253633499673947
0.174926684877677 0.253889144521218
0.171926319221477 0.253122467392095
0.168941059137771 0.25133655542478
0.165982925215288 0.248538599856442
0.163063828815279 0.244739867066631
0.160195524108669 0.239955653211555
0.157389560746162 0.234205222631848
0.154657237351871 0.227511730281876
0.152009556027749 0.219902128492907
0.149457178052005 0.211407058445583
0.147010380949901 0.202060726788702
0.144679017109787 0.191900767901118
0.142472474111006 0.180968092351377
0.140399636923428 0.169306722165297
0.138468852130813 0.156963613564809
0.136687894322062 0.143988467891826
0.135063934785693 0.130433531478475
0.133603512633592 0.116353385269562
0.13231250847032 0.101804725044371
0.131196120713989 0.0868461331227697
0.130258844664073 0.0715378424748844
0.129504454400419 0.0559414941841885
0.128935987586361 0.0401198892406147
0.128555733237122 0.0241367356631337
0.128365222502759 0.00805639197004795
0.128365222502759 -0.008056391970048
0.128555733237122 -0.0241367356631338
0.128935987586361 -0.0401198892406147
0.129504454400419 -0.0559414941841885
0.130258844664073 -0.0715378424748844
0.131196120713989 -0.0868461331227696
0.13231250847032 -0.101804725044371
0.133603512633592 -0.116353385269562
0.135063934785693 -0.130433531478475
0.136687894322062 -0.143988467891826
0.138468852130813 -0.156963613564809
0.140399636923428 -0.169306722165297
0.142472474111006 -0.180968092351377
0.144679017109787 -0.191900767901118
0.147010380949901 -0.202060726788703
0.149457178052005 -0.211407058445583
0.152009556027749 -0.219902128492907
0.154657237351871 -0.227511730281876
0.157389560746162 -0.234205222631848
0.160195524108669 -0.239955653211555
0.163063828815279 -0.244739867066631
0.165982925215288 -0.248538599856442
0.168941059137771 -0.25133655542478
0.171926319221477 -0.253122467392095
0.174926684877677 -0.253889144521218
0.177930074692835 -0.253633499673947
0.180924395076192 -0.252356562241851
0.183897588956392 -0.250063474001284
0.186837684331049 -0.246763468409259
0.189732842473777 -0.242469833423578
0.192571405604565 -0.237199857996917
0.195341943831537 -0.230974762460315
0.198033301175098 -0.223819613076387
0.200634640489122 -0.215763221106339
0.20313548709832 -0.206838026797181
0.205525770976053 -0.197079968756302
0.207795867292782 -0.186528339239389
0.209936635171856 -0.175225625934374
0.211939454496597 -0.163217340878515
0.213796260620466 -0.15055183719748
0.215499576840551 -0.137280114404385
0.217042544503609 -0.123455613042743
0.21841895062344 -0.109133999500269
0.219623252898395 -0.0943729418599709
0.220650602028262 -0.0792318776911451
0.221496861240694 -0.0637717747152547
0.222158622948522 -0.0480548853104362
0.222633222470915 -0.0321444958431451
0.222918748763103 -0.0161046718362932
0.223014052111481 -6.2192733979328e-17
};
\addplot [semithick, sandybrown24416662, opacity=0.7]
table {%
0.465876735370982 0.359203557852482
0.465759424774706 0.381382907086414
0.465407965354246 0.403472947909044
0.464823772312642 0.425384731522521
0.464009197989086 0.447030026907862
0.462967522386879 0.468321676100122
0.461702939966019 0.489173945142759
0.460220542753588 0.509502869308016
0.458526299839958 0.529226591193247
0.456627033343369 0.548265690331785
0.454530390939669 0.56654350299113
0.452244815067821 0.583986430870744
0.449779508935181 0.600524237456427
0.44714439945943 0.616090330837976
0.444350097296383 0.630622031851297
0.441407854114617 0.644060826465285
0.43832951728898 0.656352601397167
0.435127482195389 0.667447862007598
0.431814642299027 0.677301931598102
0.42840433723691 0.685875131308372
0.424910299103879 0.693132939889031
0.421346597158295 0.699046132706524
0.417727581170107 0.703590899420398
0.41406782363939 0.706748939859145
0.410382061118033 0.708507537708528
0.406685134870841 0.708859611715694
0.402991931115004 0.707803744202878
0.399317321078556 0.705344186775892
0.39567610111918 0.70149084320441
0.392082933144509 0.696259229542981
0.38855228557379 0.689670411653353
0.385098375078669 0.681750920379685
0.381735109337675 0.672532644718202
0.378476031034903 0.66205270341146
0.375334263328407 0.65035329548427
0.372322457007873 0.637481530323123
0.369452739554356 0.623489237983314
0.366736666307193 0.608432760487601
0.364185173934728 0.592372724956771
0.361808536396213 0.575373799485616
0.359616323572195 0.557504432747344
0.357617362729989 0.538836578374929
0.355819702979386 0.519445405229226
0.354230582861728 0.499408994720497
0.35285640120286 0.478808026402137
0.351702691347312 0.457725453102579
0.350774098877479 0.436246166903538
0.350074362907495 0.414456657309557
0.349606301027144 0.392444662985299
0.349371797956419 0.370298818462913
0.349371797956419 0.348108297242051
0.349606301027144 0.325962452719665
0.350074362907495 0.303950458395408
0.350774098877479 0.282160948801427
0.351702691347312 0.260681662602386
0.35285640120286 0.239599089302828
0.354230582861728 0.218998120984467
0.355819702979386 0.198961710475739
0.357617362729989 0.179570537330035
0.359616323572195 0.16090268295762
0.361808536396213 0.143033316219349
0.364185173934728 0.126034390748194
0.366736666307193 0.109974355217364
0.369452739554356 0.0949178777216509
0.372322457007873 0.080925585381841
0.375334263328407 0.068053820220694
0.378476031034903 0.0563544122935044
0.381735109337675 0.0458744709867622
0.385098375078669 0.0366561953252794
0.38855228557379 0.0287367040516119
0.392082933144509 0.0221478861619837
0.39567610111918 0.016916272500554
0.399317321078556 0.0130629289290721
0.402991931115004 0.0106033715020866
0.406685134870841 0.00954750398927062
0.410382061118033 0.00989957799643637
0.41406782363939 0.0116581758458192
0.417727581170107 0.014816216284566
0.421346597158295 0.0193609829984406
0.424910299103879 0.0252741758159332
0.42840433723691 0.0325319843965924
0.431814642299027 0.0411051841068622
0.435127482195389 0.0509592536973667
0.43832951728898 0.0620545143077974
0.441407854114617 0.0743462892396792
0.444350097296383 0.087785083853667
0.44714439945943 0.102316784866989
0.449779508935181 0.117882878248537
0.452244815067821 0.134420684834221
0.454530390939669 0.151863612713834
0.456627033343369 0.17014142537318
0.458526299839958 0.189180524511718
0.460220542753588 0.208904246396949
0.461702939966019 0.229233170562206
0.462967522386879 0.250085439604842
0.464009197989086 0.271377088797102
0.464823772312642 0.293022384182443
0.465407965354246 0.314934167795921
0.465759424774706 0.33702420861855
0.465876735370982 0.359203557852482
};
\addplot [semithick, sandybrown24416662, opacity=0.7]
table {%
0.875159308676694 0.0850077919941101
0.875063064046179 0.141489329132554
0.874774717697794 0.197743435055748
0.874295430700512 0.253543594333676
0.873627132973911 0.308665119419243
0.872772515517063 0.362886055385729
0.871735019572837 0.415988073660939
0.870518822771221 0.467757351159305
0.869128822307487 0.517985431271984
0.86757061522291 0.566470063248051
0.865850475867459 0.613016016586887
0.863975330635208 0.657435867162501
0.861952730074186 0.699550751914334
0.859790818482987 0.739191089065676
0.857498301116555 0.776197260969629
0.855084409133197 0.810420256833035
0.852558862423961 0.841722272730343
0.849931830474068 0.869977266491387
0.847213891413987 0.895071465228715
0.844415989425033 0.916903823460853
0.841549390671013 0.935386429986795
0.838625637933362 0.950444861873383
0.835656504132433 0.962018484130187
0.832653944922106 0.970060693865218
0.829630050548587 0.974539107938321
0.826596997167252 0.975435693356652
0.823566997813571 0.972746839887176
0.820552253225522 0.966483374593799
0.817564902715526 0.956670518240607
0.81461697528972 0.943347783736749
0.811720341211394 0.926568817031894
0.808886664203634 0.906401181102932
0.806127354483621 0.882926083901708
0.803453522817716 0.85623805135926
0.800875935782325 0.826444546763264
0.798404972410698 0.793665538041294
0.796050582400219 0.758033014692329
0.793822246048495 0.719690456311616
0.791728936079536 0.678792254848988
0.789779081513766 0.635503092926969
0.787980533727333 0.589997280721981
0.786340534837387 0.542458054078778
0.784865688540629 0.49307683668438
0.78356193352256 0.442052469272441
0.782434519544484 0.389590408961796
0.781487986304578 0.335901901953159
0.780726145158126 0.28120313291523
0.780152063770537 0.225714354485334
0.779768053764941 0.16965900038979
0.779575661414097 0.113262785755154
0.779575661414097 0.0567527982330659
0.779768053764941 0.000356583598429799
0.780152063770537 -0.0556987704971138
0.780726145158126 -0.111187548927009
0.781487986304578 -0.165886317964939
0.782434519544484 -0.219574824973576
0.78356193352256 -0.272036885284221
0.784865688540629 -0.32306125269616
0.786340534837387 -0.372442470090558
0.787980533727333 -0.41998169673376
0.789779081513766 -0.465487508938749
0.791728936079536 -0.508776670860767
0.793822246048495 -0.549674872323396
0.796050582400219 -0.588017430704108
0.798404972410698 -0.623649954053074
0.800875935782325 -0.656428962775043
0.803453522817716 -0.68622246737104
0.806127354483621 -0.712910499913488
0.808886664203634 -0.736385597114712
0.811720341211394 -0.756553233043674
0.81461697528972 -0.773332199748529
0.817564902715526 -0.786654934252387
0.820552253225522 -0.796467790605578
0.823566997813571 -0.802731255898955
0.826596997167252 -0.805420109368432
0.829630050548587 -0.804523523950101
0.832653944922106 -0.800045109876998
0.835656504132433 -0.792002900141967
0.838625637933362 -0.780429277885162
0.841549390671013 -0.765370845998575
0.844415989425033 -0.746888239472633
0.847213891413987 -0.725055881240495
0.849931830474068 -0.699961682503166
0.852558862423961 -0.671706688742122
0.855084409133197 -0.640404672844815
0.857498301116555 -0.606181676981409
0.859790818482987 -0.569175505077456
0.861952730074186 -0.529535167926113
0.863975330635208 -0.487420283174281
0.865850475867459 -0.443000432598667
0.86757061522291 -0.396454479259831
0.869128822307487 -0.347969847283764
0.870518822771221 -0.297741767171085
0.871735019572837 -0.245972489672719
0.872772515517063 -0.192870471397509
0.873627132973911 -0.138649535431023
0.874295430700512 -0.083528010345455
0.874774717697794 -0.0277278510675276
0.875063064046179 0.0285262548556662
0.875159308676694 0.0850077919941099
};
\end{axis}

\end{tikzpicture}

%% file: icra2024/tex/appendix/varying-obstacle-spaces/gamma-0.0-2.tex
\begin{tikzpicture}

\definecolor{sandybrown24416662}{RGB}{244,166,62}
\definecolor{seagreen4916384}{RGB}{49,163,84}

\begin{axis}[
width=0.6\linewidth, 
height=0.6\linewidth,  
tick pos=left,
xmin=0, xmax=1,
ymin=0, ymax=1,
xtick=\empty,
ytick=\empty,
xticklabels={},
yticklabels={},
]
\path [draw=sandybrown24416662, fill=sandybrown24416662, opacity=0.7]
(axis cs:0.257650231540891,0)
--(axis cs:0.257537694125508,0.0566227800020072)
--(axis cs:0.25720053502781,0.113017560053604)
--(axis cs:0.256640111868485,0.168957258279708)
--(axis cs:0.255858681273799,0.224216625259131)
--(axis cs:0.254859389788954,0.278573151024497)
--(axis cs:0.253646261208068,0.331807961031345)
--(axis cs:0.252224180371768,0.383706697488671)
--(axis cs:0.250598873497669,0.434060382502099)
--(axis cs:0.248776885122909,0.482666259554098)
--(axis cs:0.246765551751616,0.529328609932922)
--(axis cs:0.244572972313387,0.57385954082278)
--(axis cs:0.242207975551763,0.616079741881894)
--(axis cs:0.239680084473992,0.655819207261955)
--(axis cs:0.236999478005241,0.692917920161676)
--(axis cs:0.234176950001656,0.727226497157967)
--(axis cs:0.231223865787314,0.758606789720264)
--(axis cs:0.228152116390074,0.786932440485897)
--(axis cs:0.224974070660603,0.812089392056594)
--(axis cs:0.221702525467383,0.83397634626737)
--(axis cs:0.218350654168234,0.852505172078482)
--(axis cs:0.214931953565858,0.867601260448023)
--(axis cs:0.21146018956097,0.879203824756198)
--(axis cs:0.20794934172188,0.887266145571591)
--(axis cs:0.204413546993705,0.891755758773825)
--(axis cs:0.200867042773879,0.892654586275115)
--(axis cs:0.197324109583178,0.889959008814349)
--(axis cs:0.193799013563104,0.883679880530574)
--(axis cs:0.190305949031171,0.87384248525721)
--(axis cs:0.186858981325392,0.860486434712975)
--(axis cs:0.183471990168133,0.843665508999476)
--(axis cs:0.180158613777368,0.823447440047721)
--(axis cs:0.176932193950385,0.799913638885536)
--(axis cs:0.173805722341084,0.773158867824094)
--(axis cs:0.170791788147167,0.743290858883529)
--(axis cs:0.167902527417882,0.710429879994117)
--(axis cs:0.165149574186441,0.674708250719768)
--(axis cs:0.162544013623869,0.636269809453843)
--(axis cs:0.160096337402935,0.595269334232713)
--(axis cs:0.157816401451888,0.551871919499229)
--(axis cs:0.155713386268114,0.506252311325677)
--(axis cs:0.153795759951512,0.458594203773015)
--(axis cs:0.152071244106445,0.409089499219727)
--(axis cs:0.150546782749564,0.357937535638669)
--(axis cs:0.149228514348703,0.305344283933418)
--(axis cs:0.148121747105439,0.251521518566129)
--(axis cs:0.147230937580834,0.196685964816527)
--(axis cs:0.146559672750436,0.141058426105695)
--(axis cs:0.146110655560801,0.0848628948986269)
--(axis cs:0.145885694045672,0.0283256507656328)
--(axis cs:0.145885694045672,-0.028325650765633)
--(axis cs:0.146110655560801,-0.0848628948986271)
--(axis cs:0.146559672750436,-0.141058426105695)
--(axis cs:0.147230937580834,-0.196685964816527)
--(axis cs:0.148121747105439,-0.251521518566129)
--(axis cs:0.149228514348703,-0.305344283933418)
--(axis cs:0.150546782749564,-0.357937535638669)
--(axis cs:0.152071244106445,-0.409089499219726)
--(axis cs:0.153795759951512,-0.458594203773015)
--(axis cs:0.155713386268114,-0.506252311325677)
--(axis cs:0.157816401451888,-0.551871919499229)
--(axis cs:0.160096337402935,-0.595269334232713)
--(axis cs:0.162544013623869,-0.636269809453844)
--(axis cs:0.165149574186441,-0.674708250719768)
--(axis cs:0.167902527417882,-0.710429879994117)
--(axis cs:0.170791788147167,-0.743290858883529)
--(axis cs:0.173805722341084,-0.773158867824094)
--(axis cs:0.176932193950385,-0.799913638885537)
--(axis cs:0.180158613777368,-0.823447440047721)
--(axis cs:0.183471990168133,-0.843665508999476)
--(axis cs:0.186858981325392,-0.860486434712975)
--(axis cs:0.190305949031171,-0.87384248525721)
--(axis cs:0.193799013563104,-0.883679880530574)
--(axis cs:0.197324109583178,-0.889959008814348)
--(axis cs:0.200867042773879,-0.892654586275115)
--(axis cs:0.204413546993705,-0.891755758773825)
--(axis cs:0.20794934172188,-0.887266145571591)
--(axis cs:0.21146018956097,-0.879203824756198)
--(axis cs:0.214931953565858,-0.867601260448023)
--(axis cs:0.218350654168234,-0.852505172078482)
--(axis cs:0.221702525467383,-0.83397634626737)
--(axis cs:0.224974070660603,-0.812089392056594)
--(axis cs:0.228152116390074,-0.786932440485897)
--(axis cs:0.231223865787314,-0.758606789720264)
--(axis cs:0.234176950001656,-0.727226497157967)
--(axis cs:0.236999478005241,-0.692917920161675)
--(axis cs:0.239680084473992,-0.655819207261955)
--(axis cs:0.242207975551763,-0.616079741881894)
--(axis cs:0.244572972313387,-0.57385954082278)
--(axis cs:0.246765551751616,-0.529328609932922)
--(axis cs:0.248776885122909,-0.482666259554098)
--(axis cs:0.250598873497669,-0.434060382502099)
--(axis cs:0.252224180371768,-0.383706697488672)
--(axis cs:0.253646261208068,-0.331807961031344)
--(axis cs:0.254859389788954,-0.278573151024497)
--(axis cs:0.255858681273799,-0.224216625259131)
--(axis cs:0.256640111868485,-0.168957258279708)
--(axis cs:0.25720053502781,-0.113017560053604)
--(axis cs:0.257537694125508,-0.0566227800020072)
--(axis cs:0.257650231540891,-2.18664840217285e-16)
--cycle;
\path [draw=sandybrown24416662, fill=sandybrown24416662, opacity=0.7]
(axis cs:0.568369647558041,0.308099524275207)
--(axis cs:0.568197886553612,0.327458120985732)
--(axis cs:0.567683295161268,0.346738767459801)
--(axis cs:0.566827945458924,0.365863827339045)
--(axis cs:0.565635281637941,0.384756290757402)
--(axis cs:0.564110106134582,0.403340084432663)
--(axis cs:0.562258560292299,0.421540377986732)
--(axis cs:0.560088099632707,0.439283885261147)
--(axis cs:0.557607463834831,0.456499159414575)
--(axis cs:0.554826641543505,0.473116880614023)
--(axis cs:0.551756830148622,0.489070135161342)
--(axis cs:0.548410390697198,0.504294684931072)
--(axis cs:0.5448007981198,0.518729226034701)
--(axis cs:0.540942586971752,0.532315635669785)
--(axis cs:0.536851292907618,0.544999206159962)
--(axis cs:0.532543390124595,0.556728865243456)
--(axis cs:0.528036225026745,0.567457381723047)
--(axis cs:0.52334794637714,0.577141555649436)
--(axis cs:0.518497432219204,0.585742392272189)
--(axis cs:0.513504213861491,0.593225259057838)
--(axis cs:0.508388397232003,0.599560025142869)
--(axis cs:0.503170581918712,0.604721182660074)
--(axis cs:0.497871778222288,0.608687949449729)
--(axis cs:0.492513322555038,0.611444352742017)
--(axis cs:0.487116791526699,0.612979293473737)
--(axis cs:0.481703915063052,0.613286590980309)
--(axis cs:0.476296488907173,0.612365007883134)
--(axis cs:0.470916286855676,0.610218255072075)
--(axis cs:0.465584973083312,0.606854976763021)
--(axis cs:0.46032401490899,0.602288715690674)
--(axis cs:0.455154596354455,0.596537858576739)
--(axis cs:0.450097532843712,0.589625562093084)
--(axis cs:0.445173187386659,0.581579659617998)
--(axis cs:0.440401388584435,0.572432549160996)
--(axis cs:0.435801350786641,0.562221062907473)
--(axis cs:0.431391596721932,0.550986318908491)
--(axis cs:0.427189882913533,0.538773555512901)
--(axis cs:0.423213128179982,0.52563194920848)
--(axis cs:0.419477345509018,0.511614416605566)
--(axis cs:0.415997577578938,0.49677740136055)
--(axis cs:0.412787836187033,0.481180646897193)
--(axis cs:0.409861045829029,0.464886955840942)
--(axis cs:0.407228991656708,0.447961937134921)
--(axis cs:0.404902272023252,0.430473741855882)
--(axis cs:0.402890255807426,0.412492788793863)
--(axis cs:0.401201044688401,0.39409148090058)
--(axis cs:0.399841440523157,0.375343913748291)
--(axis cs:0.398816917957804,0.356325577173083)
--(axis cs:0.398131602383113,0.33711305130394)
--(axis cs:0.397788253323027,0.317783698201595)
--(axis cs:0.397788253323027,0.298415350348818)
--(axis cs:0.398131602383113,0.279085997246474)
--(axis cs:0.398816917957804,0.259873471377331)
--(axis cs:0.399841440523157,0.240855134802122)
--(axis cs:0.401201044688401,0.222107567649834)
--(axis cs:0.402890255807426,0.203706259756551)
--(axis cs:0.404902272023252,0.185725306694532)
--(axis cs:0.407228991656708,0.168237111415493)
--(axis cs:0.409861045829029,0.151312092709472)
--(axis cs:0.412787836187033,0.13501840165322)
--(axis cs:0.415997577578938,0.119421647189863)
--(axis cs:0.419477345509018,0.104584631944848)
--(axis cs:0.423213128179982,0.0905670993419342)
--(axis cs:0.427189882913533,0.0774254930375126)
--(axis cs:0.431391596721933,0.0652127296419232)
--(axis cs:0.435801350786641,0.0539779856429414)
--(axis cs:0.440401388584435,0.043766499389418)
--(axis cs:0.445173187386659,0.0346193889324158)
--(axis cs:0.450097532843712,0.0265734864573294)
--(axis cs:0.455154596354455,0.0196611899736748)
--(axis cs:0.46032401490899,0.0139103328597396)
--(axis cs:0.465584973083312,0.00934407178739255)
--(axis cs:0.470916286855676,0.00598079347833863)
--(axis cs:0.476296488907173,0.00383404066728021)
--(axis cs:0.481703915063052,0.00291245757010439)
--(axis cs:0.487116791526699,0.00321975507667654)
--(axis cs:0.492513322555038,0.00475469580839638)
--(axis cs:0.497871778222288,0.00751109910068509)
--(axis cs:0.503170581918712,0.0114778658903399)
--(axis cs:0.508388397232003,0.0166390234075447)
--(axis cs:0.513504213861491,0.0229737894925758)
--(axis cs:0.518497432219204,0.0304566562782252)
--(axis cs:0.52334794637714,0.0390574929009783)
--(axis cs:0.528036225026745,0.0487416668273668)
--(axis cs:0.532543390124595,0.0594701833069581)
--(axis cs:0.536851292907618,0.071199842390452)
--(axis cs:0.540942586971752,0.0838834128806292)
--(axis cs:0.5448007981198,0.0974698225157129)
--(axis cs:0.548410390697198,0.111904363619341)
--(axis cs:0.551756830148622,0.127128913389072)
--(axis cs:0.554826641543505,0.143082167936391)
--(axis cs:0.557607463834831,0.159699889135839)
--(axis cs:0.560088099632707,0.176915163289267)
--(axis cs:0.562258560292299,0.194658670563682)
--(axis cs:0.564110106134582,0.212858964117751)
--(axis cs:0.565635281637941,0.231442757793012)
--(axis cs:0.566827945458924,0.250335221211369)
--(axis cs:0.567683295161268,0.269460281090613)
--(axis cs:0.568197886553612,0.288740927564681)
--(axis cs:0.568369647558041,0.308099524275207)
--cycle;
\path [draw=sandybrown24416662, fill=sandybrown24416662, opacity=0.7]
(axis cs:0.925360777630831,0.0244822530166762)
--(axis cs:0.925218474733707,0.0635213084060879)
--(axis cs:0.924792139045877,0.102403167298038)
--(axis cs:0.924083487270683,0.140971266169878)
--(axis cs:0.92309537289871,0.179070304899821)
--(axis cs:0.921831774717785,0.216546872105721)
--(axis cs:0.920297780791803,0.253250062878559)
--(axis cs:0.91849956797289,0.289032086423257)
--(axis cs:0.916444377029401,0.323748861160041)
--(axis cs:0.914140483489898,0.357260594890117)
--(axis cs:0.91159716432052,0.389432347689511)
--(axis cs:0.908824660569912,0.420134575264519)
--(axis cs:0.905834136132142,0.449243650580857)
--(axis cs:0.902637632793632,0.476642361666092)
--(axis cs:0.899248021745142,0.502220383580889)
--(axis cs:0.895678951754021,0.5258747226586)
--(axis cs:0.891944794205442,0.547510131224411)
--(axis cs:0.888060585233906,0.567039491124109)
--(axis cs:0.884041965178034,0.584384164518146)
--(axis cs:0.879905115602445,0.599474310528457)
--(axis cs:0.875666694140303,0.612249166463032)
--(axis cs:0.871343767418912,0.622657292485828)
--(axis cs:0.866953742338429,0.630656778746834)
--(axis cs:0.862514295980422,0.636215414138247)
--(axis cs:0.858043304428499,0.639310815997231)
--(axis cs:0.853558770787632,0.639930520233002)
--(axis cs:0.849078752692003,0.638072031515324)
--(axis cs:0.844621289593289,0.633742833322321)
--(axis cs:0.840204330122155,0.626960357807152)
--(axis cs:0.835845659815454,0.617751915604885)
--(axis cs:0.831562829500143,0.606154585862206)
--(axis cs:0.827373084622306,0.592215066932783)
--(axis cs:0.823293295805821,0.575989488339479)
--(axis cs:0.819339890920315,0.557543184760573)
--(axis cs:0.815528788931933,0.536950432950068)
--(axis cs:0.811875335803278,0.514294152651422)
--(axis cs:0.808394242700635,0.489665572709004)
--(axis cs:0.805099526757298,0.46316386372174)
--(axis cs:0.802004454631522,0.434895738718108)
--(axis cs:0.79912148908637,0.404975023460438)
--(axis cs:0.796462238806569,0.373522198108743)
--(axis cs:0.794037411654429,0.340663912089637)
--(axis cs:0.791856771553064,0.30653247412379)
--(axis cs:0.78992909917052,0.271265319465412)
--(axis cs:0.78826215656312,0.235004456498992)
--(axis cs:0.786862655920408,0.197895894921647)
--(axis cs:0.785736232537527,0.160089057813616)
--(axis cs:0.784887422123875,0.121736179964245)
--(axis cs:0.784319642539407,0.0829916948762224)
--(axis cs:0.784035180032114,0.0440116129163751)
--(axis cs:0.784035180032114,0.00495289311697721)
--(axis cs:0.784319642539407,-0.0340271888428702)
--(axis cs:0.784887422123875,-0.0727716739308924)
--(axis cs:0.785736232537527,-0.111124551780264)
--(axis cs:0.786862655920408,-0.148931388888295)
--(axis cs:0.78826215656312,-0.186039950465639)
--(axis cs:0.78992909917052,-0.22230081343206)
--(axis cs:0.791856771553064,-0.257567968090437)
--(axis cs:0.794037411654429,-0.291699406056284)
--(axis cs:0.796462238806569,-0.324557692075391)
--(axis cs:0.79912148908637,-0.356010517427086)
--(axis cs:0.802004454631522,-0.385931232684755)
--(axis cs:0.805099526757298,-0.414199357688388)
--(axis cs:0.808394242700635,-0.440701066675652)
--(axis cs:0.811875335803278,-0.465329646618069)
--(axis cs:0.815528788931933,-0.487985926916715)
--(axis cs:0.819339890920315,-0.508578678727221)
--(axis cs:0.823293295805821,-0.527024982306127)
--(axis cs:0.827373084622306,-0.543250560899431)
--(axis cs:0.831562829500143,-0.557190079828853)
--(axis cs:0.835845659815454,-0.568787409571533)
--(axis cs:0.840204330122155,-0.5779958517738)
--(axis cs:0.844621289593289,-0.584778327288968)
--(axis cs:0.849078752692003,-0.589107525481972)
--(axis cs:0.853558770787632,-0.59096601419965)
--(axis cs:0.8580433044285,-0.590346309963878)
--(axis cs:0.862514295980422,-0.587250908104895)
--(axis cs:0.866953742338429,-0.581692272713482)
--(axis cs:0.871343767418912,-0.573692786452475)
--(axis cs:0.875666694140303,-0.56328466042968)
--(axis cs:0.879905115602445,-0.550509804495105)
--(axis cs:0.884041965178034,-0.535419658484794)
--(axis cs:0.888060585233906,-0.518074985090757)
--(axis cs:0.891944794205442,-0.498545625191058)
--(axis cs:0.895678951754021,-0.476910216625247)
--(axis cs:0.899248021745142,-0.453255877547536)
--(axis cs:0.902637632793633,-0.427677855632739)
--(axis cs:0.905834136132142,-0.400279144547504)
--(axis cs:0.908824660569912,-0.371170069231167)
--(axis cs:0.91159716432052,-0.340467841656158)
--(axis cs:0.914140483489898,-0.308296088856765)
--(axis cs:0.916444377029401,-0.274784355126689)
--(axis cs:0.91849956797289,-0.240067580389905)
--(axis cs:0.920297780791803,-0.204285556845207)
--(axis cs:0.921831774717785,-0.167582366072368)
--(axis cs:0.92309537289871,-0.130105798866469)
--(axis cs:0.924083487270683,-0.092006760136525)
--(axis cs:0.924792139045877,-0.0534386612646852)
--(axis cs:0.925218474733707,-0.0145568023727355)
--(axis cs:0.925360777630831,0.024482253016676)
--cycle;
\addplot [semithick, seagreen4916384]
table {%
0 0
0.0686536934455336 0.00715986369938762
0.123139750765481 0.0142618411966583
0.130780167919552 0.0550857897474512
0.14164469575638 0.0927490694914908
0.141392726164691 0.115291865054262
0.142098299506721 0.154673061617946
0.145480214129844 0.217050631493459
0.136454450520593 0.27313048119981
0.124967924944315 0.323778044571328
0.143642253658431 0.366275491119077
0.148493470435994 0.404828130235144
0.12161863987027 0.432186062792594
0.146956381975417 0.462749316500369
0.141487719416728 0.508811404358254
0.141082923806964 0.542820995882231
0.137691724634905 0.590035464323733
0.148214616759172 0.62378158995773
0.155095346444814 0.673560906122936
0.15057469860367 0.73485324957088
0.173635411901622 0.776919856946178
0.175027244257732 0.809528538276674
0.17756999511603 0.868744913843871
0.205035961593542 0.898197211984269
0.229314297549606 0.863852738818627
0.239492800989682 0.831289133355498
0.230590859545663 0.787588546563171
0.268407170158449 0.759707149604664
0.308680851237731 0.745744563692423
0.333255064212408 0.718206201580077
0.354645735422533 0.697026633542081
0.386117261594457 0.66782303284261
0.431835012988968 0.662317142913967
0.478080149968698 0.653162360671577
0.521527694795677 0.639538030245448
0.552324068996312 0.643594066481416
0.582791877777723 0.630435950450405
0.624345111520279 0.624268080550723
0.638193401274864 0.584783210014505
0.671575536542194 0.579480194442354
0.715842107049336 0.57011431442495
0.749160783701473 0.603157772242534
0.798518463004286 0.597580928987422
0.835405675611814 0.639364195274936
0.873923543448164 0.636996064060329
0.887607791743674 0.616062280401449
0.900900287763859 0.598632505138226
0.924964393484067 0.566223734045481
0.943534305714604 0.526074215751972
0.948203901496136 0.490110911404399
0.948435791821625 0.451600750127254
0.965261650297445 0.410590199767725
0.991088578838151 0.379289478190813
0.982030138861779 0.31091781469217
0.945352124447374 0.273658203509174
0.958173447693386 0.219827615565246
0.987306660517958 0.172536858824912
0.972322433345069 0.135254038324311
0.981390511288272 0.101587353748378
0.959813858030396 0.042841801784028
1 0
};
\addplot [semithick, sandybrown24416662, opacity=0.7]
table {%
0.257650231540891 0
0.257537694125508 0.0566227800020072
0.25720053502781 0.113017560053604
0.256640111868485 0.168957258279708
0.255858681273799 0.224216625259131
0.254859389788954 0.278573151024497
0.253646261208068 0.331807961031345
0.252224180371768 0.383706697488671
0.250598873497669 0.434060382502099
0.248776885122909 0.482666259554098
0.246765551751616 0.529328609932922
0.244572972313387 0.57385954082278
0.242207975551763 0.616079741881894
0.239680084473992 0.655819207261955
0.236999478005241 0.692917920161676
0.234176950001656 0.727226497157967
0.231223865787314 0.758606789720264
0.228152116390074 0.786932440485897
0.224974070660603 0.812089392056594
0.221702525467383 0.83397634626737
0.218350654168234 0.852505172078482
0.214931953565858 0.867601260448023
0.21146018956097 0.879203824756198
0.20794934172188 0.887266145571591
0.204413546993705 0.891755758773825
0.200867042773879 0.892654586275115
0.197324109583178 0.889959008814349
0.193799013563104 0.883679880530574
0.190305949031171 0.87384248525721
0.186858981325392 0.860486434712975
0.183471990168133 0.843665508999476
0.180158613777368 0.823447440047721
0.176932193950385 0.799913638885536
0.173805722341084 0.773158867824094
0.170791788147167 0.743290858883529
0.167902527417882 0.710429879994117
0.165149574186441 0.674708250719768
0.162544013623869 0.636269809453843
0.160096337402935 0.595269334232713
0.157816401451888 0.551871919499229
0.155713386268114 0.506252311325677
0.153795759951512 0.458594203773015
0.152071244106445 0.409089499219727
0.150546782749564 0.357937535638669
0.149228514348703 0.305344283933418
0.148121747105439 0.251521518566129
0.147230937580834 0.196685964816527
0.146559672750436 0.141058426105695
0.146110655560801 0.0848628948986269
0.145885694045672 0.0283256507656328
0.145885694045672 -0.028325650765633
0.146110655560801 -0.0848628948986271
0.146559672750436 -0.141058426105695
0.147230937580834 -0.196685964816527
0.148121747105439 -0.251521518566129
0.149228514348703 -0.305344283933418
0.150546782749564 -0.357937535638669
0.152071244106445 -0.409089499219726
0.153795759951512 -0.458594203773015
0.155713386268114 -0.506252311325677
0.157816401451888 -0.551871919499229
0.160096337402935 -0.595269334232713
0.162544013623869 -0.636269809453844
0.165149574186441 -0.674708250719768
0.167902527417882 -0.710429879994117
0.170791788147167 -0.743290858883529
0.173805722341084 -0.773158867824094
0.176932193950385 -0.799913638885537
0.180158613777368 -0.823447440047721
0.183471990168133 -0.843665508999476
0.186858981325392 -0.860486434712975
0.190305949031171 -0.87384248525721
0.193799013563104 -0.883679880530574
0.197324109583178 -0.889959008814348
0.200867042773879 -0.892654586275115
0.204413546993705 -0.891755758773825
0.20794934172188 -0.887266145571591
0.21146018956097 -0.879203824756198
0.214931953565858 -0.867601260448023
0.218350654168234 -0.852505172078482
0.221702525467383 -0.83397634626737
0.224974070660603 -0.812089392056594
0.228152116390074 -0.786932440485897
0.231223865787314 -0.758606789720264
0.234176950001656 -0.727226497157967
0.236999478005241 -0.692917920161675
0.239680084473992 -0.655819207261955
0.242207975551763 -0.616079741881894
0.244572972313387 -0.57385954082278
0.246765551751616 -0.529328609932922
0.248776885122909 -0.482666259554098
0.250598873497669 -0.434060382502099
0.252224180371768 -0.383706697488672
0.253646261208068 -0.331807961031344
0.254859389788954 -0.278573151024497
0.255858681273799 -0.224216625259131
0.256640111868485 -0.168957258279708
0.25720053502781 -0.113017560053604
0.257537694125508 -0.0566227800020072
0.257650231540891 -2.18664840217285e-16
};
\addplot [semithick, sandybrown24416662, opacity=0.7]
table {%
0.568369647558041 0.308099524275207
0.568197886553612 0.327458120985732
0.567683295161268 0.346738767459801
0.566827945458924 0.365863827339045
0.565635281637941 0.384756290757402
0.564110106134582 0.403340084432663
0.562258560292299 0.421540377986732
0.560088099632707 0.439283885261147
0.557607463834831 0.456499159414575
0.554826641543505 0.473116880614023
0.551756830148622 0.489070135161342
0.548410390697198 0.504294684931072
0.5448007981198 0.518729226034701
0.540942586971752 0.532315635669785
0.536851292907618 0.544999206159962
0.532543390124595 0.556728865243456
0.528036225026745 0.567457381723047
0.52334794637714 0.577141555649436
0.518497432219204 0.585742392272189
0.513504213861491 0.593225259057838
0.508388397232003 0.599560025142869
0.503170581918712 0.604721182660074
0.497871778222288 0.608687949449729
0.492513322555038 0.611444352742017
0.487116791526699 0.612979293473737
0.481703915063052 0.613286590980309
0.476296488907173 0.612365007883134
0.470916286855676 0.610218255072075
0.465584973083312 0.606854976763021
0.46032401490899 0.602288715690674
0.455154596354455 0.596537858576739
0.450097532843712 0.589625562093084
0.445173187386659 0.581579659617998
0.440401388584435 0.572432549160996
0.435801350786641 0.562221062907473
0.431391596721932 0.550986318908491
0.427189882913533 0.538773555512901
0.423213128179982 0.52563194920848
0.419477345509018 0.511614416605566
0.415997577578938 0.49677740136055
0.412787836187033 0.481180646897193
0.409861045829029 0.464886955840942
0.407228991656708 0.447961937134921
0.404902272023252 0.430473741855882
0.402890255807426 0.412492788793863
0.401201044688401 0.39409148090058
0.399841440523157 0.375343913748291
0.398816917957804 0.356325577173083
0.398131602383113 0.33711305130394
0.397788253323027 0.317783698201595
0.397788253323027 0.298415350348818
0.398131602383113 0.279085997246474
0.398816917957804 0.259873471377331
0.399841440523157 0.240855134802122
0.401201044688401 0.222107567649834
0.402890255807426 0.203706259756551
0.404902272023252 0.185725306694532
0.407228991656708 0.168237111415493
0.409861045829029 0.151312092709472
0.412787836187033 0.13501840165322
0.415997577578938 0.119421647189863
0.419477345509018 0.104584631944848
0.423213128179982 0.0905670993419342
0.427189882913533 0.0774254930375126
0.431391596721933 0.0652127296419232
0.435801350786641 0.0539779856429414
0.440401388584435 0.043766499389418
0.445173187386659 0.0346193889324158
0.450097532843712 0.0265734864573294
0.455154596354455 0.0196611899736748
0.46032401490899 0.0139103328597396
0.465584973083312 0.00934407178739255
0.470916286855676 0.00598079347833863
0.476296488907173 0.00383404066728021
0.481703915063052 0.00291245757010439
0.487116791526699 0.00321975507667654
0.492513322555038 0.00475469580839638
0.497871778222288 0.00751109910068509
0.503170581918712 0.0114778658903399
0.508388397232003 0.0166390234075447
0.513504213861491 0.0229737894925758
0.518497432219204 0.0304566562782252
0.52334794637714 0.0390574929009783
0.528036225026745 0.0487416668273668
0.532543390124595 0.0594701833069581
0.536851292907618 0.071199842390452
0.540942586971752 0.0838834128806292
0.5448007981198 0.0974698225157129
0.548410390697198 0.111904363619341
0.551756830148622 0.127128913389072
0.554826641543505 0.143082167936391
0.557607463834831 0.159699889135839
0.560088099632707 0.176915163289267
0.562258560292299 0.194658670563682
0.564110106134582 0.212858964117751
0.565635281637941 0.231442757793012
0.566827945458924 0.250335221211369
0.567683295161268 0.269460281090613
0.568197886553612 0.288740927564681
0.568369647558041 0.308099524275207
};
\addplot [semithick, sandybrown24416662, opacity=0.7]
table {%
0.925360777630831 0.0244822530166762
0.925218474733707 0.0635213084060879
0.924792139045877 0.102403167298038
0.924083487270683 0.140971266169878
0.92309537289871 0.179070304899821
0.921831774717785 0.216546872105721
0.920297780791803 0.253250062878559
0.91849956797289 0.289032086423257
0.916444377029401 0.323748861160041
0.914140483489898 0.357260594890117
0.91159716432052 0.389432347689511
0.908824660569912 0.420134575264519
0.905834136132142 0.449243650580857
0.902637632793632 0.476642361666092
0.899248021745142 0.502220383580889
0.895678951754021 0.5258747226586
0.891944794205442 0.547510131224411
0.888060585233906 0.567039491124109
0.884041965178034 0.584384164518146
0.879905115602445 0.599474310528457
0.875666694140303 0.612249166463032
0.871343767418912 0.622657292485828
0.866953742338429 0.630656778746834
0.862514295980422 0.636215414138247
0.858043304428499 0.639310815997231
0.853558770787632 0.639930520233002
0.849078752692003 0.638072031515324
0.844621289593289 0.633742833322321
0.840204330122155 0.626960357807152
0.835845659815454 0.617751915604885
0.831562829500143 0.606154585862206
0.827373084622306 0.592215066932783
0.823293295805821 0.575989488339479
0.819339890920315 0.557543184760573
0.815528788931933 0.536950432950068
0.811875335803278 0.514294152651422
0.808394242700635 0.489665572709004
0.805099526757298 0.46316386372174
0.802004454631522 0.434895738718108
0.79912148908637 0.404975023460438
0.796462238806569 0.373522198108743
0.794037411654429 0.340663912089637
0.791856771553064 0.30653247412379
0.78992909917052 0.271265319465412
0.78826215656312 0.235004456498992
0.786862655920408 0.197895894921647
0.785736232537527 0.160089057813616
0.784887422123875 0.121736179964245
0.784319642539407 0.0829916948762224
0.784035180032114 0.0440116129163751
0.784035180032114 0.00495289311697721
0.784319642539407 -0.0340271888428702
0.784887422123875 -0.0727716739308924
0.785736232537527 -0.111124551780264
0.786862655920408 -0.148931388888295
0.78826215656312 -0.186039950465639
0.78992909917052 -0.22230081343206
0.791856771553064 -0.257567968090437
0.794037411654429 -0.291699406056284
0.796462238806569 -0.324557692075391
0.79912148908637 -0.356010517427086
0.802004454631522 -0.385931232684755
0.805099526757298 -0.414199357688388
0.808394242700635 -0.440701066675652
0.811875335803278 -0.465329646618069
0.815528788931933 -0.487985926916715
0.819339890920315 -0.508578678727221
0.823293295805821 -0.527024982306127
0.827373084622306 -0.543250560899431
0.831562829500143 -0.557190079828853
0.835845659815454 -0.568787409571533
0.840204330122155 -0.5779958517738
0.844621289593289 -0.584778327288968
0.849078752692003 -0.589107525481972
0.853558770787632 -0.59096601419965
0.8580433044285 -0.590346309963878
0.862514295980422 -0.587250908104895
0.866953742338429 -0.581692272713482
0.871343767418912 -0.573692786452475
0.875666694140303 -0.56328466042968
0.879905115602445 -0.550509804495105
0.884041965178034 -0.535419658484794
0.888060585233906 -0.518074985090757
0.891944794205442 -0.498545625191058
0.895678951754021 -0.476910216625247
0.899248021745142 -0.453255877547536
0.902637632793633 -0.427677855632739
0.905834136132142 -0.400279144547504
0.908824660569912 -0.371170069231167
0.91159716432052 -0.340467841656158
0.914140483489898 -0.308296088856765
0.916444377029401 -0.274784355126689
0.91849956797289 -0.240067580389905
0.920297780791803 -0.204285556845207
0.921831774717785 -0.167582366072368
0.92309537289871 -0.130105798866469
0.924083487270683 -0.092006760136525
0.924792139045877 -0.0534386612646852
0.925218474733707 -0.0145568023727355
0.925360777630831 0.024482253016676
};
\end{axis}

\end{tikzpicture}

%% file: icra2024/tex/appendix/varying-obstacle-spaces/gamma-0.0-3.tex
\begin{tikzpicture}

\definecolor{sandybrown24416662}{RGB}{244,166,62}
\definecolor{seagreen4916384}{RGB}{49,163,84}

\begin{axis}[
width=0.6\linewidth, 
height=0.6\linewidth,  
tick pos=left,
xmin=0, xmax=1,
ymin=0, ymax=1,
xtick=\empty,
ytick=\empty,
xticklabels={},
yticklabels={},
]
\path [draw=sandybrown24416662, fill=sandybrown24416662, opacity=0.7]
(axis cs:0.231775630189209,0)
--(axis cs:0.231657406306348,0.00965617192296304)
--(axis cs:0.231303210703617,0.019273461849678)
--(axis cs:0.230714469601696,0.0288131443479726)
--(axis cs:0.22989355365321,0.0382368064833977)
--(axis cs:0.228843768396946,0.0475065024945567)
--(axis cs:0.227569340947617,0.0565849065872941)
--(axis cs:0.226075402974759,0.0654354632324948)
--(axis cs:0.224367970039313,0.0740225343622971)
--(axis cs:0.222453917371083,0.0823115428720148)
--(axis cs:0.220340952184609,0.0902691118499329)
--(axis cs:0.218037582644942,0.097863198974352)
--(axis cs:0.215553083608256,0.105063225536708)
--(axis cs:0.212897459275283,0.111840199571238)
--(axis cs:0.210081402907928,0.118166832595396)
--(axis cs:0.207116253771271,0.12401764949094)
--(axis cs:0.204013951474355,0.12936909108324)
--(axis cs:0.200786987893592,0.134199609005763)
--(axis cs:0.197448356872384,0.138489752467735)
--(axis cs:0.194011501899505,0.142222246575615)
--(axis cs:0.190490261976913,0.145382061892991)
--(axis cs:0.186898815894979,0.147956474958816)
--(axis cs:0.183251625139495,0.149935119520298)
--(axis cs:0.179563375660379,0.151310028274139)
--(axis cs:0.175848918736541,0.15207566494805)
--(axis cs:0.172123211175017,0.152228946593375)
--(axis cs:0.168401255085193,0.151769255999025)
--(axis cs:0.164698037470602,0.150698444176781)
--(axis cs:0.161028469881549,0.149020822907914)
--(axis cs:0.157407328371564,0.146743147381168)
--(axis cs:0.15384919399945,0.143874588991997)
--(axis cs:0.150368394116501,0.140426698412595)
--(axis cs:0.14697894467532,0.136413359081414)
--(axis cs:0.14369449379252,0.131850731299457)
--(axis cs:0.140528266792581,0.126757187158451)
--(axis cs:0.137493012954131,0.121153236562917)
--(axis cs:0.134600954173103,0.115061444644025)
--(axis cs:0.131863735749475,0.10850634089777)
--(axis cs:0.129292379495754,0.101514320413358)
--(axis cs:0.126897239356031,0.0941135375894873)
--(axis cs:0.124687959714297,0.0863337927665307)
--(axis cs:0.122673436559919,0.0782064122310774)
--(axis cs:0.120861781666626,0.0697641220760355)
--(axis cs:0.11926028992926,0.0610409164242056)
--(axis cs:0.117875409989813,0.0520719205459465)
--(axis cs:0.116712718271014,0.0428932494221087)
--(axis cs:0.115776896522052,0.0335418623217532)
--(axis cs:0.115071712966816,0.0240554139802193)
--(axis cs:0.114600007130594,0.0144721029767955)
--(axis cs:0.114363678406302,0.00483051792252278)
--(axis cs:0.114363678406302,-0.00483051792252281)
--(axis cs:0.114600007130594,-0.0144721029767955)
--(axis cs:0.115071712966816,-0.0240554139802193)
--(axis cs:0.115776896522052,-0.0335418623217532)
--(axis cs:0.116712718271014,-0.0428932494221087)
--(axis cs:0.117875409989813,-0.0520719205459464)
--(axis cs:0.11926028992926,-0.0610409164242056)
--(axis cs:0.120861781666626,-0.0697641220760355)
--(axis cs:0.122673436559919,-0.0782064122310775)
--(axis cs:0.124687959714297,-0.0863337927665307)
--(axis cs:0.126897239356031,-0.0941135375894873)
--(axis cs:0.129292379495754,-0.101514320413358)
--(axis cs:0.131863735749475,-0.10850634089777)
--(axis cs:0.134600954173103,-0.115061444644025)
--(axis cs:0.137493012954131,-0.121153236562917)
--(axis cs:0.140528266792581,-0.126757187158451)
--(axis cs:0.14369449379252,-0.131850731299457)
--(axis cs:0.14697894467532,-0.136413359081414)
--(axis cs:0.150368394116501,-0.140426698412595)
--(axis cs:0.15384919399945,-0.143874588991997)
--(axis cs:0.157407328371565,-0.146743147381168)
--(axis cs:0.161028469881549,-0.149020822907914)
--(axis cs:0.164698037470602,-0.150698444176781)
--(axis cs:0.168401255085193,-0.151769255999025)
--(axis cs:0.172123211175017,-0.152228946593375)
--(axis cs:0.175848918736541,-0.15207566494805)
--(axis cs:0.179563375660379,-0.151310028274139)
--(axis cs:0.183251625139494,-0.149935119520298)
--(axis cs:0.186898815894979,-0.147956474958816)
--(axis cs:0.190490261976913,-0.145382061892991)
--(axis cs:0.194011501899505,-0.142222246575615)
--(axis cs:0.197448356872384,-0.138489752467735)
--(axis cs:0.200786987893592,-0.134199609005763)
--(axis cs:0.204013951474355,-0.12936909108324)
--(axis cs:0.207116253771271,-0.12401764949094)
--(axis cs:0.210081402907928,-0.118166832595396)
--(axis cs:0.212897459275283,-0.111840199571238)
--(axis cs:0.215553083608256,-0.105063225536708)
--(axis cs:0.218037582644942,-0.0978631989743521)
--(axis cs:0.220340952184609,-0.0902691118499329)
--(axis cs:0.222453917371083,-0.0823115428720147)
--(axis cs:0.224367970039313,-0.0740225343622971)
--(axis cs:0.226075402974759,-0.0654354632324948)
--(axis cs:0.227569340947617,-0.0565849065872941)
--(axis cs:0.228843768396946,-0.0475065024945567)
--(axis cs:0.22989355365321,-0.0382368064833977)
--(axis cs:0.230714469601696,-0.0288131443479725)
--(axis cs:0.231303210703617,-0.019273461849678)
--(axis cs:0.231657406306348,-0.00965617192296304)
--(axis cs:0.231775630189209,-3.72900322197267e-17)
--cycle;
\path [draw=sandybrown24416662, fill=sandybrown24416662, opacity=0.7]
(axis cs:0.538342895953703,0.532696642255931)
--(axis cs:0.538187144864051,0.549296447500034)
--(axis cs:0.537720518749762,0.565829411187218)
--(axis cs:0.536944896549506,0.58222896090792)
--(axis cs:0.535863401420131,0.598429061463533)
--(axis cs:0.534480388160816,0.614364480766845)
--(axis cs:0.532801425677833,0.629971052508628)
--(axis cs:0.530833274560518,0.645185934532721)
--(axis cs:0.528583859858777,0.659947861879209)
--(axis cs:0.526062239171694,0.6741973934768)
--(axis cs:0.523278566175788,0.687877151491033)
--(axis cs:0.520244049739733,0.70093205236457)
--(axis cs:0.516970908790202,0.713309528619239)
--(axis cs:0.513472323110566,0.724959740526716)
--(axis cs:0.509762380270551,0.735835776795519)
--(axis cs:0.505856018900573,0.745893843466222)
--(axis cs:0.501768968539144,0.755093440254276)
--(axis cs:0.49751768629557,0.763397523630354)
--(axis cs:0.493119290582982,0.770772655981576)
--(axis cs:0.488591492188528,0.77718914025297)
--(axis cs:0.483952522958285,0.782621139527034)
--(axis cs:0.479221062384048,0.787046781059881)
--(axis cs:0.474416162387609,0.790448244355062)
--(axis cs:0.469557170605395,0.79281183292041)
--(axis cs:0.464663652482361,0.794128029418989)
--(axis cs:0.459755312488853,0.794391533992041)
--(axis cs:0.454851914777656,0.793601285599653)
--(axis cs:0.44997320360073,0.791760466293189)
--(axis cs:0.445138823806073,0.788876488402288)
--(axis cs:0.440368241734854,0.784960964688037)
--(axis cs:0.435680666837328,0.780029661582472)
--(axis cs:0.431094974323155,0.774102435702727)
--(axis cs:0.426629629157598,0.767203153895442)
--(axis cs:0.422302611709625,0.759359597133398)
--(axis cs:0.418131345351304,0.750603348651339)
--(axis cs:0.414132626300036,0.74096966677144)
--(axis cs:0.41032255598612,0.730497342930483)
--(axis cs:0.406716476217968,0.719228545480435)
--(axis cs:0.403328907406064,0.707208649891374)
--(axis cs:0.400173490094391,0.694486056040491)
--(axis cs:0.397262930034779,0.681111993322864)
--(axis cs:0.394608947025324,0.667140314368763)
--(axis cs:0.392222227718898,0.65262727819812)
--(axis cs:0.39011238259177,0.637631323685308)
--(axis cs:0.388287907245611,0.622212834246422)
--(axis cs:0.386756148198703,0.606433894696573)
--(axis cs:0.385523273304102,0.590358041256249)
--(axis cs:0.384594246913874,0.574050005713382)
--(axis cs:0.383972809889398,0.557575454771284)
--(axis cs:0.383661464538239,0.541000725632009)
--(axis cs:0.383661464538239,0.524392558879852)
--(axis cs:0.383972809889398,0.507817829740577)
--(axis cs:0.384594246913874,0.491343278798479)
--(axis cs:0.385523273304102,0.475035243255612)
--(axis cs:0.386756148198703,0.458959389815288)
--(axis cs:0.388287907245611,0.443180450265439)
--(axis cs:0.39011238259177,0.427761960826553)
--(axis cs:0.392222227718898,0.412766006313742)
--(axis cs:0.394608947025324,0.398252970143099)
--(axis cs:0.397262930034779,0.384281291188998)
--(axis cs:0.400173490094391,0.37090722847137)
--(axis cs:0.403328907406064,0.358184634620487)
--(axis cs:0.406716476217968,0.346164739031427)
--(axis cs:0.41032255598612,0.334895941581378)
--(axis cs:0.414132626300036,0.324423617740422)
--(axis cs:0.418131345351304,0.314789935860523)
--(axis cs:0.422302611709625,0.306033687378464)
--(axis cs:0.426629629157599,0.298190130616419)
--(axis cs:0.431094974323155,0.291290848809134)
--(axis cs:0.435680666837328,0.28536362292939)
--(axis cs:0.440368241734854,0.280432319823825)
--(axis cs:0.445138823806073,0.276516796109573)
--(axis cs:0.44997320360073,0.273632818218673)
--(axis cs:0.454851914777656,0.271791998912208)
--(axis cs:0.459755312488853,0.27100175051982)
--(axis cs:0.464663652482361,0.271265255092872)
--(axis cs:0.469557170605395,0.272581451591451)
--(axis cs:0.474416162387609,0.2749450401568)
--(axis cs:0.479221062384048,0.27834650345198)
--(axis cs:0.483952522958285,0.282772144984827)
--(axis cs:0.488591492188528,0.288204144258891)
--(axis cs:0.493119290582982,0.294620628530286)
--(axis cs:0.49751768629557,0.301995760881508)
--(axis cs:0.501768968539144,0.310299844257586)
--(axis cs:0.505856018900573,0.319499441045639)
--(axis cs:0.509762380270551,0.329557507716343)
--(axis cs:0.513472323110566,0.340433543985146)
--(axis cs:0.516970908790202,0.352083755892623)
--(axis cs:0.520244049739733,0.364461232147291)
--(axis cs:0.523278566175788,0.377516133020828)
--(axis cs:0.526062239171694,0.391195891035061)
--(axis cs:0.528583859858777,0.405445422632652)
--(axis cs:0.530833274560518,0.420207349979141)
--(axis cs:0.532801425677833,0.435422232003233)
--(axis cs:0.534480388160816,0.451028803745017)
--(axis cs:0.535863401420131,0.466964223048329)
--(axis cs:0.536944896549506,0.483164323603942)
--(axis cs:0.537720518749762,0.499563873324644)
--(axis cs:0.538187144864051,0.516096837011827)
--(axis cs:0.538342895953703,0.532696642255931)
--cycle;
\path [draw=sandybrown24416662, fill=sandybrown24416662, opacity=0.7]
(axis cs:0.910682445947848,0.0718552235052037)
--(axis cs:0.910533330323642,0.114348447465358)
--(axis cs:0.910086583887014,0.156670566210344)
--(axis cs:0.909344005528188,0.198651163505302)
--(axis cs:0.908308585348127,0.240121198301717)
--(axis cs:0.906984492618449,0.280913685406054)
--(axis cs:0.905377058993249,0.320864367870198)
--(axis cs:0.9034927570404,0.359812378396209)
--(axis cs:0.901339174178814,0.397600887092159)
--(axis cs:0.898924982126581,0.434077732970745)
--(axis cs:0.896259901983021,0.469096036647891)
--(axis cs:0.893354665085249,0.502514791774182)
--(axis cs:0.890220969796867,0.534199432817665)
--(axis cs:0.886871434402783,0.564022376911753)
--(axis cs:0.883319546299832,0.591863537586404)
--(axis cs:0.879579607687787,0.617610808313944)
--(axis cs:0.875666677979451,0.641160513922492)
--(axis cs:0.871596513161716,0.662417828059285)
--(axis cs:0.867385502351762,0.681297155022929)
--(axis cs:0.863050601803872,0.697722474427084)
--(axis cs:0.858609266632576,0.711627647307722)
--(axis cs:0.854079380527066,0.722956682441399)
--(axis cs:0.849479183739896,0.731663961802142)
--(axis cs:0.844827199639916,0.737714424249146)
--(axis cs:0.840142160125207,0.74108370670561)
--(axis cs:0.835442930196341,0.741758242260249)
--(axis cs:0.830748431993676,0.739735314796457)
--(axis cs:0.826077568604582,0.735023069929153)
--(axis cs:0.821449147947373,0.727640482205258)
--(axis cs:0.816881807038464,0.717617278699898)
--(axis cs:0.81239393694768,0.704993819315963)
--(axis cs:0.808003608743913,0.689820934269021)
--(axis cs:0.803728500729298,0.672159719411979)
--(axis cs:0.799585827254939,0.652081290223652)
--(axis cs:0.795592269404785,0.629666495451825)
--(axis cs:0.791763907826798,0.605005591563881)
--(axis cs:0.788116157981857,0.578197879315865)
--(axis cs:0.784663708071135,0.549351303903385)
--(axis cs:0.781420459891903,0.518582020304407)
--(axis cs:0.778399472859895,0.486013925564158)
--(axis cs:0.775612911423649,0.451778159905452)
--(axis cs:0.77307199608257,0.416012578673294)
--(axis cs:0.770786958205937,0.378861197240069)
--(axis cs:0.768766998834789,0.340473611106458)
--(axis cs:0.76702025163258,0.301004393533163)
--(axis cs:0.765553750133792,0.260612473128938)
--(axis cs:0.764373399422366,0.219460493901173)
--(axis cs:0.763483952354021,0.177714160345883)
--(axis cs:0.762888990418176,0.135541570214188)
--(axis cs:0.762590909316558,0.0931125376419961)
--(axis cs:0.762590909316558,0.0505979093684111)
--(axis cs:0.762888990418176,0.00816887679621876)
--(axis cs:0.763483952354021,-0.034003713335476)
--(axis cs:0.764373399422366,-0.0757500468907652)
--(axis cs:0.765553750133792,-0.116902026118531)
--(axis cs:0.76702025163258,-0.157293946522756)
--(axis cs:0.768766998834789,-0.196763164096051)
--(axis cs:0.770786958205937,-0.235150750229661)
--(axis cs:0.77307199608257,-0.272302131662887)
--(axis cs:0.775612911423649,-0.308067712895044)
--(axis cs:0.778399472859895,-0.342303478553751)
--(axis cs:0.781420459891903,-0.374871573294)
--(axis cs:0.784663708071135,-0.405640856892977)
--(axis cs:0.788116157981856,-0.434487432305457)
--(axis cs:0.791763907826798,-0.461295144553474)
--(axis cs:0.795592269404785,-0.485956048441417)
--(axis cs:0.799585827254939,-0.508370843213245)
--(axis cs:0.803728500729298,-0.528449272401572)
--(axis cs:0.808003608743913,-0.546110487258613)
--(axis cs:0.81239393694768,-0.561283372305556)
--(axis cs:0.816881807038464,-0.573906831689491)
--(axis cs:0.821449147947373,-0.58393003519485)
--(axis cs:0.826077568604582,-0.591312622918745)
--(axis cs:0.830748431993676,-0.59602486778605)
--(axis cs:0.835442930196341,-0.598047795249842)
--(axis cs:0.840142160125207,-0.597373259695202)
--(axis cs:0.844827199639916,-0.594003977238738)
--(axis cs:0.849479183739896,-0.587953514791735)
--(axis cs:0.854079380527066,-0.579246235430991)
--(axis cs:0.858609266632576,-0.567917200297315)
--(axis cs:0.863050601803872,-0.554012027416676)
--(axis cs:0.867385502351762,-0.537586708012522)
--(axis cs:0.871596513161716,-0.518707381048877)
--(axis cs:0.875666677979451,-0.497450066912085)
--(axis cs:0.879579607687787,-0.473900361303537)
--(axis cs:0.883319546299832,-0.448153090575997)
--(axis cs:0.886871434402783,-0.420311929901346)
--(axis cs:0.890220969796867,-0.390488985807257)
--(axis cs:0.893354665085249,-0.358804344763775)
--(axis cs:0.896259901983021,-0.325385589637484)
--(axis cs:0.898924982126581,-0.290367285960338)
--(axis cs:0.901339174178814,-0.253890440081751)
--(axis cs:0.9034927570404,-0.216101931385802)
--(axis cs:0.905377058993249,-0.17715392085979)
--(axis cs:0.906984492618449,-0.137203238395647)
--(axis cs:0.908308585348127,-0.0964107512913101)
--(axis cs:0.909344005528188,-0.0549407164948946)
--(axis cs:0.910086583887014,-0.0129601191999361)
--(axis cs:0.910533330323642,0.0293619995450492)
--(axis cs:0.910682445947848,0.0718552235052035)
--cycle;
\addplot [semithick, seagreen4916384]
table {%
0 0
0.0859412935021262 0.0465494189302596
0.119259602240981 0.0897600288520093
0.140117575103063 0.139781342291709
0.186822615897508 0.163182103237869
0.215425045959832 0.140832832245309
0.249058498739271 0.10213877177644
0.245746267496302 0.053758792479179
0.254137735072834 0.00682968462660728
0.269415057165449 0.00185161956697188
0.292741832312316 0.00308810691658497
0.319583492913601 0.00116313224469688
0.34010036774253 0.00335386316731882
0.38538762297856 0.0015381441125463
0.42081506068664 0.00245020745665639
0.44497493774051 0.000880271284412571
0.469174448852428 0.000530031993126826
0.48100665246506 0.00167628322397068
0.513961054329787 0.0105324001936366
0.536249122624091 0.00966618960471205
0.592848159341136 0.00867978860674855
0.628210744112179 0.00111551150670476
0.676476839206911 0.00345761847738593
0.69983983736764 0.00291080766235009
0.753751029344747 0.010426180875858
0.731685628130066 0.0933952090552266
0.755026587339414 0.119280333114669
0.751419448853149 0.172002363265596
0.758358151589533 0.227649993595389
0.76394312117633 0.292152510903742
0.766540943419611 0.346865101591336
0.761984097193497 0.383817767428041
0.753017600334763 0.420978418792954
0.767617744472319 0.472562909102689
0.748688133181422 0.527012025132842
0.75777586504401 0.553238176436014
0.781018951457322 0.590127690663199
0.789199744935624 0.628474155879346
0.758631256111561 0.685695010159881
0.804463954629502 0.708659895237291
0.840490131842164 0.751419216156917
0.883631595163531 0.723059872026144
0.878489818707948 0.692505199719623
0.891131387627615 0.641553903250158
0.904931577010529 0.583210409537343
0.908099408932041 0.530069191017722
0.923903017975337 0.489135412993151
0.931713183256398 0.451292082151056
0.920363868946483 0.409992392088285
0.929752397867104 0.353328310371823
0.937543243999796 0.297194779534104
0.913241756692203 0.252422575275794
0.933554567580112 0.192973985968325
0.917641804996762 0.164390491766806
0.933457743996526 0.111791142272493
0.956146624468737 0.0707472926611881
0.944535964257008 0.038425602517749
0.918379905767665 0.00705970201602354
0.939481141039463 0.00044423631645738
0.960390340740608 0.00362033096393804
0.984419166326454 0.00371578523862819
1 0
};
\addplot [semithick, sandybrown24416662, opacity=0.7]
table {%
0.231775630189209 0
0.231657406306348 0.00965617192296304
0.231303210703617 0.019273461849678
0.230714469601696 0.0288131443479726
0.22989355365321 0.0382368064833977
0.228843768396946 0.0475065024945567
0.227569340947617 0.0565849065872941
0.226075402974759 0.0654354632324948
0.224367970039313 0.0740225343622971
0.222453917371083 0.0823115428720148
0.220340952184609 0.0902691118499329
0.218037582644942 0.097863198974352
0.215553083608256 0.105063225536708
0.212897459275283 0.111840199571238
0.210081402907928 0.118166832595396
0.207116253771271 0.12401764949094
0.204013951474355 0.12936909108324
0.200786987893592 0.134199609005763
0.197448356872384 0.138489752467735
0.194011501899505 0.142222246575615
0.190490261976913 0.145382061892991
0.186898815894979 0.147956474958816
0.183251625139495 0.149935119520298
0.179563375660379 0.151310028274139
0.175848918736541 0.15207566494805
0.172123211175017 0.152228946593375
0.168401255085193 0.151769255999025
0.164698037470602 0.150698444176781
0.161028469881549 0.149020822907914
0.157407328371564 0.146743147381168
0.15384919399945 0.143874588991997
0.150368394116501 0.140426698412595
0.14697894467532 0.136413359081414
0.14369449379252 0.131850731299457
0.140528266792581 0.126757187158451
0.137493012954131 0.121153236562917
0.134600954173103 0.115061444644025
0.131863735749475 0.10850634089777
0.129292379495754 0.101514320413358
0.126897239356031 0.0941135375894873
0.124687959714297 0.0863337927665307
0.122673436559919 0.0782064122310774
0.120861781666626 0.0697641220760355
0.11926028992926 0.0610409164242056
0.117875409989813 0.0520719205459465
0.116712718271014 0.0428932494221087
0.115776896522052 0.0335418623217532
0.115071712966816 0.0240554139802193
0.114600007130594 0.0144721029767955
0.114363678406302 0.00483051792252278
0.114363678406302 -0.00483051792252281
0.114600007130594 -0.0144721029767955
0.115071712966816 -0.0240554139802193
0.115776896522052 -0.0335418623217532
0.116712718271014 -0.0428932494221087
0.117875409989813 -0.0520719205459464
0.11926028992926 -0.0610409164242056
0.120861781666626 -0.0697641220760355
0.122673436559919 -0.0782064122310775
0.124687959714297 -0.0863337927665307
0.126897239356031 -0.0941135375894873
0.129292379495754 -0.101514320413358
0.131863735749475 -0.10850634089777
0.134600954173103 -0.115061444644025
0.137493012954131 -0.121153236562917
0.140528266792581 -0.126757187158451
0.14369449379252 -0.131850731299457
0.14697894467532 -0.136413359081414
0.150368394116501 -0.140426698412595
0.15384919399945 -0.143874588991997
0.157407328371565 -0.146743147381168
0.161028469881549 -0.149020822907914
0.164698037470602 -0.150698444176781
0.168401255085193 -0.151769255999025
0.172123211175017 -0.152228946593375
0.175848918736541 -0.15207566494805
0.179563375660379 -0.151310028274139
0.183251625139494 -0.149935119520298
0.186898815894979 -0.147956474958816
0.190490261976913 -0.145382061892991
0.194011501899505 -0.142222246575615
0.197448356872384 -0.138489752467735
0.200786987893592 -0.134199609005763
0.204013951474355 -0.12936909108324
0.207116253771271 -0.12401764949094
0.210081402907928 -0.118166832595396
0.212897459275283 -0.111840199571238
0.215553083608256 -0.105063225536708
0.218037582644942 -0.0978631989743521
0.220340952184609 -0.0902691118499329
0.222453917371083 -0.0823115428720147
0.224367970039313 -0.0740225343622971
0.226075402974759 -0.0654354632324948
0.227569340947617 -0.0565849065872941
0.228843768396946 -0.0475065024945567
0.22989355365321 -0.0382368064833977
0.230714469601696 -0.0288131443479725
0.231303210703617 -0.019273461849678
0.231657406306348 -0.00965617192296304
0.231775630189209 -3.72900322197267e-17
};
\addplot [semithick, sandybrown24416662, opacity=0.7]
table {%
0.538342895953703 0.532696642255931
0.538187144864051 0.549296447500034
0.537720518749762 0.565829411187218
0.536944896549506 0.58222896090792
0.535863401420131 0.598429061463533
0.534480388160816 0.614364480766845
0.532801425677833 0.629971052508628
0.530833274560518 0.645185934532721
0.528583859858777 0.659947861879209
0.526062239171694 0.6741973934768
0.523278566175788 0.687877151491033
0.520244049739733 0.70093205236457
0.516970908790202 0.713309528619239
0.513472323110566 0.724959740526716
0.509762380270551 0.735835776795519
0.505856018900573 0.745893843466222
0.501768968539144 0.755093440254276
0.49751768629557 0.763397523630354
0.493119290582982 0.770772655981576
0.488591492188528 0.77718914025297
0.483952522958285 0.782621139527034
0.479221062384048 0.787046781059881
0.474416162387609 0.790448244355062
0.469557170605395 0.79281183292041
0.464663652482361 0.794128029418989
0.459755312488853 0.794391533992041
0.454851914777656 0.793601285599653
0.44997320360073 0.791760466293189
0.445138823806073 0.788876488402288
0.440368241734854 0.784960964688037
0.435680666837328 0.780029661582472
0.431094974323155 0.774102435702727
0.426629629157598 0.767203153895442
0.422302611709625 0.759359597133398
0.418131345351304 0.750603348651339
0.414132626300036 0.74096966677144
0.41032255598612 0.730497342930483
0.406716476217968 0.719228545480435
0.403328907406064 0.707208649891374
0.400173490094391 0.694486056040491
0.397262930034779 0.681111993322864
0.394608947025324 0.667140314368763
0.392222227718898 0.65262727819812
0.39011238259177 0.637631323685308
0.388287907245611 0.622212834246422
0.386756148198703 0.606433894696573
0.385523273304102 0.590358041256249
0.384594246913874 0.574050005713382
0.383972809889398 0.557575454771284
0.383661464538239 0.541000725632009
0.383661464538239 0.524392558879852
0.383972809889398 0.507817829740577
0.384594246913874 0.491343278798479
0.385523273304102 0.475035243255612
0.386756148198703 0.458959389815288
0.388287907245611 0.443180450265439
0.39011238259177 0.427761960826553
0.392222227718898 0.412766006313742
0.394608947025324 0.398252970143099
0.397262930034779 0.384281291188998
0.400173490094391 0.37090722847137
0.403328907406064 0.358184634620487
0.406716476217968 0.346164739031427
0.41032255598612 0.334895941581378
0.414132626300036 0.324423617740422
0.418131345351304 0.314789935860523
0.422302611709625 0.306033687378464
0.426629629157599 0.298190130616419
0.431094974323155 0.291290848809134
0.435680666837328 0.28536362292939
0.440368241734854 0.280432319823825
0.445138823806073 0.276516796109573
0.44997320360073 0.273632818218673
0.454851914777656 0.271791998912208
0.459755312488853 0.27100175051982
0.464663652482361 0.271265255092872
0.469557170605395 0.272581451591451
0.474416162387609 0.2749450401568
0.479221062384048 0.27834650345198
0.483952522958285 0.282772144984827
0.488591492188528 0.288204144258891
0.493119290582982 0.294620628530286
0.49751768629557 0.301995760881508
0.501768968539144 0.310299844257586
0.505856018900573 0.319499441045639
0.509762380270551 0.329557507716343
0.513472323110566 0.340433543985146
0.516970908790202 0.352083755892623
0.520244049739733 0.364461232147291
0.523278566175788 0.377516133020828
0.526062239171694 0.391195891035061
0.528583859858777 0.405445422632652
0.530833274560518 0.420207349979141
0.532801425677833 0.435422232003233
0.534480388160816 0.451028803745017
0.535863401420131 0.466964223048329
0.536944896549506 0.483164323603942
0.537720518749762 0.499563873324644
0.538187144864051 0.516096837011827
0.538342895953703 0.532696642255931
};
\addplot [semithick, sandybrown24416662, opacity=0.7]
table {%
0.910682445947848 0.0718552235052037
0.910533330323642 0.114348447465358
0.910086583887014 0.156670566210344
0.909344005528188 0.198651163505302
0.908308585348127 0.240121198301717
0.906984492618449 0.280913685406054
0.905377058993249 0.320864367870198
0.9034927570404 0.359812378396209
0.901339174178814 0.397600887092159
0.898924982126581 0.434077732970745
0.896259901983021 0.469096036647891
0.893354665085249 0.502514791774182
0.890220969796867 0.534199432817665
0.886871434402783 0.564022376911753
0.883319546299832 0.591863537586404
0.879579607687787 0.617610808313944
0.875666677979451 0.641160513922492
0.871596513161716 0.662417828059285
0.867385502351762 0.681297155022929
0.863050601803872 0.697722474427084
0.858609266632576 0.711627647307722
0.854079380527066 0.722956682441399
0.849479183739896 0.731663961802142
0.844827199639916 0.737714424249146
0.840142160125207 0.74108370670561
0.835442930196341 0.741758242260249
0.830748431993676 0.739735314796457
0.826077568604582 0.735023069929153
0.821449147947373 0.727640482205258
0.816881807038464 0.717617278699898
0.81239393694768 0.704993819315963
0.808003608743913 0.689820934269021
0.803728500729298 0.672159719411979
0.799585827254939 0.652081290223652
0.795592269404785 0.629666495451825
0.791763907826798 0.605005591563881
0.788116157981857 0.578197879315865
0.784663708071135 0.549351303903385
0.781420459891903 0.518582020304407
0.778399472859895 0.486013925564158
0.775612911423649 0.451778159905452
0.77307199608257 0.416012578673294
0.770786958205937 0.378861197240069
0.768766998834789 0.340473611106458
0.76702025163258 0.301004393533163
0.765553750133792 0.260612473128938
0.764373399422366 0.219460493901173
0.763483952354021 0.177714160345883
0.762888990418176 0.135541570214188
0.762590909316558 0.0931125376419961
0.762590909316558 0.0505979093684111
0.762888990418176 0.00816887679621876
0.763483952354021 -0.034003713335476
0.764373399422366 -0.0757500468907652
0.765553750133792 -0.116902026118531
0.76702025163258 -0.157293946522756
0.768766998834789 -0.196763164096051
0.770786958205937 -0.235150750229661
0.77307199608257 -0.272302131662887
0.775612911423649 -0.308067712895044
0.778399472859895 -0.342303478553751
0.781420459891903 -0.374871573294
0.784663708071135 -0.405640856892977
0.788116157981856 -0.434487432305457
0.791763907826798 -0.461295144553474
0.795592269404785 -0.485956048441417
0.799585827254939 -0.508370843213245
0.803728500729298 -0.528449272401572
0.808003608743913 -0.546110487258613
0.81239393694768 -0.561283372305556
0.816881807038464 -0.573906831689491
0.821449147947373 -0.58393003519485
0.826077568604582 -0.591312622918745
0.830748431993676 -0.59602486778605
0.835442930196341 -0.598047795249842
0.840142160125207 -0.597373259695202
0.844827199639916 -0.594003977238738
0.849479183739896 -0.587953514791735
0.854079380527066 -0.579246235430991
0.858609266632576 -0.567917200297315
0.863050601803872 -0.554012027416676
0.867385502351762 -0.537586708012522
0.871596513161716 -0.518707381048877
0.875666677979451 -0.497450066912085
0.879579607687787 -0.473900361303537
0.883319546299832 -0.448153090575997
0.886871434402783 -0.420311929901346
0.890220969796867 -0.390488985807257
0.893354665085249 -0.358804344763775
0.896259901983021 -0.325385589637484
0.898924982126581 -0.290367285960338
0.901339174178814 -0.253890440081751
0.9034927570404 -0.216101931385802
0.905377058993249 -0.17715392085979
0.906984492618449 -0.137203238395647
0.908308585348127 -0.0964107512913101
0.909344005528188 -0.0549407164948946
0.910086583887014 -0.0129601191999361
0.910533330323642 0.0293619995450492
0.910682445947848 0.0718552235052035
};
\end{axis}

\end{tikzpicture}

%% file: icra2024/tex/appendix/gamma-normalized-cost.tex
\begin{tikzpicture}

\definecolor{darkgray176}{RGB}{176,176,176}
\definecolor{seagreen4916384}{RGB}{49,163,84}

\begin{axis}[
width=0.75\linewidth, 
log basis x={10},
tick align=outside,
tick pos=left,
x grid style={darkgray176},
xlabel={\(\displaystyle \gamma\)},
xmajorgrids,
xmin=0.0001, xmax=1,
xmode=log,
xtick style={color=black},
xtick={1e-05,0.0001,0.001,0.01,0.1,1,10},
xticklabels={
  \(\displaystyle {10^{-5}}\),
  \(\displaystyle {10^{-4}}\),
  \(\displaystyle {10^{-3}}\),
  \(\displaystyle {10^{-2}}\),
  \(\displaystyle {10^{-1}}\),
  \(\displaystyle {10^{0}}\),
  \(\displaystyle {10^{1}}\)
},
y grid style={darkgray176},
ylabel={\(\displaystyle \overline{{C}}(\hat{{\sigma}}^*)\)},
ymajorgrids,
ymin=0.99, ymax=5,
ytick style={color=black}
]
\path [draw=seagreen4916384, fill=seagreen4916384, opacity=0.4]
--(axis cs:0.0001,1.00127545)
--(axis cs:0.001,1.01587516)
--(axis cs:0.01,1.03247676)
--(axis cs:0.05,1.15634403)
--(axis cs:0.1,1.32802248)
--(axis cs:0.2,1.60604588)
--(axis cs:0.3,1.81209518)
--(axis cs:0.5,2.3158804)
--(axis cs:0.7,3.24755313)
--(axis cs:1,3.82404074)
--(axis cs:1,5.32736544)
--(axis cs:1,5.32736544)
--(axis cs:0.7,3.98022879)
--(axis cs:0.5,3.77377058)
--(axis cs:0.3,2.40175802)
--(axis cs:0.2,2.07224274)
--(axis cs:0.1,1.4583281)
--(axis cs:0.05,1.22278841)
--(axis cs:0.01,1.04965766)
--(axis cs:0.001,1.0472612)
--(axis cs:0.0001,1.02095637)
--cycle;

\addplot [thick, seagreen4916384, mark=asterisk, mark size=3, mark options={solid}]
table {%
0.0001 1.01111591
0.001 1.03156818
0.01 1.04106721
0.05 1.18956622
0.1 1.39317529
0.2 1.83914431
0.3 2.1069266
0.5 3.04482549
0.7 3.61389096
1 4.57570309
};
\end{axis}

\end{tikzpicture}

%% file: icra2024/tex/appendix/varying-gammas/gamma-0.0.tex
\begin{tikzpicture}

\definecolor{sandybrown24416662}{RGB}{244,166,62}
\definecolor{seagreen4916384}{RGB}{49,163,84}

\begin{axis}[
width=0.33\linewidth, 
height=0.33\linewidth, 
tick pos=left,
xlabel={x},
xmin=0, xmax=1,
xtick style={color=black},
ylabel={y},
ymin=0, ymax=1,
ytick style={color=black},
]
\path [draw=sandybrown24416662, fill=sandybrown24416662, opacity=0.7]
(axis cs:0.25,0)
--(axis cs:0.249899333823594,0.0380543517939387)
--(axis cs:0.24959774064154,0.0759554721442496)
--(axis cs:0.249096434863135,0.113550746616246)
--(axis cs:0.248397435069818,0.150688792308648)
--(axis cs:0.247503555887047,0.187220067419092)
--(axis cs:0.246418396650804,0.222997473396197)
--(axis cs:0.245146326914331,0.257876947253503)
--(axis cs:0.243692468853489,0.291718041660281)
--(axis cs:0.242062676641559,0.324384490473359)
--(axis cs:0.240263512876553,0.355744757432784)
--(axis cs:0.238302222155949,0.385672565811924)
--(axis cs:0.236186701905254,0.414047406889267)
--(axis cs:0.233925470577857,0.44075502519452)
--(axis cs:0.231527633354226,0.465687878575054)
--(axis cs:0.22900284547856,0.488745571230201)
--(axis cs:0.226361273380525,0.509835257969709)
--(axis cs:0.223613553738634,0.528872018068549)
--(axis cs:0.220770750650094,0.545779197212711)
--(axis cs:0.217844311079594,0.560488716159064)
--(axis cs:0.214846018766414,0.572941344866444)
--(axis cs:0.211787946775471,0.583086940994125)
--(axis cs:0.208682408883347,0.590884651807325)
--(axis cs:0.205541909995051,0.596303078676752)
--(axis cs:0.202379095791187,0.599320403509805)
--(axis cs:0.19920670180826,0.599924476604325)
--(axis cs:0.196037502157161,0.598112865571165)
--(axis cs:0.192884258086336,0.59389286512856)
--(axis cs:0.18975966659674,0.587281467728867)
--(axis cs:0.186676309315498,0.578305295135965)
--(axis cs:0.183646601834129,0.567000491228801)
--(axis cs:0.180682743715344,0.553412576462749)
--(axis cs:0.177796669369711,0.537596264574801)
--(axis cs:0.175,0.519615242270663)
--(axis cs:0.172303996806695,0.499541912780863)
--(axis cs:0.169719515643117,0.477457104318499)
--(axis cs:0.167256963302736,0.453449744612555)
--(axis cs:0.164926255614684,0.427616502827318)
--(axis cs:0.162736777516212,0.400061400309775)
--(axis cs:0.160697345262861,0.370895391732363)
--(axis cs:0.158816170928508,0.340235918317662)
--(axis cs:0.157100829338251,0.308206434944044)
--(axis cs:0.155558227567254,0.274935913036446)
--(axis cs:0.154194577128397,0.240558321243968)
--(axis cs:0.153015368960705,0.205212085995401)
--(axis cs:0.152025351319275,0.169039534104858)
--(axis cs:0.15122851065573,0.132186319671924)
--(axis cs:0.15062805556618,0.0948008375840099)
--(axis cs:0.150226403871346,0.0570336259825095)
--(axis cs:0.150025172880841,0.0190367600988406)
--(axis cs:0.150025172880841,-0.0190367600988407)
--(axis cs:0.150226403871346,-0.0570336259825096)
--(axis cs:0.15062805556618,-0.09480083758401)
--(axis cs:0.15122851065573,-0.132186319671924)
--(axis cs:0.152025351319275,-0.169039534104858)
--(axis cs:0.153015368960705,-0.205212085995401)
--(axis cs:0.154194577128397,-0.240558321243968)
--(axis cs:0.155558227567254,-0.274935913036446)
--(axis cs:0.157100829338251,-0.308206434944044)
--(axis cs:0.158816170928508,-0.340235918317662)
--(axis cs:0.160697345262861,-0.370895391732363)
--(axis cs:0.162736777516212,-0.400061400309775)
--(axis cs:0.164926255614684,-0.427616502827318)
--(axis cs:0.167256963302736,-0.453449744612555)
--(axis cs:0.169719515643117,-0.477457104318499)
--(axis cs:0.172303996806694,-0.499541912780863)
--(axis cs:0.175,-0.519615242270663)
--(axis cs:0.177796669369711,-0.537596264574802)
--(axis cs:0.180682743715344,-0.553412576462749)
--(axis cs:0.183646601834129,-0.567000491228801)
--(axis cs:0.186676309315498,-0.578305295135965)
--(axis cs:0.18975966659674,-0.587281467728867)
--(axis cs:0.192884258086336,-0.59389286512856)
--(axis cs:0.196037502157161,-0.598112865571165)
--(axis cs:0.19920670180826,-0.599924476604325)
--(axis cs:0.202379095791187,-0.599320403509805)
--(axis cs:0.205541909995051,-0.596303078676752)
--(axis cs:0.208682408883347,-0.590884651807325)
--(axis cs:0.211787946775471,-0.583086940994125)
--(axis cs:0.214846018766414,-0.572941344866444)
--(axis cs:0.217844311079594,-0.560488716159064)
--(axis cs:0.220770750650094,-0.545779197212711)
--(axis cs:0.223613553738634,-0.528872018068549)
--(axis cs:0.226361273380525,-0.509835257969709)
--(axis cs:0.22900284547856,-0.488745571230201)
--(axis cs:0.231527633354226,-0.465687878575054)
--(axis cs:0.233925470577857,-0.44075502519452)
--(axis cs:0.236186701905254,-0.414047406889267)
--(axis cs:0.238302222155949,-0.385672565811924)
--(axis cs:0.240263512876553,-0.355744757432784)
--(axis cs:0.242062676641559,-0.324384490473358)
--(axis cs:0.243692468853489,-0.291718041660281)
--(axis cs:0.245146326914331,-0.257876947253503)
--(axis cs:0.246418396650804,-0.222997473396196)
--(axis cs:0.247503555887047,-0.187220067419092)
--(axis cs:0.248397435069818,-0.150688792308648)
--(axis cs:0.249096434863135,-0.113550746616246)
--(axis cs:0.24959774064154,-0.0759554721442494)
--(axis cs:0.249899333823594,-0.0380543517939387)
--(axis cs:0.25,-1.46957615897682e-16)
--cycle;
\path [draw=sandybrown24416662, fill=sandybrown24416662, opacity=0.7]
(axis cs:0.55,0.6)
--(axis cs:0.549899333823594,0.619027175896969)
--(axis cs:0.54959774064154,0.637977736072125)
--(axis cs:0.549096434863135,0.656775373308123)
--(axis cs:0.548397435069818,0.675344396154324)
--(axis cs:0.547503555887047,0.693610033709546)
--(axis cs:0.546418396650804,0.711498736698098)
--(axis cs:0.545146326914331,0.728938473626751)
--(axis cs:0.543692468853489,0.745859020830141)
--(axis cs:0.542062676641559,0.762192245236679)
--(axis cs:0.540263512876553,0.777872378716392)
--(axis cs:0.538302222155949,0.792836282905962)
--(axis cs:0.536186701905254,0.807023703444634)
--(axis cs:0.533925470577857,0.82037751259726)
--(axis cs:0.531527633354226,0.832843939287527)
--(axis cs:0.52900284547856,0.844372785615101)
--(axis cs:0.526361273380525,0.854917628984854)
--(axis cs:0.523613553738634,0.864436009034275)
--(axis cs:0.520770750650094,0.872889598606355)
--(axis cs:0.517844311079594,0.880244358079532)
--(axis cs:0.514846018766414,0.886470672433222)
--(axis cs:0.511787946775471,0.891543470497062)
--(axis cs:0.508682408883346,0.895442325903662)
--(axis cs:0.505541909995051,0.898151539338376)
--(axis cs:0.502379095791187,0.899660201754902)
--(axis cs:0.49920670180826,0.899962238302162)
--(axis cs:0.496037502157161,0.899056432785583)
--(axis cs:0.492884258086336,0.89694643256428)
--(axis cs:0.48975966659674,0.893640733864434)
--(axis cs:0.486676309315498,0.889152647567983)
--(axis cs:0.483646601834129,0.883500245614401)
--(axis cs:0.480682743715344,0.876706288231374)
--(axis cs:0.477796669369711,0.868798132287401)
--(axis cs:0.475,0.859807621135331)
--(axis cs:0.472303996806694,0.849770956390431)
--(axis cs:0.469719515643117,0.83872855215925)
--(axis cs:0.467256963302736,0.826724872306277)
--(axis cs:0.464926255614684,0.813808251413659)
--(axis cs:0.462736777516212,0.800030700154888)
--(axis cs:0.460697345262861,0.785447695866181)
--(axis cs:0.458816170928508,0.770117959158831)
--(axis cs:0.457100829338251,0.754103217472022)
--(axis cs:0.455558227567254,0.737467956518223)
--(axis cs:0.454194577128397,0.720279160621984)
--(axis cs:0.453015368960705,0.702606042997701)
--(axis cs:0.452025351319275,0.684519767052429)
--(axis cs:0.45122851065573,0.666093159835962)
--(axis cs:0.45062805556618,0.647400418792005)
--(axis cs:0.450226403871346,0.628516812991255)
--(axis cs:0.450025172880841,0.60951838004942)
--(axis cs:0.450025172880841,0.59048161995058)
--(axis cs:0.450226403871346,0.571483187008745)
--(axis cs:0.45062805556618,0.552599581207995)
--(axis cs:0.45122851065573,0.533906840164038)
--(axis cs:0.452025351319275,0.515480232947571)
--(axis cs:0.453015368960705,0.497393957002299)
--(axis cs:0.454194577128397,0.479720839378016)
--(axis cs:0.455558227567254,0.462532043481777)
--(axis cs:0.457100829338251,0.445896782527978)
--(axis cs:0.458816170928508,0.429882040841169)
--(axis cs:0.460697345262861,0.414552304133818)
--(axis cs:0.462736777516212,0.399969299845113)
--(axis cs:0.464926255614684,0.386191748586341)
--(axis cs:0.467256963302736,0.373275127693723)
--(axis cs:0.469719515643117,0.36127144784075)
--(axis cs:0.472303996806694,0.350229043609569)
--(axis cs:0.475,0.340192378864668)
--(axis cs:0.477796669369711,0.331201867712599)
--(axis cs:0.480682743715344,0.323293711768626)
--(axis cs:0.483646601834129,0.316499754385599)
--(axis cs:0.486676309315498,0.310847352432017)
--(axis cs:0.48975966659674,0.306359266135566)
--(axis cs:0.492884258086336,0.30305356743572)
--(axis cs:0.496037502157161,0.300943567214417)
--(axis cs:0.49920670180826,0.300037761697837)
--(axis cs:0.502379095791187,0.300339798245098)
--(axis cs:0.505541909995051,0.301848460661624)
--(axis cs:0.508682408883346,0.304557674096338)
--(axis cs:0.511787946775471,0.308456529502938)
--(axis cs:0.514846018766414,0.313529327566778)
--(axis cs:0.517844311079594,0.319755641920468)
--(axis cs:0.520770750650094,0.327110401393645)
--(axis cs:0.523613553738634,0.335563990965725)
--(axis cs:0.526361273380525,0.345082371015146)
--(axis cs:0.52900284547856,0.355627214384899)
--(axis cs:0.531527633354226,0.367156060712473)
--(axis cs:0.533925470577857,0.37962248740274)
--(axis cs:0.536186701905254,0.392976296555366)
--(axis cs:0.538302222155949,0.407163717094038)
--(axis cs:0.540263512876553,0.422127621283608)
--(axis cs:0.542062676641559,0.437807754763321)
--(axis cs:0.543692468853489,0.454140979169859)
--(axis cs:0.545146326914331,0.471061526373248)
--(axis cs:0.546418396650804,0.488501263301902)
--(axis cs:0.547503555887047,0.506389966290454)
--(axis cs:0.548397435069818,0.524655603845676)
--(axis cs:0.549096434863135,0.543224626691877)
--(axis cs:0.54959774064154,0.562022263927875)
--(axis cs:0.549899333823594,0.580972824103031)
--(axis cs:0.55,0.6)
--cycle;
\path [draw=sandybrown24416662, fill=sandybrown24416662, opacity=0.7]
(axis cs:0.9,0.25)
--(axis cs:0.899899333823594,0.288054351793939)
--(axis cs:0.89959774064154,0.32595547214425)
--(axis cs:0.899096434863135,0.363550746616246)
--(axis cs:0.898397435069818,0.400688792308648)
--(axis cs:0.897503555887047,0.437220067419092)
--(axis cs:0.896418396650804,0.472997473396197)
--(axis cs:0.895146326914331,0.507876947253503)
--(axis cs:0.893692468853489,0.541718041660281)
--(axis cs:0.892062676641559,0.574384490473359)
--(axis cs:0.890263512876553,0.605744757432784)
--(axis cs:0.888302222155949,0.635672565811924)
--(axis cs:0.886186701905254,0.664047406889267)
--(axis cs:0.883925470577857,0.69075502519452)
--(axis cs:0.881527633354226,0.715687878575054)
--(axis cs:0.87900284547856,0.738745571230201)
--(axis cs:0.876361273380525,0.759835257969709)
--(axis cs:0.873613553738634,0.778872018068549)
--(axis cs:0.870770750650094,0.795779197212711)
--(axis cs:0.867844311079594,0.810488716159064)
--(axis cs:0.864846018766414,0.822941344866444)
--(axis cs:0.861787946775471,0.833086940994125)
--(axis cs:0.858682408883346,0.840884651807325)
--(axis cs:0.855541909995051,0.846303078676752)
--(axis cs:0.852379095791187,0.849320403509805)
--(axis cs:0.84920670180826,0.849924476604325)
--(axis cs:0.846037502157161,0.848112865571165)
--(axis cs:0.842884258086336,0.84389286512856)
--(axis cs:0.83975966659674,0.837281467728867)
--(axis cs:0.836676309315498,0.828305295135965)
--(axis cs:0.833646601834129,0.817000491228801)
--(axis cs:0.830682743715344,0.803412576462749)
--(axis cs:0.827796669369711,0.787596264574801)
--(axis cs:0.825,0.769615242270663)
--(axis cs:0.822303996806694,0.749541912780863)
--(axis cs:0.819719515643117,0.727457104318499)
--(axis cs:0.817256963302736,0.703449744612555)
--(axis cs:0.814926255614684,0.677616502827318)
--(axis cs:0.812736777516212,0.650061400309775)
--(axis cs:0.810697345262861,0.620895391732363)
--(axis cs:0.808816170928508,0.590235918317662)
--(axis cs:0.807100829338251,0.558206434944044)
--(axis cs:0.805558227567254,0.524935913036446)
--(axis cs:0.804194577128396,0.490558321243968)
--(axis cs:0.803015368960705,0.455212085995401)
--(axis cs:0.802025351319275,0.419039534104858)
--(axis cs:0.80122851065573,0.382186319671925)
--(axis cs:0.80062805556618,0.34480083758401)
--(axis cs:0.800226403871346,0.307033625982509)
--(axis cs:0.800025172880841,0.269036760098841)
--(axis cs:0.800025172880841,0.230963239901159)
--(axis cs:0.800226403871346,0.19296637401749)
--(axis cs:0.80062805556618,0.15519916241599)
--(axis cs:0.80122851065573,0.117813680328076)
--(axis cs:0.802025351319275,0.0809604658951421)
--(axis cs:0.803015368960705,0.0447879140045988)
--(axis cs:0.804194577128396,0.00944167875603175)
--(axis cs:0.805558227567254,-0.0249359130364462)
--(axis cs:0.807100829338251,-0.0582064349440438)
--(axis cs:0.808816170928508,-0.0902359183176624)
--(axis cs:0.810697345262861,-0.120895391732363)
--(axis cs:0.812736777516212,-0.150061400309775)
--(axis cs:0.814926255614684,-0.177616502827318)
--(axis cs:0.817256963302736,-0.203449744612555)
--(axis cs:0.819719515643117,-0.227457104318499)
--(axis cs:0.822303996806694,-0.249541912780863)
--(axis cs:0.825,-0.269615242270663)
--(axis cs:0.827796669369711,-0.287596264574802)
--(axis cs:0.830682743715344,-0.303412576462749)
--(axis cs:0.833646601834129,-0.317000491228801)
--(axis cs:0.836676309315498,-0.328305295135965)
--(axis cs:0.83975966659674,-0.337281467728867)
--(axis cs:0.842884258086336,-0.34389286512856)
--(axis cs:0.846037502157161,-0.348112865571165)
--(axis cs:0.84920670180826,-0.349924476604325)
--(axis cs:0.852379095791187,-0.349320403509805)
--(axis cs:0.855541909995051,-0.346303078676752)
--(axis cs:0.858682408883346,-0.340884651807325)
--(axis cs:0.861787946775471,-0.333086940994125)
--(axis cs:0.864846018766414,-0.322941344866444)
--(axis cs:0.867844311079594,-0.310488716159064)
--(axis cs:0.870770750650094,-0.295779197212711)
--(axis cs:0.873613553738634,-0.278872018068549)
--(axis cs:0.876361273380525,-0.259835257969709)
--(axis cs:0.87900284547856,-0.238745571230201)
--(axis cs:0.881527633354226,-0.215687878575054)
--(axis cs:0.883925470577857,-0.19075502519452)
--(axis cs:0.886186701905254,-0.164047406889267)
--(axis cs:0.888302222155949,-0.135672565811924)
--(axis cs:0.890263512876553,-0.105744757432784)
--(axis cs:0.892062676641559,-0.0743844904733584)
--(axis cs:0.893692468853489,-0.0417180416602813)
--(axis cs:0.895146326914331,-0.00787694725350313)
--(axis cs:0.896418396650804,0.0270025266038037)
--(axis cs:0.897503555887047,0.0627799325809078)
--(axis cs:0.898397435069818,0.0993112076913524)
--(axis cs:0.899096434863135,0.136449253383754)
--(axis cs:0.89959774064154,0.174044527855751)
--(axis cs:0.899899333823594,0.211945648206061)
--(axis cs:0.9,0.25)
--cycle;
\addplot [semithick, seagreen4916384]
table {%
0 0
0.0415696381644869 0.00451078787595934
0.10751609652625 0.00160698758124196
0.14017673351052 0.000421583852257978
0.146476057046068 0.0838763730654501
0.148032978294555 0.142323854805112
0.144628471715774 0.202704994349931
0.135745001108529 0.249223598683357
0.13049379861698 0.276933226517324
0.147855508566828 0.317757212403772
0.146313370633972 0.37080837745746
0.158520884255071 0.405374836862201
0.159429729903208 0.460425304554041
0.154945002648916 0.523273514847922
0.1662709469262 0.561843611950958
0.182300669572691 0.589555848599087
0.203619353100965 0.600553052046533
0.2133457518594 0.590526834281456
0.218812469865081 0.565663270318327
0.238021937735015 0.525283287862512
0.23702030206668 0.488722942794848
0.241744471816555 0.431572861179458
0.254651541046017 0.376818791147109
0.244358136430242 0.330599675346923
0.256849630268135 0.26974238967654
0.258533862361876 0.213917906441688
0.269664247237174 0.151418791521504
0.267730628890688 0.104465257982925
0.297427602967777 0.0684613962475035
0.292781723942878 0.025258041757871
0.329008178560421 0.00446819681778509
0.37576022400058 0.00337357357896381
0.401872828020156 0.0064439565260806
0.414809840119547 0.00054521478286407
0.432546201105769 0.00109471823266228
0.480629591765708 0.0088965948047152
0.513142603919598 0.000397445105582779
0.539391112374575 0.0014066463998513
0.560233807455654 0.00763946287797793
0.579792595131749 0.00464428996029921
0.606405765113705 0.00487151899960593
0.621399333039233 0.00363842415386143
0.652174972000437 0.000690479361215648
0.710821720587591 0.0104222887844351
0.731255635379985 0.00110362051512484
0.750653101770275 0.000474580317309978
0.764492254876648 0.000268615529553541
0.795514063507726 0.00174667419224552
0.798205190848392 0.0347107411819475
0.786740144888324 0.104183508324982
0.790301887586875 0.135192494473712
0.786734414110438 0.156558724069501
0.78423888915872 0.202395236860114
0.796944209243069 0.229245979518855
0.791617635751641 0.299601181640891
0.792679447898817 0.349248479877286
0.798991412667341 0.419488373367411
0.802119929629888 0.482588543463718
0.797004148689488 0.51241365954229
0.803117059022193 0.538515787779778
0.804587384508714 0.580741120879927
0.80885112766389 0.619588254358192
0.810291147044502 0.657919726450113
0.786702083408471 0.718321069747742
0.805443024537409 0.760477415795059
0.817310012248351 0.797758081505238
0.818684321341695 0.841439733671602
0.848330677988143 0.870382574303793
0.86810801472403 0.828263279905625
0.884058682465209 0.796382087301896
0.889144960207982 0.753859766275873
0.897550966176097 0.717229189544371
0.886260059378327 0.681999600484523
0.89401259639922 0.615674007777054
0.894569654882713 0.561359727013777
0.913017521611945 0.509179161241898
0.916493463514091 0.461180231734816
0.9115558070429 0.420502780231134
0.910320084013942 0.374745725592601
0.915404923740473 0.330177069939572
0.930823466492532 0.28604283706084
0.906382112002761 0.261292796295379
0.899824745618471 0.199251255475081
0.918149484566555 0.14364332236741
0.920154594181534 0.0919272851375059
0.902192821475319 0.0537859937057365
0.936364004220845 0.00809263375055899
1 0
};
\addplot [semithick, sandybrown24416662, opacity=0.7]
table {%
0.25 0
0.249899333823594 0.0380543517939387
0.24959774064154 0.0759554721442496
0.249096434863135 0.113550746616246
0.248397435069818 0.150688792308648
0.247503555887047 0.187220067419092
0.246418396650804 0.222997473396197
0.245146326914331 0.257876947253503
0.243692468853489 0.291718041660281
0.242062676641559 0.324384490473359
0.240263512876553 0.355744757432784
0.238302222155949 0.385672565811924
0.236186701905254 0.414047406889267
0.233925470577857 0.44075502519452
0.231527633354226 0.465687878575054
0.22900284547856 0.488745571230201
0.226361273380525 0.509835257969709
0.223613553738634 0.528872018068549
0.220770750650094 0.545779197212711
0.217844311079594 0.560488716159064
0.214846018766414 0.572941344866444
0.211787946775471 0.583086940994125
0.208682408883347 0.590884651807325
0.205541909995051 0.596303078676752
0.202379095791187 0.599320403509805
0.19920670180826 0.599924476604325
0.196037502157161 0.598112865571165
0.192884258086336 0.59389286512856
0.18975966659674 0.587281467728867
0.186676309315498 0.578305295135965
0.183646601834129 0.567000491228801
0.180682743715344 0.553412576462749
0.177796669369711 0.537596264574801
0.175 0.519615242270663
0.172303996806695 0.499541912780863
0.169719515643117 0.477457104318499
0.167256963302736 0.453449744612555
0.164926255614684 0.427616502827318
0.162736777516212 0.400061400309775
0.160697345262861 0.370895391732363
0.158816170928508 0.340235918317662
0.157100829338251 0.308206434944044
0.155558227567254 0.274935913036446
0.154194577128397 0.240558321243968
0.153015368960705 0.205212085995401
0.152025351319275 0.169039534104858
0.15122851065573 0.132186319671924
0.15062805556618 0.0948008375840099
0.150226403871346 0.0570336259825095
0.150025172880841 0.0190367600988406
0.150025172880841 -0.0190367600988407
0.150226403871346 -0.0570336259825096
0.15062805556618 -0.09480083758401
0.15122851065573 -0.132186319671924
0.152025351319275 -0.169039534104858
0.153015368960705 -0.205212085995401
0.154194577128397 -0.240558321243968
0.155558227567254 -0.274935913036446
0.157100829338251 -0.308206434944044
0.158816170928508 -0.340235918317662
0.160697345262861 -0.370895391732363
0.162736777516212 -0.400061400309775
0.164926255614684 -0.427616502827318
0.167256963302736 -0.453449744612555
0.169719515643117 -0.477457104318499
0.172303996806694 -0.499541912780863
0.175 -0.519615242270663
0.177796669369711 -0.537596264574802
0.180682743715344 -0.553412576462749
0.183646601834129 -0.567000491228801
0.186676309315498 -0.578305295135965
0.18975966659674 -0.587281467728867
0.192884258086336 -0.59389286512856
0.196037502157161 -0.598112865571165
0.19920670180826 -0.599924476604325
0.202379095791187 -0.599320403509805
0.205541909995051 -0.596303078676752
0.208682408883347 -0.590884651807325
0.211787946775471 -0.583086940994125
0.214846018766414 -0.572941344866444
0.217844311079594 -0.560488716159064
0.220770750650094 -0.545779197212711
0.223613553738634 -0.528872018068549
0.226361273380525 -0.509835257969709
0.22900284547856 -0.488745571230201
0.231527633354226 -0.465687878575054
0.233925470577857 -0.44075502519452
0.236186701905254 -0.414047406889267
0.238302222155949 -0.385672565811924
0.240263512876553 -0.355744757432784
0.242062676641559 -0.324384490473358
0.243692468853489 -0.291718041660281
0.245146326914331 -0.257876947253503
0.246418396650804 -0.222997473396196
0.247503555887047 -0.187220067419092
0.248397435069818 -0.150688792308648
0.249096434863135 -0.113550746616246
0.24959774064154 -0.0759554721442494
0.249899333823594 -0.0380543517939387
0.25 -1.46957615897682e-16
};
\addplot [semithick, sandybrown24416662, opacity=0.7]
table {%
0.55 0.6
0.549899333823594 0.619027175896969
0.54959774064154 0.637977736072125
0.549096434863135 0.656775373308123
0.548397435069818 0.675344396154324
0.547503555887047 0.693610033709546
0.546418396650804 0.711498736698098
0.545146326914331 0.728938473626751
0.543692468853489 0.745859020830141
0.542062676641559 0.762192245236679
0.540263512876553 0.777872378716392
0.538302222155949 0.792836282905962
0.536186701905254 0.807023703444634
0.533925470577857 0.82037751259726
0.531527633354226 0.832843939287527
0.52900284547856 0.844372785615101
0.526361273380525 0.854917628984854
0.523613553738634 0.864436009034275
0.520770750650094 0.872889598606355
0.517844311079594 0.880244358079532
0.514846018766414 0.886470672433222
0.511787946775471 0.891543470497062
0.508682408883346 0.895442325903662
0.505541909995051 0.898151539338376
0.502379095791187 0.899660201754902
0.49920670180826 0.899962238302162
0.496037502157161 0.899056432785583
0.492884258086336 0.89694643256428
0.48975966659674 0.893640733864434
0.486676309315498 0.889152647567983
0.483646601834129 0.883500245614401
0.480682743715344 0.876706288231374
0.477796669369711 0.868798132287401
0.475 0.859807621135331
0.472303996806694 0.849770956390431
0.469719515643117 0.83872855215925
0.467256963302736 0.826724872306277
0.464926255614684 0.813808251413659
0.462736777516212 0.800030700154888
0.460697345262861 0.785447695866181
0.458816170928508 0.770117959158831
0.457100829338251 0.754103217472022
0.455558227567254 0.737467956518223
0.454194577128397 0.720279160621984
0.453015368960705 0.702606042997701
0.452025351319275 0.684519767052429
0.45122851065573 0.666093159835962
0.45062805556618 0.647400418792005
0.450226403871346 0.628516812991255
0.450025172880841 0.60951838004942
0.450025172880841 0.59048161995058
0.450226403871346 0.571483187008745
0.45062805556618 0.552599581207995
0.45122851065573 0.533906840164038
0.452025351319275 0.515480232947571
0.453015368960705 0.497393957002299
0.454194577128397 0.479720839378016
0.455558227567254 0.462532043481777
0.457100829338251 0.445896782527978
0.458816170928508 0.429882040841169
0.460697345262861 0.414552304133818
0.462736777516212 0.399969299845113
0.464926255614684 0.386191748586341
0.467256963302736 0.373275127693723
0.469719515643117 0.36127144784075
0.472303996806694 0.350229043609569
0.475 0.340192378864668
0.477796669369711 0.331201867712599
0.480682743715344 0.323293711768626
0.483646601834129 0.316499754385599
0.486676309315498 0.310847352432017
0.48975966659674 0.306359266135566
0.492884258086336 0.30305356743572
0.496037502157161 0.300943567214417
0.49920670180826 0.300037761697837
0.502379095791187 0.300339798245098
0.505541909995051 0.301848460661624
0.508682408883346 0.304557674096338
0.511787946775471 0.308456529502938
0.514846018766414 0.313529327566778
0.517844311079594 0.319755641920468
0.520770750650094 0.327110401393645
0.523613553738634 0.335563990965725
0.526361273380525 0.345082371015146
0.52900284547856 0.355627214384899
0.531527633354226 0.367156060712473
0.533925470577857 0.37962248740274
0.536186701905254 0.392976296555366
0.538302222155949 0.407163717094038
0.540263512876553 0.422127621283608
0.542062676641559 0.437807754763321
0.543692468853489 0.454140979169859
0.545146326914331 0.471061526373248
0.546418396650804 0.488501263301902
0.547503555887047 0.506389966290454
0.548397435069818 0.524655603845676
0.549096434863135 0.543224626691877
0.54959774064154 0.562022263927875
0.549899333823594 0.580972824103031
0.55 0.6
};
\addplot [semithick, sandybrown24416662, opacity=0.7]
table {%
0.9 0.25
0.899899333823594 0.288054351793939
0.89959774064154 0.32595547214425
0.899096434863135 0.363550746616246
0.898397435069818 0.400688792308648
0.897503555887047 0.437220067419092
0.896418396650804 0.472997473396197
0.895146326914331 0.507876947253503
0.893692468853489 0.541718041660281
0.892062676641559 0.574384490473359
0.890263512876553 0.605744757432784
0.888302222155949 0.635672565811924
0.886186701905254 0.664047406889267
0.883925470577857 0.69075502519452
0.881527633354226 0.715687878575054
0.87900284547856 0.738745571230201
0.876361273380525 0.759835257969709
0.873613553738634 0.778872018068549
0.870770750650094 0.795779197212711
0.867844311079594 0.810488716159064
0.864846018766414 0.822941344866444
0.861787946775471 0.833086940994125
0.858682408883346 0.840884651807325
0.855541909995051 0.846303078676752
0.852379095791187 0.849320403509805
0.84920670180826 0.849924476604325
0.846037502157161 0.848112865571165
0.842884258086336 0.84389286512856
0.83975966659674 0.837281467728867
0.836676309315498 0.828305295135965
0.833646601834129 0.817000491228801
0.830682743715344 0.803412576462749
0.827796669369711 0.787596264574801
0.825 0.769615242270663
0.822303996806694 0.749541912780863
0.819719515643117 0.727457104318499
0.817256963302736 0.703449744612555
0.814926255614684 0.677616502827318
0.812736777516212 0.650061400309775
0.810697345262861 0.620895391732363
0.808816170928508 0.590235918317662
0.807100829338251 0.558206434944044
0.805558227567254 0.524935913036446
0.804194577128396 0.490558321243968
0.803015368960705 0.455212085995401
0.802025351319275 0.419039534104858
0.80122851065573 0.382186319671925
0.80062805556618 0.34480083758401
0.800226403871346 0.307033625982509
0.800025172880841 0.269036760098841
0.800025172880841 0.230963239901159
0.800226403871346 0.19296637401749
0.80062805556618 0.15519916241599
0.80122851065573 0.117813680328076
0.802025351319275 0.0809604658951421
0.803015368960705 0.0447879140045988
0.804194577128396 0.00944167875603175
0.805558227567254 -0.0249359130364462
0.807100829338251 -0.0582064349440438
0.808816170928508 -0.0902359183176624
0.810697345262861 -0.120895391732363
0.812736777516212 -0.150061400309775
0.814926255614684 -0.177616502827318
0.817256963302736 -0.203449744612555
0.819719515643117 -0.227457104318499
0.822303996806694 -0.249541912780863
0.825 -0.269615242270663
0.827796669369711 -0.287596264574802
0.830682743715344 -0.303412576462749
0.833646601834129 -0.317000491228801
0.836676309315498 -0.328305295135965
0.83975966659674 -0.337281467728867
0.842884258086336 -0.34389286512856
0.846037502157161 -0.348112865571165
0.84920670180826 -0.349924476604325
0.852379095791187 -0.349320403509805
0.855541909995051 -0.346303078676752
0.858682408883346 -0.340884651807325
0.861787946775471 -0.333086940994125
0.864846018766414 -0.322941344866444
0.867844311079594 -0.310488716159064
0.870770750650094 -0.295779197212711
0.873613553738634 -0.278872018068549
0.876361273380525 -0.259835257969709
0.87900284547856 -0.238745571230201
0.881527633354226 -0.215687878575054
0.883925470577857 -0.19075502519452
0.886186701905254 -0.164047406889267
0.888302222155949 -0.135672565811924
0.890263512876553 -0.105744757432784
0.892062676641559 -0.0743844904733584
0.893692468853489 -0.0417180416602813
0.895146326914331 -0.00787694725350313
0.896418396650804 0.0270025266038037
0.897503555887047 0.0627799325809078
0.898397435069818 0.0993112076913524
0.899096434863135 0.136449253383754
0.89959774064154 0.174044527855751
0.899899333823594 0.211945648206061
0.9 0.25
};
\draw (axis cs:0.02,0.98) node[
  scale=0.85,
  anchor=north west,
  text=black,
  rotate=0.0
]{$\gamma=$0.0};
\draw (axis cs:0.02,0.895) node[
  scale=0.85,
  anchor=north west,
  text=black,
  rotate=0.0
]{$\overline{C}(\sigma^*)=1.02$};
\end{axis}

\end{tikzpicture}

%% file: icra2024/tex/appendix/varying-gammas/gamma-1.0.tex
\begin{tikzpicture}

\definecolor{sandybrown24416662}{RGB}{244,166,62}
\definecolor{seagreen4916384}{RGB}{49,163,84}

\begin{axis}[
width=0.33\linewidth, 
height=0.33\linewidth, 
tick pos=left,
xlabel={x},
xmin=0, xmax=1,
xtick style={color=black},
ylabel={y},
ymin=0, ymax=1,
ytick style={color=black},
]
\path [draw=sandybrown24416662, fill=sandybrown24416662, opacity=0.7]
(axis cs:0.25,0)
--(axis cs:0.249899333823594,0.0380543517939387)
--(axis cs:0.24959774064154,0.0759554721442496)
--(axis cs:0.249096434863135,0.113550746616246)
--(axis cs:0.248397435069818,0.150688792308648)
--(axis cs:0.247503555887047,0.187220067419092)
--(axis cs:0.246418396650804,0.222997473396197)
--(axis cs:0.245146326914331,0.257876947253503)
--(axis cs:0.243692468853489,0.291718041660281)
--(axis cs:0.242062676641559,0.324384490473359)
--(axis cs:0.240263512876553,0.355744757432784)
--(axis cs:0.238302222155949,0.385672565811924)
--(axis cs:0.236186701905254,0.414047406889267)
--(axis cs:0.233925470577857,0.44075502519452)
--(axis cs:0.231527633354226,0.465687878575054)
--(axis cs:0.22900284547856,0.488745571230201)
--(axis cs:0.226361273380525,0.509835257969709)
--(axis cs:0.223613553738634,0.528872018068549)
--(axis cs:0.220770750650094,0.545779197212711)
--(axis cs:0.217844311079594,0.560488716159064)
--(axis cs:0.214846018766414,0.572941344866444)
--(axis cs:0.211787946775471,0.583086940994125)
--(axis cs:0.208682408883347,0.590884651807325)
--(axis cs:0.205541909995051,0.596303078676752)
--(axis cs:0.202379095791187,0.599320403509805)
--(axis cs:0.19920670180826,0.599924476604325)
--(axis cs:0.196037502157161,0.598112865571165)
--(axis cs:0.192884258086336,0.59389286512856)
--(axis cs:0.18975966659674,0.587281467728867)
--(axis cs:0.186676309315498,0.578305295135965)
--(axis cs:0.183646601834129,0.567000491228801)
--(axis cs:0.180682743715344,0.553412576462749)
--(axis cs:0.177796669369711,0.537596264574801)
--(axis cs:0.175,0.519615242270663)
--(axis cs:0.172303996806695,0.499541912780863)
--(axis cs:0.169719515643117,0.477457104318499)
--(axis cs:0.167256963302736,0.453449744612555)
--(axis cs:0.164926255614684,0.427616502827318)
--(axis cs:0.162736777516212,0.400061400309775)
--(axis cs:0.160697345262861,0.370895391732363)
--(axis cs:0.158816170928508,0.340235918317662)
--(axis cs:0.157100829338251,0.308206434944044)
--(axis cs:0.155558227567254,0.274935913036446)
--(axis cs:0.154194577128397,0.240558321243968)
--(axis cs:0.153015368960705,0.205212085995401)
--(axis cs:0.152025351319275,0.169039534104858)
--(axis cs:0.15122851065573,0.132186319671924)
--(axis cs:0.15062805556618,0.0948008375840099)
--(axis cs:0.150226403871346,0.0570336259825095)
--(axis cs:0.150025172880841,0.0190367600988406)
--(axis cs:0.150025172880841,-0.0190367600988407)
--(axis cs:0.150226403871346,-0.0570336259825096)
--(axis cs:0.15062805556618,-0.09480083758401)
--(axis cs:0.15122851065573,-0.132186319671924)
--(axis cs:0.152025351319275,-0.169039534104858)
--(axis cs:0.153015368960705,-0.205212085995401)
--(axis cs:0.154194577128397,-0.240558321243968)
--(axis cs:0.155558227567254,-0.274935913036446)
--(axis cs:0.157100829338251,-0.308206434944044)
--(axis cs:0.158816170928508,-0.340235918317662)
--(axis cs:0.160697345262861,-0.370895391732363)
--(axis cs:0.162736777516212,-0.400061400309775)
--(axis cs:0.164926255614684,-0.427616502827318)
--(axis cs:0.167256963302736,-0.453449744612555)
--(axis cs:0.169719515643117,-0.477457104318499)
--(axis cs:0.172303996806694,-0.499541912780863)
--(axis cs:0.175,-0.519615242270663)
--(axis cs:0.177796669369711,-0.537596264574802)
--(axis cs:0.180682743715344,-0.553412576462749)
--(axis cs:0.183646601834129,-0.567000491228801)
--(axis cs:0.186676309315498,-0.578305295135965)
--(axis cs:0.18975966659674,-0.587281467728867)
--(axis cs:0.192884258086336,-0.59389286512856)
--(axis cs:0.196037502157161,-0.598112865571165)
--(axis cs:0.19920670180826,-0.599924476604325)
--(axis cs:0.202379095791187,-0.599320403509805)
--(axis cs:0.205541909995051,-0.596303078676752)
--(axis cs:0.208682408883347,-0.590884651807325)
--(axis cs:0.211787946775471,-0.583086940994125)
--(axis cs:0.214846018766414,-0.572941344866444)
--(axis cs:0.217844311079594,-0.560488716159064)
--(axis cs:0.220770750650094,-0.545779197212711)
--(axis cs:0.223613553738634,-0.528872018068549)
--(axis cs:0.226361273380525,-0.509835257969709)
--(axis cs:0.22900284547856,-0.488745571230201)
--(axis cs:0.231527633354226,-0.465687878575054)
--(axis cs:0.233925470577857,-0.44075502519452)
--(axis cs:0.236186701905254,-0.414047406889267)
--(axis cs:0.238302222155949,-0.385672565811924)
--(axis cs:0.240263512876553,-0.355744757432784)
--(axis cs:0.242062676641559,-0.324384490473358)
--(axis cs:0.243692468853489,-0.291718041660281)
--(axis cs:0.245146326914331,-0.257876947253503)
--(axis cs:0.246418396650804,-0.222997473396196)
--(axis cs:0.247503555887047,-0.187220067419092)
--(axis cs:0.248397435069818,-0.150688792308648)
--(axis cs:0.249096434863135,-0.113550746616246)
--(axis cs:0.24959774064154,-0.0759554721442494)
--(axis cs:0.249899333823594,-0.0380543517939387)
--(axis cs:0.25,-1.46957615897682e-16)
--cycle;
\path [draw=sandybrown24416662, fill=sandybrown24416662, opacity=0.7]
(axis cs:0.55,0.6)
--(axis cs:0.549899333823594,0.619027175896969)
--(axis cs:0.54959774064154,0.637977736072125)
--(axis cs:0.549096434863135,0.656775373308123)
--(axis cs:0.548397435069818,0.675344396154324)
--(axis cs:0.547503555887047,0.693610033709546)
--(axis cs:0.546418396650804,0.711498736698098)
--(axis cs:0.545146326914331,0.728938473626751)
--(axis cs:0.543692468853489,0.745859020830141)
--(axis cs:0.542062676641559,0.762192245236679)
--(axis cs:0.540263512876553,0.777872378716392)
--(axis cs:0.538302222155949,0.792836282905962)
--(axis cs:0.536186701905254,0.807023703444634)
--(axis cs:0.533925470577857,0.82037751259726)
--(axis cs:0.531527633354226,0.832843939287527)
--(axis cs:0.52900284547856,0.844372785615101)
--(axis cs:0.526361273380525,0.854917628984854)
--(axis cs:0.523613553738634,0.864436009034275)
--(axis cs:0.520770750650094,0.872889598606355)
--(axis cs:0.517844311079594,0.880244358079532)
--(axis cs:0.514846018766414,0.886470672433222)
--(axis cs:0.511787946775471,0.891543470497062)
--(axis cs:0.508682408883346,0.895442325903662)
--(axis cs:0.505541909995051,0.898151539338376)
--(axis cs:0.502379095791187,0.899660201754902)
--(axis cs:0.49920670180826,0.899962238302162)
--(axis cs:0.496037502157161,0.899056432785583)
--(axis cs:0.492884258086336,0.89694643256428)
--(axis cs:0.48975966659674,0.893640733864434)
--(axis cs:0.486676309315498,0.889152647567983)
--(axis cs:0.483646601834129,0.883500245614401)
--(axis cs:0.480682743715344,0.876706288231374)
--(axis cs:0.477796669369711,0.868798132287401)
--(axis cs:0.475,0.859807621135331)
--(axis cs:0.472303996806694,0.849770956390431)
--(axis cs:0.469719515643117,0.83872855215925)
--(axis cs:0.467256963302736,0.826724872306277)
--(axis cs:0.464926255614684,0.813808251413659)
--(axis cs:0.462736777516212,0.800030700154888)
--(axis cs:0.460697345262861,0.785447695866181)
--(axis cs:0.458816170928508,0.770117959158831)
--(axis cs:0.457100829338251,0.754103217472022)
--(axis cs:0.455558227567254,0.737467956518223)
--(axis cs:0.454194577128397,0.720279160621984)
--(axis cs:0.453015368960705,0.702606042997701)
--(axis cs:0.452025351319275,0.684519767052429)
--(axis cs:0.45122851065573,0.666093159835962)
--(axis cs:0.45062805556618,0.647400418792005)
--(axis cs:0.450226403871346,0.628516812991255)
--(axis cs:0.450025172880841,0.60951838004942)
--(axis cs:0.450025172880841,0.59048161995058)
--(axis cs:0.450226403871346,0.571483187008745)
--(axis cs:0.45062805556618,0.552599581207995)
--(axis cs:0.45122851065573,0.533906840164038)
--(axis cs:0.452025351319275,0.515480232947571)
--(axis cs:0.453015368960705,0.497393957002299)
--(axis cs:0.454194577128397,0.479720839378016)
--(axis cs:0.455558227567254,0.462532043481777)
--(axis cs:0.457100829338251,0.445896782527978)
--(axis cs:0.458816170928508,0.429882040841169)
--(axis cs:0.460697345262861,0.414552304133818)
--(axis cs:0.462736777516212,0.399969299845113)
--(axis cs:0.464926255614684,0.386191748586341)
--(axis cs:0.467256963302736,0.373275127693723)
--(axis cs:0.469719515643117,0.36127144784075)
--(axis cs:0.472303996806694,0.350229043609569)
--(axis cs:0.475,0.340192378864668)
--(axis cs:0.477796669369711,0.331201867712599)
--(axis cs:0.480682743715344,0.323293711768626)
--(axis cs:0.483646601834129,0.316499754385599)
--(axis cs:0.486676309315498,0.310847352432017)
--(axis cs:0.48975966659674,0.306359266135566)
--(axis cs:0.492884258086336,0.30305356743572)
--(axis cs:0.496037502157161,0.300943567214417)
--(axis cs:0.49920670180826,0.300037761697837)
--(axis cs:0.502379095791187,0.300339798245098)
--(axis cs:0.505541909995051,0.301848460661624)
--(axis cs:0.508682408883346,0.304557674096338)
--(axis cs:0.511787946775471,0.308456529502938)
--(axis cs:0.514846018766414,0.313529327566778)
--(axis cs:0.517844311079594,0.319755641920468)
--(axis cs:0.520770750650094,0.327110401393645)
--(axis cs:0.523613553738634,0.335563990965725)
--(axis cs:0.526361273380525,0.345082371015146)
--(axis cs:0.52900284547856,0.355627214384899)
--(axis cs:0.531527633354226,0.367156060712473)
--(axis cs:0.533925470577857,0.37962248740274)
--(axis cs:0.536186701905254,0.392976296555366)
--(axis cs:0.538302222155949,0.407163717094038)
--(axis cs:0.540263512876553,0.422127621283608)
--(axis cs:0.542062676641559,0.437807754763321)
--(axis cs:0.543692468853489,0.454140979169859)
--(axis cs:0.545146326914331,0.471061526373248)
--(axis cs:0.546418396650804,0.488501263301902)
--(axis cs:0.547503555887047,0.506389966290454)
--(axis cs:0.548397435069818,0.524655603845676)
--(axis cs:0.549096434863135,0.543224626691877)
--(axis cs:0.54959774064154,0.562022263927875)
--(axis cs:0.549899333823594,0.580972824103031)
--(axis cs:0.55,0.6)
--cycle;
\path [draw=sandybrown24416662, fill=sandybrown24416662, opacity=0.7]
(axis cs:0.9,0.25)
--(axis cs:0.899899333823594,0.288054351793939)
--(axis cs:0.89959774064154,0.32595547214425)
--(axis cs:0.899096434863135,0.363550746616246)
--(axis cs:0.898397435069818,0.400688792308648)
--(axis cs:0.897503555887047,0.437220067419092)
--(axis cs:0.896418396650804,0.472997473396197)
--(axis cs:0.895146326914331,0.507876947253503)
--(axis cs:0.893692468853489,0.541718041660281)
--(axis cs:0.892062676641559,0.574384490473359)
--(axis cs:0.890263512876553,0.605744757432784)
--(axis cs:0.888302222155949,0.635672565811924)
--(axis cs:0.886186701905254,0.664047406889267)
--(axis cs:0.883925470577857,0.69075502519452)
--(axis cs:0.881527633354226,0.715687878575054)
--(axis cs:0.87900284547856,0.738745571230201)
--(axis cs:0.876361273380525,0.759835257969709)
--(axis cs:0.873613553738634,0.778872018068549)
--(axis cs:0.870770750650094,0.795779197212711)
--(axis cs:0.867844311079594,0.810488716159064)
--(axis cs:0.864846018766414,0.822941344866444)
--(axis cs:0.861787946775471,0.833086940994125)
--(axis cs:0.858682408883346,0.840884651807325)
--(axis cs:0.855541909995051,0.846303078676752)
--(axis cs:0.852379095791187,0.849320403509805)
--(axis cs:0.84920670180826,0.849924476604325)
--(axis cs:0.846037502157161,0.848112865571165)
--(axis cs:0.842884258086336,0.84389286512856)
--(axis cs:0.83975966659674,0.837281467728867)
--(axis cs:0.836676309315498,0.828305295135965)
--(axis cs:0.833646601834129,0.817000491228801)
--(axis cs:0.830682743715344,0.803412576462749)
--(axis cs:0.827796669369711,0.787596264574801)
--(axis cs:0.825,0.769615242270663)
--(axis cs:0.822303996806694,0.749541912780863)
--(axis cs:0.819719515643117,0.727457104318499)
--(axis cs:0.817256963302736,0.703449744612555)
--(axis cs:0.814926255614684,0.677616502827318)
--(axis cs:0.812736777516212,0.650061400309775)
--(axis cs:0.810697345262861,0.620895391732363)
--(axis cs:0.808816170928508,0.590235918317662)
--(axis cs:0.807100829338251,0.558206434944044)
--(axis cs:0.805558227567254,0.524935913036446)
--(axis cs:0.804194577128396,0.490558321243968)
--(axis cs:0.803015368960705,0.455212085995401)
--(axis cs:0.802025351319275,0.419039534104858)
--(axis cs:0.80122851065573,0.382186319671925)
--(axis cs:0.80062805556618,0.34480083758401)
--(axis cs:0.800226403871346,0.307033625982509)
--(axis cs:0.800025172880841,0.269036760098841)
--(axis cs:0.800025172880841,0.230963239901159)
--(axis cs:0.800226403871346,0.19296637401749)
--(axis cs:0.80062805556618,0.15519916241599)
--(axis cs:0.80122851065573,0.117813680328076)
--(axis cs:0.802025351319275,0.0809604658951421)
--(axis cs:0.803015368960705,0.0447879140045988)
--(axis cs:0.804194577128396,0.00944167875603175)
--(axis cs:0.805558227567254,-0.0249359130364462)
--(axis cs:0.807100829338251,-0.0582064349440438)
--(axis cs:0.808816170928508,-0.0902359183176624)
--(axis cs:0.810697345262861,-0.120895391732363)
--(axis cs:0.812736777516212,-0.150061400309775)
--(axis cs:0.814926255614684,-0.177616502827318)
--(axis cs:0.817256963302736,-0.203449744612555)
--(axis cs:0.819719515643117,-0.227457104318499)
--(axis cs:0.822303996806694,-0.249541912780863)
--(axis cs:0.825,-0.269615242270663)
--(axis cs:0.827796669369711,-0.287596264574802)
--(axis cs:0.830682743715344,-0.303412576462749)
--(axis cs:0.833646601834129,-0.317000491228801)
--(axis cs:0.836676309315498,-0.328305295135965)
--(axis cs:0.83975966659674,-0.337281467728867)
--(axis cs:0.842884258086336,-0.34389286512856)
--(axis cs:0.846037502157161,-0.348112865571165)
--(axis cs:0.84920670180826,-0.349924476604325)
--(axis cs:0.852379095791187,-0.349320403509805)
--(axis cs:0.855541909995051,-0.346303078676752)
--(axis cs:0.858682408883346,-0.340884651807325)
--(axis cs:0.861787946775471,-0.333086940994125)
--(axis cs:0.864846018766414,-0.322941344866444)
--(axis cs:0.867844311079594,-0.310488716159064)
--(axis cs:0.870770750650094,-0.295779197212711)
--(axis cs:0.873613553738634,-0.278872018068549)
--(axis cs:0.876361273380525,-0.259835257969709)
--(axis cs:0.87900284547856,-0.238745571230201)
--(axis cs:0.881527633354226,-0.215687878575054)
--(axis cs:0.883925470577857,-0.19075502519452)
--(axis cs:0.886186701905254,-0.164047406889267)
--(axis cs:0.888302222155949,-0.135672565811924)
--(axis cs:0.890263512876553,-0.105744757432784)
--(axis cs:0.892062676641559,-0.0743844904733584)
--(axis cs:0.893692468853489,-0.0417180416602813)
--(axis cs:0.895146326914331,-0.00787694725350313)
--(axis cs:0.896418396650804,0.0270025266038037)
--(axis cs:0.897503555887047,0.0627799325809078)
--(axis cs:0.898397435069818,0.0993112076913524)
--(axis cs:0.899096434863135,0.136449253383754)
--(axis cs:0.89959774064154,0.174044527855751)
--(axis cs:0.899899333823594,0.211945648206061)
--(axis cs:0.9,0.25)
--cycle;
\addplot [semithick, seagreen4916384]
table {%
0 0
0.0505204976939265 0.0182798371148536
0.0713315428371826 0.0548734708501134
0.095048645965283 0.0817574736119185
0.133061416244574 0.10649909659226
0.146483577746418 0.153385856107389
0.152254673106304 0.198189058378051
0.138246517915944 0.246370481820829
0.147076801009949 0.289699258803772
0.14807587807814 0.32626985535305
0.145978423684528 0.370865875003341
0.157926898423926 0.428544923907377
0.14662556815667 0.465559219178399
0.166752629782772 0.50109411627452
0.178854792972924 0.552356242386681
0.182091875419026 0.60468822631357
0.223963041912746 0.595624377955626
0.235341580934606 0.574092046645532
0.26623245083368 0.540928866442042
0.27582832008662 0.486469308111204
0.30893714378574 0.474271192581891
0.341022478253002 0.441495191942871
0.359881071225778 0.408163000049208
0.401748449345228 0.38964377240894
0.415691093470473 0.343999349584227
0.430860091192157 0.31923687601873
0.466093351457339 0.289205598954272
0.494961989666634 0.290004478525121
0.529271229127352 0.308887164279222
0.557066444529498 0.31634772512194
0.593485944240497 0.321571268456648
0.60720613775827 0.357840102004949
0.640142046438189 0.396110083450989
0.685319169618678 0.40188296357846
0.719922372430511 0.423273269924387
0.756353417856807 0.449817518605631
0.799349870014707 0.485079475571023
0.799013944023749 0.54770061684922
0.80721869299208 0.602148896484936
0.803964021662291 0.626784433421355
0.810892673769595 0.685311056430057
0.808000186304584 0.716872454803799
0.817781312667145 0.7677109016527
0.822733819210302 0.817328704813591
0.846802103080967 0.868705853312726
0.873567143910192 0.836845203575459
0.874365973449426 0.82345961829952
0.873594668502718 0.782392470110538
0.893956107232108 0.733674038480577
0.897536199678369 0.69194845926789
0.905869291405653 0.655370342151573
0.924767176825919 0.601091058195213
0.942228577670404 0.557220882109151
0.941021543553543 0.518993373399821
0.952846249580552 0.499846773402802
0.963422633889253 0.453056895194349
0.988860490338262 0.412083543382325
0.989061323539868 0.363727439501407
0.986468679911234 0.307057103675884
0.996351101467136 0.276677097925319
0.990971022999264 0.226090297812658
0.976361732780205 0.185243006073056
0.997171434626326 0.150976688850726
0.978777520783788 0.114939795058411
0.974064140199064 0.0747544477889905
0.977041050074366 0.0339727065733974
1 0
};
\addplot [semithick, sandybrown24416662, opacity=0.7]
table {%
0.25 0
0.249899333823594 0.0380543517939387
0.24959774064154 0.0759554721442496
0.249096434863135 0.113550746616246
0.248397435069818 0.150688792308648
0.247503555887047 0.187220067419092
0.246418396650804 0.222997473396197
0.245146326914331 0.257876947253503
0.243692468853489 0.291718041660281
0.242062676641559 0.324384490473359
0.240263512876553 0.355744757432784
0.238302222155949 0.385672565811924
0.236186701905254 0.414047406889267
0.233925470577857 0.44075502519452
0.231527633354226 0.465687878575054
0.22900284547856 0.488745571230201
0.226361273380525 0.509835257969709
0.223613553738634 0.528872018068549
0.220770750650094 0.545779197212711
0.217844311079594 0.560488716159064
0.214846018766414 0.572941344866444
0.211787946775471 0.583086940994125
0.208682408883347 0.590884651807325
0.205541909995051 0.596303078676752
0.202379095791187 0.599320403509805
0.19920670180826 0.599924476604325
0.196037502157161 0.598112865571165
0.192884258086336 0.59389286512856
0.18975966659674 0.587281467728867
0.186676309315498 0.578305295135965
0.183646601834129 0.567000491228801
0.180682743715344 0.553412576462749
0.177796669369711 0.537596264574801
0.175 0.519615242270663
0.172303996806695 0.499541912780863
0.169719515643117 0.477457104318499
0.167256963302736 0.453449744612555
0.164926255614684 0.427616502827318
0.162736777516212 0.400061400309775
0.160697345262861 0.370895391732363
0.158816170928508 0.340235918317662
0.157100829338251 0.308206434944044
0.155558227567254 0.274935913036446
0.154194577128397 0.240558321243968
0.153015368960705 0.205212085995401
0.152025351319275 0.169039534104858
0.15122851065573 0.132186319671924
0.15062805556618 0.0948008375840099
0.150226403871346 0.0570336259825095
0.150025172880841 0.0190367600988406
0.150025172880841 -0.0190367600988407
0.150226403871346 -0.0570336259825096
0.15062805556618 -0.09480083758401
0.15122851065573 -0.132186319671924
0.152025351319275 -0.169039534104858
0.153015368960705 -0.205212085995401
0.154194577128397 -0.240558321243968
0.155558227567254 -0.274935913036446
0.157100829338251 -0.308206434944044
0.158816170928508 -0.340235918317662
0.160697345262861 -0.370895391732363
0.162736777516212 -0.400061400309775
0.164926255614684 -0.427616502827318
0.167256963302736 -0.453449744612555
0.169719515643117 -0.477457104318499
0.172303996806694 -0.499541912780863
0.175 -0.519615242270663
0.177796669369711 -0.537596264574802
0.180682743715344 -0.553412576462749
0.183646601834129 -0.567000491228801
0.186676309315498 -0.578305295135965
0.18975966659674 -0.587281467728867
0.192884258086336 -0.59389286512856
0.196037502157161 -0.598112865571165
0.19920670180826 -0.599924476604325
0.202379095791187 -0.599320403509805
0.205541909995051 -0.596303078676752
0.208682408883347 -0.590884651807325
0.211787946775471 -0.583086940994125
0.214846018766414 -0.572941344866444
0.217844311079594 -0.560488716159064
0.220770750650094 -0.545779197212711
0.223613553738634 -0.528872018068549
0.226361273380525 -0.509835257969709
0.22900284547856 -0.488745571230201
0.231527633354226 -0.465687878575054
0.233925470577857 -0.44075502519452
0.236186701905254 -0.414047406889267
0.238302222155949 -0.385672565811924
0.240263512876553 -0.355744757432784
0.242062676641559 -0.324384490473358
0.243692468853489 -0.291718041660281
0.245146326914331 -0.257876947253503
0.246418396650804 -0.222997473396196
0.247503555887047 -0.187220067419092
0.248397435069818 -0.150688792308648
0.249096434863135 -0.113550746616246
0.24959774064154 -0.0759554721442494
0.249899333823594 -0.0380543517939387
0.25 -1.46957615897682e-16
};
\addplot [semithick, sandybrown24416662, opacity=0.7]
table {%
0.55 0.6
0.549899333823594 0.619027175896969
0.54959774064154 0.637977736072125
0.549096434863135 0.656775373308123
0.548397435069818 0.675344396154324
0.547503555887047 0.693610033709546
0.546418396650804 0.711498736698098
0.545146326914331 0.728938473626751
0.543692468853489 0.745859020830141
0.542062676641559 0.762192245236679
0.540263512876553 0.777872378716392
0.538302222155949 0.792836282905962
0.536186701905254 0.807023703444634
0.533925470577857 0.82037751259726
0.531527633354226 0.832843939287527
0.52900284547856 0.844372785615101
0.526361273380525 0.854917628984854
0.523613553738634 0.864436009034275
0.520770750650094 0.872889598606355
0.517844311079594 0.880244358079532
0.514846018766414 0.886470672433222
0.511787946775471 0.891543470497062
0.508682408883346 0.895442325903662
0.505541909995051 0.898151539338376
0.502379095791187 0.899660201754902
0.49920670180826 0.899962238302162
0.496037502157161 0.899056432785583
0.492884258086336 0.89694643256428
0.48975966659674 0.893640733864434
0.486676309315498 0.889152647567983
0.483646601834129 0.883500245614401
0.480682743715344 0.876706288231374
0.477796669369711 0.868798132287401
0.475 0.859807621135331
0.472303996806694 0.849770956390431
0.469719515643117 0.83872855215925
0.467256963302736 0.826724872306277
0.464926255614684 0.813808251413659
0.462736777516212 0.800030700154888
0.460697345262861 0.785447695866181
0.458816170928508 0.770117959158831
0.457100829338251 0.754103217472022
0.455558227567254 0.737467956518223
0.454194577128397 0.720279160621984
0.453015368960705 0.702606042997701
0.452025351319275 0.684519767052429
0.45122851065573 0.666093159835962
0.45062805556618 0.647400418792005
0.450226403871346 0.628516812991255
0.450025172880841 0.60951838004942
0.450025172880841 0.59048161995058
0.450226403871346 0.571483187008745
0.45062805556618 0.552599581207995
0.45122851065573 0.533906840164038
0.452025351319275 0.515480232947571
0.453015368960705 0.497393957002299
0.454194577128397 0.479720839378016
0.455558227567254 0.462532043481777
0.457100829338251 0.445896782527978
0.458816170928508 0.429882040841169
0.460697345262861 0.414552304133818
0.462736777516212 0.399969299845113
0.464926255614684 0.386191748586341
0.467256963302736 0.373275127693723
0.469719515643117 0.36127144784075
0.472303996806694 0.350229043609569
0.475 0.340192378864668
0.477796669369711 0.331201867712599
0.480682743715344 0.323293711768626
0.483646601834129 0.316499754385599
0.486676309315498 0.310847352432017
0.48975966659674 0.306359266135566
0.492884258086336 0.30305356743572
0.496037502157161 0.300943567214417
0.49920670180826 0.300037761697837
0.502379095791187 0.300339798245098
0.505541909995051 0.301848460661624
0.508682408883346 0.304557674096338
0.511787946775471 0.308456529502938
0.514846018766414 0.313529327566778
0.517844311079594 0.319755641920468
0.520770750650094 0.327110401393645
0.523613553738634 0.335563990965725
0.526361273380525 0.345082371015146
0.52900284547856 0.355627214384899
0.531527633354226 0.367156060712473
0.533925470577857 0.37962248740274
0.536186701905254 0.392976296555366
0.538302222155949 0.407163717094038
0.540263512876553 0.422127621283608
0.542062676641559 0.437807754763321
0.543692468853489 0.454140979169859
0.545146326914331 0.471061526373248
0.546418396650804 0.488501263301902
0.547503555887047 0.506389966290454
0.548397435069818 0.524655603845676
0.549096434863135 0.543224626691877
0.54959774064154 0.562022263927875
0.549899333823594 0.580972824103031
0.55 0.6
};
\addplot [semithick, sandybrown24416662, opacity=0.7]
table {%
0.9 0.25
0.899899333823594 0.288054351793939
0.89959774064154 0.32595547214425
0.899096434863135 0.363550746616246
0.898397435069818 0.400688792308648
0.897503555887047 0.437220067419092
0.896418396650804 0.472997473396197
0.895146326914331 0.507876947253503
0.893692468853489 0.541718041660281
0.892062676641559 0.574384490473359
0.890263512876553 0.605744757432784
0.888302222155949 0.635672565811924
0.886186701905254 0.664047406889267
0.883925470577857 0.69075502519452
0.881527633354226 0.715687878575054
0.87900284547856 0.738745571230201
0.876361273380525 0.759835257969709
0.873613553738634 0.778872018068549
0.870770750650094 0.795779197212711
0.867844311079594 0.810488716159064
0.864846018766414 0.822941344866444
0.861787946775471 0.833086940994125
0.858682408883346 0.840884651807325
0.855541909995051 0.846303078676752
0.852379095791187 0.849320403509805
0.84920670180826 0.849924476604325
0.846037502157161 0.848112865571165
0.842884258086336 0.84389286512856
0.83975966659674 0.837281467728867
0.836676309315498 0.828305295135965
0.833646601834129 0.817000491228801
0.830682743715344 0.803412576462749
0.827796669369711 0.787596264574801
0.825 0.769615242270663
0.822303996806694 0.749541912780863
0.819719515643117 0.727457104318499
0.817256963302736 0.703449744612555
0.814926255614684 0.677616502827318
0.812736777516212 0.650061400309775
0.810697345262861 0.620895391732363
0.808816170928508 0.590235918317662
0.807100829338251 0.558206434944044
0.805558227567254 0.524935913036446
0.804194577128396 0.490558321243968
0.803015368960705 0.455212085995401
0.802025351319275 0.419039534104858
0.80122851065573 0.382186319671925
0.80062805556618 0.34480083758401
0.800226403871346 0.307033625982509
0.800025172880841 0.269036760098841
0.800025172880841 0.230963239901159
0.800226403871346 0.19296637401749
0.80062805556618 0.15519916241599
0.80122851065573 0.117813680328076
0.802025351319275 0.0809604658951421
0.803015368960705 0.0447879140045988
0.804194577128396 0.00944167875603175
0.805558227567254 -0.0249359130364462
0.807100829338251 -0.0582064349440438
0.808816170928508 -0.0902359183176624
0.810697345262861 -0.120895391732363
0.812736777516212 -0.150061400309775
0.814926255614684 -0.177616502827318
0.817256963302736 -0.203449744612555
0.819719515643117 -0.227457104318499
0.822303996806694 -0.249541912780863
0.825 -0.269615242270663
0.827796669369711 -0.287596264574802
0.830682743715344 -0.303412576462749
0.833646601834129 -0.317000491228801
0.836676309315498 -0.328305295135965
0.83975966659674 -0.337281467728867
0.842884258086336 -0.34389286512856
0.846037502157161 -0.348112865571165
0.84920670180826 -0.349924476604325
0.852379095791187 -0.349320403509805
0.855541909995051 -0.346303078676752
0.858682408883346 -0.340884651807325
0.861787946775471 -0.333086940994125
0.864846018766414 -0.322941344866444
0.867844311079594 -0.310488716159064
0.870770750650094 -0.295779197212711
0.873613553738634 -0.278872018068549
0.876361273380525 -0.259835257969709
0.87900284547856 -0.238745571230201
0.881527633354226 -0.215687878575054
0.883925470577857 -0.19075502519452
0.886186701905254 -0.164047406889267
0.888302222155949 -0.135672565811924
0.890263512876553 -0.105744757432784
0.892062676641559 -0.0743844904733584
0.893692468853489 -0.0417180416602813
0.895146326914331 -0.00787694725350313
0.896418396650804 0.0270025266038037
0.897503555887047 0.0627799325809078
0.898397435069818 0.0993112076913524
0.899096434863135 0.136449253383754
0.89959774064154 0.174044527855751
0.899899333823594 0.211945648206061
0.9 0.25
};
\draw (axis cs:0.02,0.98) node[
  scale=0.85,
  anchor=north west,
  text=black,
  rotate=0.0
]{$\gamma=$1.0};
\draw (axis cs:0.02,0.895) node[
  scale=0.85,
  anchor=north west,
  text=black,
  rotate=0.0
]{$\overline{C}(\sigma^*)=4.40$};
\end{axis}

\end{tikzpicture}

%% file: icra2024/tex/appendix/varying-gammas/gamma-10.0.tex
\begin{tikzpicture}

\definecolor{sandybrown24416662}{RGB}{244,166,62}
\definecolor{seagreen4916384}{RGB}{49,163,84}

\begin{axis}[
width=0.33\linewidth, 
height=0.33\linewidth, 
tick pos=left,
xlabel={x},
xmin=0, xmax=1,
xtick style={color=black},
ylabel={y},
ymin=0, ymax=1,
ytick style={color=black},
]
\path [draw=sandybrown24416662, fill=sandybrown24416662, opacity=0.7]
(axis cs:0.25,0)
--(axis cs:0.249899333823594,0.0380543517939387)
--(axis cs:0.24959774064154,0.0759554721442496)
--(axis cs:0.249096434863135,0.113550746616246)
--(axis cs:0.248397435069818,0.150688792308648)
--(axis cs:0.247503555887047,0.187220067419092)
--(axis cs:0.246418396650804,0.222997473396197)
--(axis cs:0.245146326914331,0.257876947253503)
--(axis cs:0.243692468853489,0.291718041660281)
--(axis cs:0.242062676641559,0.324384490473359)
--(axis cs:0.240263512876553,0.355744757432784)
--(axis cs:0.238302222155949,0.385672565811924)
--(axis cs:0.236186701905254,0.414047406889267)
--(axis cs:0.233925470577857,0.44075502519452)
--(axis cs:0.231527633354226,0.465687878575054)
--(axis cs:0.22900284547856,0.488745571230201)
--(axis cs:0.226361273380525,0.509835257969709)
--(axis cs:0.223613553738634,0.528872018068549)
--(axis cs:0.220770750650094,0.545779197212711)
--(axis cs:0.217844311079594,0.560488716159064)
--(axis cs:0.214846018766414,0.572941344866444)
--(axis cs:0.211787946775471,0.583086940994125)
--(axis cs:0.208682408883347,0.590884651807325)
--(axis cs:0.205541909995051,0.596303078676752)
--(axis cs:0.202379095791187,0.599320403509805)
--(axis cs:0.19920670180826,0.599924476604325)
--(axis cs:0.196037502157161,0.598112865571165)
--(axis cs:0.192884258086336,0.59389286512856)
--(axis cs:0.18975966659674,0.587281467728867)
--(axis cs:0.186676309315498,0.578305295135965)
--(axis cs:0.183646601834129,0.567000491228801)
--(axis cs:0.180682743715344,0.553412576462749)
--(axis cs:0.177796669369711,0.537596264574801)
--(axis cs:0.175,0.519615242270663)
--(axis cs:0.172303996806695,0.499541912780863)
--(axis cs:0.169719515643117,0.477457104318499)
--(axis cs:0.167256963302736,0.453449744612555)
--(axis cs:0.164926255614684,0.427616502827318)
--(axis cs:0.162736777516212,0.400061400309775)
--(axis cs:0.160697345262861,0.370895391732363)
--(axis cs:0.158816170928508,0.340235918317662)
--(axis cs:0.157100829338251,0.308206434944044)
--(axis cs:0.155558227567254,0.274935913036446)
--(axis cs:0.154194577128397,0.240558321243968)
--(axis cs:0.153015368960705,0.205212085995401)
--(axis cs:0.152025351319275,0.169039534104858)
--(axis cs:0.15122851065573,0.132186319671924)
--(axis cs:0.15062805556618,0.0948008375840099)
--(axis cs:0.150226403871346,0.0570336259825095)
--(axis cs:0.150025172880841,0.0190367600988406)
--(axis cs:0.150025172880841,-0.0190367600988407)
--(axis cs:0.150226403871346,-0.0570336259825096)
--(axis cs:0.15062805556618,-0.09480083758401)
--(axis cs:0.15122851065573,-0.132186319671924)
--(axis cs:0.152025351319275,-0.169039534104858)
--(axis cs:0.153015368960705,-0.205212085995401)
--(axis cs:0.154194577128397,-0.240558321243968)
--(axis cs:0.155558227567254,-0.274935913036446)
--(axis cs:0.157100829338251,-0.308206434944044)
--(axis cs:0.158816170928508,-0.340235918317662)
--(axis cs:0.160697345262861,-0.370895391732363)
--(axis cs:0.162736777516212,-0.400061400309775)
--(axis cs:0.164926255614684,-0.427616502827318)
--(axis cs:0.167256963302736,-0.453449744612555)
--(axis cs:0.169719515643117,-0.477457104318499)
--(axis cs:0.172303996806694,-0.499541912780863)
--(axis cs:0.175,-0.519615242270663)
--(axis cs:0.177796669369711,-0.537596264574802)
--(axis cs:0.180682743715344,-0.553412576462749)
--(axis cs:0.183646601834129,-0.567000491228801)
--(axis cs:0.186676309315498,-0.578305295135965)
--(axis cs:0.18975966659674,-0.587281467728867)
--(axis cs:0.192884258086336,-0.59389286512856)
--(axis cs:0.196037502157161,-0.598112865571165)
--(axis cs:0.19920670180826,-0.599924476604325)
--(axis cs:0.202379095791187,-0.599320403509805)
--(axis cs:0.205541909995051,-0.596303078676752)
--(axis cs:0.208682408883347,-0.590884651807325)
--(axis cs:0.211787946775471,-0.583086940994125)
--(axis cs:0.214846018766414,-0.572941344866444)
--(axis cs:0.217844311079594,-0.560488716159064)
--(axis cs:0.220770750650094,-0.545779197212711)
--(axis cs:0.223613553738634,-0.528872018068549)
--(axis cs:0.226361273380525,-0.509835257969709)
--(axis cs:0.22900284547856,-0.488745571230201)
--(axis cs:0.231527633354226,-0.465687878575054)
--(axis cs:0.233925470577857,-0.44075502519452)
--(axis cs:0.236186701905254,-0.414047406889267)
--(axis cs:0.238302222155949,-0.385672565811924)
--(axis cs:0.240263512876553,-0.355744757432784)
--(axis cs:0.242062676641559,-0.324384490473358)
--(axis cs:0.243692468853489,-0.291718041660281)
--(axis cs:0.245146326914331,-0.257876947253503)
--(axis cs:0.246418396650804,-0.222997473396196)
--(axis cs:0.247503555887047,-0.187220067419092)
--(axis cs:0.248397435069818,-0.150688792308648)
--(axis cs:0.249096434863135,-0.113550746616246)
--(axis cs:0.24959774064154,-0.0759554721442494)
--(axis cs:0.249899333823594,-0.0380543517939387)
--(axis cs:0.25,-1.46957615897682e-16)
--cycle;
\path [draw=sandybrown24416662, fill=sandybrown24416662, opacity=0.7]
(axis cs:0.55,0.6)
--(axis cs:0.549899333823594,0.619027175896969)
--(axis cs:0.54959774064154,0.637977736072125)
--(axis cs:0.549096434863135,0.656775373308123)
--(axis cs:0.548397435069818,0.675344396154324)
--(axis cs:0.547503555887047,0.693610033709546)
--(axis cs:0.546418396650804,0.711498736698098)
--(axis cs:0.545146326914331,0.728938473626751)
--(axis cs:0.543692468853489,0.745859020830141)
--(axis cs:0.542062676641559,0.762192245236679)
--(axis cs:0.540263512876553,0.777872378716392)
--(axis cs:0.538302222155949,0.792836282905962)
--(axis cs:0.536186701905254,0.807023703444634)
--(axis cs:0.533925470577857,0.82037751259726)
--(axis cs:0.531527633354226,0.832843939287527)
--(axis cs:0.52900284547856,0.844372785615101)
--(axis cs:0.526361273380525,0.854917628984854)
--(axis cs:0.523613553738634,0.864436009034275)
--(axis cs:0.520770750650094,0.872889598606355)
--(axis cs:0.517844311079594,0.880244358079532)
--(axis cs:0.514846018766414,0.886470672433222)
--(axis cs:0.511787946775471,0.891543470497062)
--(axis cs:0.508682408883346,0.895442325903662)
--(axis cs:0.505541909995051,0.898151539338376)
--(axis cs:0.502379095791187,0.899660201754902)
--(axis cs:0.49920670180826,0.899962238302162)
--(axis cs:0.496037502157161,0.899056432785583)
--(axis cs:0.492884258086336,0.89694643256428)
--(axis cs:0.48975966659674,0.893640733864434)
--(axis cs:0.486676309315498,0.889152647567983)
--(axis cs:0.483646601834129,0.883500245614401)
--(axis cs:0.480682743715344,0.876706288231374)
--(axis cs:0.477796669369711,0.868798132287401)
--(axis cs:0.475,0.859807621135331)
--(axis cs:0.472303996806694,0.849770956390431)
--(axis cs:0.469719515643117,0.83872855215925)
--(axis cs:0.467256963302736,0.826724872306277)
--(axis cs:0.464926255614684,0.813808251413659)
--(axis cs:0.462736777516212,0.800030700154888)
--(axis cs:0.460697345262861,0.785447695866181)
--(axis cs:0.458816170928508,0.770117959158831)
--(axis cs:0.457100829338251,0.754103217472022)
--(axis cs:0.455558227567254,0.737467956518223)
--(axis cs:0.454194577128397,0.720279160621984)
--(axis cs:0.453015368960705,0.702606042997701)
--(axis cs:0.452025351319275,0.684519767052429)
--(axis cs:0.45122851065573,0.666093159835962)
--(axis cs:0.45062805556618,0.647400418792005)
--(axis cs:0.450226403871346,0.628516812991255)
--(axis cs:0.450025172880841,0.60951838004942)
--(axis cs:0.450025172880841,0.59048161995058)
--(axis cs:0.450226403871346,0.571483187008745)
--(axis cs:0.45062805556618,0.552599581207995)
--(axis cs:0.45122851065573,0.533906840164038)
--(axis cs:0.452025351319275,0.515480232947571)
--(axis cs:0.453015368960705,0.497393957002299)
--(axis cs:0.454194577128397,0.479720839378016)
--(axis cs:0.455558227567254,0.462532043481777)
--(axis cs:0.457100829338251,0.445896782527978)
--(axis cs:0.458816170928508,0.429882040841169)
--(axis cs:0.460697345262861,0.414552304133818)
--(axis cs:0.462736777516212,0.399969299845113)
--(axis cs:0.464926255614684,0.386191748586341)
--(axis cs:0.467256963302736,0.373275127693723)
--(axis cs:0.469719515643117,0.36127144784075)
--(axis cs:0.472303996806694,0.350229043609569)
--(axis cs:0.475,0.340192378864668)
--(axis cs:0.477796669369711,0.331201867712599)
--(axis cs:0.480682743715344,0.323293711768626)
--(axis cs:0.483646601834129,0.316499754385599)
--(axis cs:0.486676309315498,0.310847352432017)
--(axis cs:0.48975966659674,0.306359266135566)
--(axis cs:0.492884258086336,0.30305356743572)
--(axis cs:0.496037502157161,0.300943567214417)
--(axis cs:0.49920670180826,0.300037761697837)
--(axis cs:0.502379095791187,0.300339798245098)
--(axis cs:0.505541909995051,0.301848460661624)
--(axis cs:0.508682408883346,0.304557674096338)
--(axis cs:0.511787946775471,0.308456529502938)
--(axis cs:0.514846018766414,0.313529327566778)
--(axis cs:0.517844311079594,0.319755641920468)
--(axis cs:0.520770750650094,0.327110401393645)
--(axis cs:0.523613553738634,0.335563990965725)
--(axis cs:0.526361273380525,0.345082371015146)
--(axis cs:0.52900284547856,0.355627214384899)
--(axis cs:0.531527633354226,0.367156060712473)
--(axis cs:0.533925470577857,0.37962248740274)
--(axis cs:0.536186701905254,0.392976296555366)
--(axis cs:0.538302222155949,0.407163717094038)
--(axis cs:0.540263512876553,0.422127621283608)
--(axis cs:0.542062676641559,0.437807754763321)
--(axis cs:0.543692468853489,0.454140979169859)
--(axis cs:0.545146326914331,0.471061526373248)
--(axis cs:0.546418396650804,0.488501263301902)
--(axis cs:0.547503555887047,0.506389966290454)
--(axis cs:0.548397435069818,0.524655603845676)
--(axis cs:0.549096434863135,0.543224626691877)
--(axis cs:0.54959774064154,0.562022263927875)
--(axis cs:0.549899333823594,0.580972824103031)
--(axis cs:0.55,0.6)
--cycle;
\path [draw=sandybrown24416662, fill=sandybrown24416662, opacity=0.7]
(axis cs:0.9,0.25)
--(axis cs:0.899899333823594,0.288054351793939)
--(axis cs:0.89959774064154,0.32595547214425)
--(axis cs:0.899096434863135,0.363550746616246)
--(axis cs:0.898397435069818,0.400688792308648)
--(axis cs:0.897503555887047,0.437220067419092)
--(axis cs:0.896418396650804,0.472997473396197)
--(axis cs:0.895146326914331,0.507876947253503)
--(axis cs:0.893692468853489,0.541718041660281)
--(axis cs:0.892062676641559,0.574384490473359)
--(axis cs:0.890263512876553,0.605744757432784)
--(axis cs:0.888302222155949,0.635672565811924)
--(axis cs:0.886186701905254,0.664047406889267)
--(axis cs:0.883925470577857,0.69075502519452)
--(axis cs:0.881527633354226,0.715687878575054)
--(axis cs:0.87900284547856,0.738745571230201)
--(axis cs:0.876361273380525,0.759835257969709)
--(axis cs:0.873613553738634,0.778872018068549)
--(axis cs:0.870770750650094,0.795779197212711)
--(axis cs:0.867844311079594,0.810488716159064)
--(axis cs:0.864846018766414,0.822941344866444)
--(axis cs:0.861787946775471,0.833086940994125)
--(axis cs:0.858682408883346,0.840884651807325)
--(axis cs:0.855541909995051,0.846303078676752)
--(axis cs:0.852379095791187,0.849320403509805)
--(axis cs:0.84920670180826,0.849924476604325)
--(axis cs:0.846037502157161,0.848112865571165)
--(axis cs:0.842884258086336,0.84389286512856)
--(axis cs:0.83975966659674,0.837281467728867)
--(axis cs:0.836676309315498,0.828305295135965)
--(axis cs:0.833646601834129,0.817000491228801)
--(axis cs:0.830682743715344,0.803412576462749)
--(axis cs:0.827796669369711,0.787596264574801)
--(axis cs:0.825,0.769615242270663)
--(axis cs:0.822303996806694,0.749541912780863)
--(axis cs:0.819719515643117,0.727457104318499)
--(axis cs:0.817256963302736,0.703449744612555)
--(axis cs:0.814926255614684,0.677616502827318)
--(axis cs:0.812736777516212,0.650061400309775)
--(axis cs:0.810697345262861,0.620895391732363)
--(axis cs:0.808816170928508,0.590235918317662)
--(axis cs:0.807100829338251,0.558206434944044)
--(axis cs:0.805558227567254,0.524935913036446)
--(axis cs:0.804194577128396,0.490558321243968)
--(axis cs:0.803015368960705,0.455212085995401)
--(axis cs:0.802025351319275,0.419039534104858)
--(axis cs:0.80122851065573,0.382186319671925)
--(axis cs:0.80062805556618,0.34480083758401)
--(axis cs:0.800226403871346,0.307033625982509)
--(axis cs:0.800025172880841,0.269036760098841)
--(axis cs:0.800025172880841,0.230963239901159)
--(axis cs:0.800226403871346,0.19296637401749)
--(axis cs:0.80062805556618,0.15519916241599)
--(axis cs:0.80122851065573,0.117813680328076)
--(axis cs:0.802025351319275,0.0809604658951421)
--(axis cs:0.803015368960705,0.0447879140045988)
--(axis cs:0.804194577128396,0.00944167875603175)
--(axis cs:0.805558227567254,-0.0249359130364462)
--(axis cs:0.807100829338251,-0.0582064349440438)
--(axis cs:0.808816170928508,-0.0902359183176624)
--(axis cs:0.810697345262861,-0.120895391732363)
--(axis cs:0.812736777516212,-0.150061400309775)
--(axis cs:0.814926255614684,-0.177616502827318)
--(axis cs:0.817256963302736,-0.203449744612555)
--(axis cs:0.819719515643117,-0.227457104318499)
--(axis cs:0.822303996806694,-0.249541912780863)
--(axis cs:0.825,-0.269615242270663)
--(axis cs:0.827796669369711,-0.287596264574802)
--(axis cs:0.830682743715344,-0.303412576462749)
--(axis cs:0.833646601834129,-0.317000491228801)
--(axis cs:0.836676309315498,-0.328305295135965)
--(axis cs:0.83975966659674,-0.337281467728867)
--(axis cs:0.842884258086336,-0.34389286512856)
--(axis cs:0.846037502157161,-0.348112865571165)
--(axis cs:0.84920670180826,-0.349924476604325)
--(axis cs:0.852379095791187,-0.349320403509805)
--(axis cs:0.855541909995051,-0.346303078676752)
--(axis cs:0.858682408883346,-0.340884651807325)
--(axis cs:0.861787946775471,-0.333086940994125)
--(axis cs:0.864846018766414,-0.322941344866444)
--(axis cs:0.867844311079594,-0.310488716159064)
--(axis cs:0.870770750650094,-0.295779197212711)
--(axis cs:0.873613553738634,-0.278872018068549)
--(axis cs:0.876361273380525,-0.259835257969709)
--(axis cs:0.87900284547856,-0.238745571230201)
--(axis cs:0.881527633354226,-0.215687878575054)
--(axis cs:0.883925470577857,-0.19075502519452)
--(axis cs:0.886186701905254,-0.164047406889267)
--(axis cs:0.888302222155949,-0.135672565811924)
--(axis cs:0.890263512876553,-0.105744757432784)
--(axis cs:0.892062676641559,-0.0743844904733584)
--(axis cs:0.893692468853489,-0.0417180416602813)
--(axis cs:0.895146326914331,-0.00787694725350313)
--(axis cs:0.896418396650804,0.0270025266038037)
--(axis cs:0.897503555887047,0.0627799325809078)
--(axis cs:0.898397435069818,0.0993112076913524)
--(axis cs:0.899096434863135,0.136449253383754)
--(axis cs:0.89959774064154,0.174044527855751)
--(axis cs:0.899899333823594,0.211945648206061)
--(axis cs:0.9,0.25)
--cycle;
\addplot [semithick, seagreen4916384]
table {%
0 0
0.0490156140555273 0.0832509305821063
0.0636057776523686 0.108620392718662
0.0808313107970092 0.139772945848483
0.0950554399047567 0.181673579966508
0.112216374050462 0.217006194005412
0.14043180409918 0.236171345062902
0.155699871480102 0.287811924286973
0.152204581899302 0.355331976591327
0.145050099712613 0.407526994479077
0.152950109360554 0.438281167065104
0.163211878169489 0.485283742799866
0.178010439050361 0.551073300704557
0.195220092196048 0.610197861067546
0.233417920086439 0.632833839709447
0.263900537222504 0.635353071236955
0.290505797164809 0.654644581650745
0.33162418348926 0.672543078702931
0.361827219764965 0.690488413482085
0.381780343266851 0.70457772459223
0.41052367177761 0.733267237385156
0.43616267677187 0.759911072007614
0.458750475989444 0.795676329087484
0.465322996048963 0.855790166636034
0.478092232661824 0.896186783889623
0.505994582454472 0.903108900719496
0.542318279208502 0.897796230101103
0.578947206536471 0.868060720544178
0.616604273472453 0.855779857151239
0.644578051581519 0.851949830732186
0.668257147786704 0.850984705307893
0.701131219539018 0.837629788973769
0.73234123968047 0.835938779562769
0.770940378520366 0.834320180238877
0.815908781077284 0.829904752690322
0.849215229416412 0.853285387098393
0.879404762392081 0.84550784523172
0.893184335077229 0.817211692499823
0.905551167410039 0.776171463363608
0.886761499817621 0.724145671658784
0.912407746489833 0.670254660847651
0.917360122014134 0.632132057571701
0.904905546712521 0.594539769522423
0.908474144422893 0.554302601490306
0.905459704924582 0.550365854462306
0.904730785440181 0.48606361575828
0.90025976005171 0.43587768693215
0.909671871934522 0.385749272815494
0.922517551113239 0.333211800233191
0.936030190571041 0.294286903583482
0.948243117201689 0.239202765409149
0.969495187001755 0.215193213480727
0.974206430423224 0.205323736482922
0.972335996781098 0.146280509882264
0.987846428417866 0.101561070907336
0.98928959471885 0.0790639739702621
0.987271268326539 0.0586796432134907
0.994116614563094 0.0316367776202866
1 0
};
\addplot [semithick, sandybrown24416662, opacity=0.7]
table {%
0.25 0
0.249899333823594 0.0380543517939387
0.24959774064154 0.0759554721442496
0.249096434863135 0.113550746616246
0.248397435069818 0.150688792308648
0.247503555887047 0.187220067419092
0.246418396650804 0.222997473396197
0.245146326914331 0.257876947253503
0.243692468853489 0.291718041660281
0.242062676641559 0.324384490473359
0.240263512876553 0.355744757432784
0.238302222155949 0.385672565811924
0.236186701905254 0.414047406889267
0.233925470577857 0.44075502519452
0.231527633354226 0.465687878575054
0.22900284547856 0.488745571230201
0.226361273380525 0.509835257969709
0.223613553738634 0.528872018068549
0.220770750650094 0.545779197212711
0.217844311079594 0.560488716159064
0.214846018766414 0.572941344866444
0.211787946775471 0.583086940994125
0.208682408883347 0.590884651807325
0.205541909995051 0.596303078676752
0.202379095791187 0.599320403509805
0.19920670180826 0.599924476604325
0.196037502157161 0.598112865571165
0.192884258086336 0.59389286512856
0.18975966659674 0.587281467728867
0.186676309315498 0.578305295135965
0.183646601834129 0.567000491228801
0.180682743715344 0.553412576462749
0.177796669369711 0.537596264574801
0.175 0.519615242270663
0.172303996806695 0.499541912780863
0.169719515643117 0.477457104318499
0.167256963302736 0.453449744612555
0.164926255614684 0.427616502827318
0.162736777516212 0.400061400309775
0.160697345262861 0.370895391732363
0.158816170928508 0.340235918317662
0.157100829338251 0.308206434944044
0.155558227567254 0.274935913036446
0.154194577128397 0.240558321243968
0.153015368960705 0.205212085995401
0.152025351319275 0.169039534104858
0.15122851065573 0.132186319671924
0.15062805556618 0.0948008375840099
0.150226403871346 0.0570336259825095
0.150025172880841 0.0190367600988406
0.150025172880841 -0.0190367600988407
0.150226403871346 -0.0570336259825096
0.15062805556618 -0.09480083758401
0.15122851065573 -0.132186319671924
0.152025351319275 -0.169039534104858
0.153015368960705 -0.205212085995401
0.154194577128397 -0.240558321243968
0.155558227567254 -0.274935913036446
0.157100829338251 -0.308206434944044
0.158816170928508 -0.340235918317662
0.160697345262861 -0.370895391732363
0.162736777516212 -0.400061400309775
0.164926255614684 -0.427616502827318
0.167256963302736 -0.453449744612555
0.169719515643117 -0.477457104318499
0.172303996806694 -0.499541912780863
0.175 -0.519615242270663
0.177796669369711 -0.537596264574802
0.180682743715344 -0.553412576462749
0.183646601834129 -0.567000491228801
0.186676309315498 -0.578305295135965
0.18975966659674 -0.587281467728867
0.192884258086336 -0.59389286512856
0.196037502157161 -0.598112865571165
0.19920670180826 -0.599924476604325
0.202379095791187 -0.599320403509805
0.205541909995051 -0.596303078676752
0.208682408883347 -0.590884651807325
0.211787946775471 -0.583086940994125
0.214846018766414 -0.572941344866444
0.217844311079594 -0.560488716159064
0.220770750650094 -0.545779197212711
0.223613553738634 -0.528872018068549
0.226361273380525 -0.509835257969709
0.22900284547856 -0.488745571230201
0.231527633354226 -0.465687878575054
0.233925470577857 -0.44075502519452
0.236186701905254 -0.414047406889267
0.238302222155949 -0.385672565811924
0.240263512876553 -0.355744757432784
0.242062676641559 -0.324384490473358
0.243692468853489 -0.291718041660281
0.245146326914331 -0.257876947253503
0.246418396650804 -0.222997473396196
0.247503555887047 -0.187220067419092
0.248397435069818 -0.150688792308648
0.249096434863135 -0.113550746616246
0.24959774064154 -0.0759554721442494
0.249899333823594 -0.0380543517939387
0.25 -1.46957615897682e-16
};
\addplot [semithick, sandybrown24416662, opacity=0.7]
table {%
0.55 0.6
0.549899333823594 0.619027175896969
0.54959774064154 0.637977736072125
0.549096434863135 0.656775373308123
0.548397435069818 0.675344396154324
0.547503555887047 0.693610033709546
0.546418396650804 0.711498736698098
0.545146326914331 0.728938473626751
0.543692468853489 0.745859020830141
0.542062676641559 0.762192245236679
0.540263512876553 0.777872378716392
0.538302222155949 0.792836282905962
0.536186701905254 0.807023703444634
0.533925470577857 0.82037751259726
0.531527633354226 0.832843939287527
0.52900284547856 0.844372785615101
0.526361273380525 0.854917628984854
0.523613553738634 0.864436009034275
0.520770750650094 0.872889598606355
0.517844311079594 0.880244358079532
0.514846018766414 0.886470672433222
0.511787946775471 0.891543470497062
0.508682408883346 0.895442325903662
0.505541909995051 0.898151539338376
0.502379095791187 0.899660201754902
0.49920670180826 0.899962238302162
0.496037502157161 0.899056432785583
0.492884258086336 0.89694643256428
0.48975966659674 0.893640733864434
0.486676309315498 0.889152647567983
0.483646601834129 0.883500245614401
0.480682743715344 0.876706288231374
0.477796669369711 0.868798132287401
0.475 0.859807621135331
0.472303996806694 0.849770956390431
0.469719515643117 0.83872855215925
0.467256963302736 0.826724872306277
0.464926255614684 0.813808251413659
0.462736777516212 0.800030700154888
0.460697345262861 0.785447695866181
0.458816170928508 0.770117959158831
0.457100829338251 0.754103217472022
0.455558227567254 0.737467956518223
0.454194577128397 0.720279160621984
0.453015368960705 0.702606042997701
0.452025351319275 0.684519767052429
0.45122851065573 0.666093159835962
0.45062805556618 0.647400418792005
0.450226403871346 0.628516812991255
0.450025172880841 0.60951838004942
0.450025172880841 0.59048161995058
0.450226403871346 0.571483187008745
0.45062805556618 0.552599581207995
0.45122851065573 0.533906840164038
0.452025351319275 0.515480232947571
0.453015368960705 0.497393957002299
0.454194577128397 0.479720839378016
0.455558227567254 0.462532043481777
0.457100829338251 0.445896782527978
0.458816170928508 0.429882040841169
0.460697345262861 0.414552304133818
0.462736777516212 0.399969299845113
0.464926255614684 0.386191748586341
0.467256963302736 0.373275127693723
0.469719515643117 0.36127144784075
0.472303996806694 0.350229043609569
0.475 0.340192378864668
0.477796669369711 0.331201867712599
0.480682743715344 0.323293711768626
0.483646601834129 0.316499754385599
0.486676309315498 0.310847352432017
0.48975966659674 0.306359266135566
0.492884258086336 0.30305356743572
0.496037502157161 0.300943567214417
0.49920670180826 0.300037761697837
0.502379095791187 0.300339798245098
0.505541909995051 0.301848460661624
0.508682408883346 0.304557674096338
0.511787946775471 0.308456529502938
0.514846018766414 0.313529327566778
0.517844311079594 0.319755641920468
0.520770750650094 0.327110401393645
0.523613553738634 0.335563990965725
0.526361273380525 0.345082371015146
0.52900284547856 0.355627214384899
0.531527633354226 0.367156060712473
0.533925470577857 0.37962248740274
0.536186701905254 0.392976296555366
0.538302222155949 0.407163717094038
0.540263512876553 0.422127621283608
0.542062676641559 0.437807754763321
0.543692468853489 0.454140979169859
0.545146326914331 0.471061526373248
0.546418396650804 0.488501263301902
0.547503555887047 0.506389966290454
0.548397435069818 0.524655603845676
0.549096434863135 0.543224626691877
0.54959774064154 0.562022263927875
0.549899333823594 0.580972824103031
0.55 0.6
};
\addplot [semithick, sandybrown24416662, opacity=0.7]
table {%
0.9 0.25
0.899899333823594 0.288054351793939
0.89959774064154 0.32595547214425
0.899096434863135 0.363550746616246
0.898397435069818 0.400688792308648
0.897503555887047 0.437220067419092
0.896418396650804 0.472997473396197
0.895146326914331 0.507876947253503
0.893692468853489 0.541718041660281
0.892062676641559 0.574384490473359
0.890263512876553 0.605744757432784
0.888302222155949 0.635672565811924
0.886186701905254 0.664047406889267
0.883925470577857 0.69075502519452
0.881527633354226 0.715687878575054
0.87900284547856 0.738745571230201
0.876361273380525 0.759835257969709
0.873613553738634 0.778872018068549
0.870770750650094 0.795779197212711
0.867844311079594 0.810488716159064
0.864846018766414 0.822941344866444
0.861787946775471 0.833086940994125
0.858682408883346 0.840884651807325
0.855541909995051 0.846303078676752
0.852379095791187 0.849320403509805
0.84920670180826 0.849924476604325
0.846037502157161 0.848112865571165
0.842884258086336 0.84389286512856
0.83975966659674 0.837281467728867
0.836676309315498 0.828305295135965
0.833646601834129 0.817000491228801
0.830682743715344 0.803412576462749
0.827796669369711 0.787596264574801
0.825 0.769615242270663
0.822303996806694 0.749541912780863
0.819719515643117 0.727457104318499
0.817256963302736 0.703449744612555
0.814926255614684 0.677616502827318
0.812736777516212 0.650061400309775
0.810697345262861 0.620895391732363
0.808816170928508 0.590235918317662
0.807100829338251 0.558206434944044
0.805558227567254 0.524935913036446
0.804194577128396 0.490558321243968
0.803015368960705 0.455212085995401
0.802025351319275 0.419039534104858
0.80122851065573 0.382186319671925
0.80062805556618 0.34480083758401
0.800226403871346 0.307033625982509
0.800025172880841 0.269036760098841
0.800025172880841 0.230963239901159
0.800226403871346 0.19296637401749
0.80062805556618 0.15519916241599
0.80122851065573 0.117813680328076
0.802025351319275 0.0809604658951421
0.803015368960705 0.0447879140045988
0.804194577128396 0.00944167875603175
0.805558227567254 -0.0249359130364462
0.807100829338251 -0.0582064349440438
0.808816170928508 -0.0902359183176624
0.810697345262861 -0.120895391732363
0.812736777516212 -0.150061400309775
0.814926255614684 -0.177616502827318
0.817256963302736 -0.203449744612555
0.819719515643117 -0.227457104318499
0.822303996806694 -0.249541912780863
0.825 -0.269615242270663
0.827796669369711 -0.287596264574802
0.830682743715344 -0.303412576462749
0.833646601834129 -0.317000491228801
0.836676309315498 -0.328305295135965
0.83975966659674 -0.337281467728867
0.842884258086336 -0.34389286512856
0.846037502157161 -0.348112865571165
0.84920670180826 -0.349924476604325
0.852379095791187 -0.349320403509805
0.855541909995051 -0.346303078676752
0.858682408883346 -0.340884651807325
0.861787946775471 -0.333086940994125
0.864846018766414 -0.322941344866444
0.867844311079594 -0.310488716159064
0.870770750650094 -0.295779197212711
0.873613553738634 -0.278872018068549
0.876361273380525 -0.259835257969709
0.87900284547856 -0.238745571230201
0.881527633354226 -0.215687878575054
0.883925470577857 -0.19075502519452
0.886186701905254 -0.164047406889267
0.888302222155949 -0.135672565811924
0.890263512876553 -0.105744757432784
0.892062676641559 -0.0743844904733584
0.893692468853489 -0.0417180416602813
0.895146326914331 -0.00787694725350313
0.896418396650804 0.0270025266038037
0.897503555887047 0.0627799325809078
0.898397435069818 0.0993112076913524
0.899096434863135 0.136449253383754
0.89959774064154 0.174044527855751
0.899899333823594 0.211945648206061
0.9 0.25
};
\draw (axis cs:0.02,0.98) node[
  scale=0.85,
  anchor=north west,
  text=black,
  rotate=0.0
]{$\gamma=$10.0};
\draw (axis cs:0.02,0.895) node[
  scale=0.85,
  anchor=north west,
  text=black,
  rotate=0.0
]{$\overline{C}(\sigma^*)=29.37$};
\end{axis}

\end{tikzpicture}